\documentclass[a4paper,fleqn,colorlinks,hidelinks]{cas-dc}

\usepackage[numbers]{natbib}

\def\tsc#1{\csdef{#1}{\textsc{\lowercase{#1}}\xspace}}
\tsc{WGM}
\tsc{QE}

\usepackage{array}
\usepackage{subcaption}
\usepackage{graphicx}
\usepackage{multirow}
\usepackage{paralist}

\usepackage{caption}
\usepackage{placeins}

\newcommand{\bx}{\mathbf{x}}
\newcommand{\by}{\mathbf{y}}
\usepackage{rotating}
\usepackage{amsmath,amsfonts}
\usepackage{lastpage}
\usepackage{float}

\begin{document}
\let\WriteBookmarks\relax
\def\floatpagepagefraction{1}
\def\textpagefraction{.001}
\shorttitle{Self-Supervised Sentinel-5P Super-Resolution}
\shortauthors{H.O. Ali et~al.}

\title [mode = title]{Self-Supervised Super-Resolution for Sentinel-5P Hyperspectral Images}                      


\author[label1,label2]{Hyam Omar Ali}
\author[label1,label3]{Antoine Crosnier}
\author[label1]{Romain Abraham}
\author[label4]{Baptiste Combelles}
\author[label4]{Fabrice J\'egou}
\author[label1,label5]{Bruno Galerne}

\affiliation[label1]{
    organization={Universit\'e d'Orl\'eans, Universit\'e de Tours, CNRS, IDP, UMR 7013},
    city={Orl\'eans},
    country={France}
}

\affiliation[label2]{
    organization={Faculty of Mathematical Sciences, University of Khartoum},
    city={Khartoum},
    country={Sudan}
}

\affiliation[label3]{
    organization={ENS Lyon},
    city={Lyon},
    country={France}
}

\affiliation[label4]{
    organization={Laboratory of Physics and Chemistry of the Environment and Space (LPC2E), CNRS UMR 7328, University of Orl\'eans},
    city={Orl\'eans},
    country={France}
}

\affiliation[label5]{
    organization={Institut universitaire de France (IUF)},
    country={France}
}

\begin{abstract}
Sentinel-5P (S5P) mission provides global hyperspectral observations for atmospheric and environmental monitoring; however, its native spatial resolution limits its use for applications requiring fine-scale spatial information. Existing super-resolution (SR) approaches for S5P rely on supervised learning. Because high-resolution (HR) data do not exist for S5P, these methods depend on synthetic low-resolution (LR) data, limiting their applicability to real observations. In this study, we propose a self-supervised hyperspectral SR framework tailored specifically for S5P that enables training directly from S5P observations without HR ground-truth (GT). The proposed framework combines Stein's Unbiased Risk Estimator with an Equivariant Imaging constraint while explicitly incorporating the S5P degradation operator and band-dependent noise statistics derived from sensor signal-to-noise ratio (SNR) metadata. We further investigate efficient Depthwise Separable Convolution-based architectures adapted to the spectral and spatial characteristics of S5P. The framework is evaluated in two complementary settings: 
\begin{inparaenum}[(i)]
    \item LR-HR, where synthetic LR images are used to recover the available GT, enabling a direct comparison between self‑supervised and supervised learning, and     
    \item GT-SHR, where native S5P observations are super-resolved beyond S5P’s spatial resolution without HR GT. 
\end{inparaenum}
In the LR-HR setting, the self-supervised models achieve performance comparable to their supervised counterparts while maintaining strong measurement consistency. In the GT-SHR setting, uncertainty analysis demonstrates stable reconstructions, while comparisons with alternative SR baselines show that the proposed models consistently achieve higher measurement consistency. Qualitative cross-sensor comparison with EMIT observations provides independent supporting evidence for the spatial plausibility of the reconstructed structures. These results demonstrate the potential of the proposed framework for enhancing real S5P observations where HR reference data are inherently unavailable. The code is available at \url{https://github.com/hyamomar/Sentinel-5P-Super-Resolution/tree/main/self_supervised}.
\end{abstract}


\maketitle

\section{Introduction}
In 2017, the European Space Agency (ESA) launched the Sentinel‑5 Precursor (S5P) mission for enhanced Earth observation capabilities \cite{s5p_mission}. S5P is designed for global monitoring of atmospheric composition, providing daily observations of main trace gases such NO\textsubscript{2}, CH\textsubscript{4}, and CO \cite{s5p_applications}, supporting air quality monitoring and climate studies. It is equipped with the TROPOspheric Monitoring Instrument (TROPOMI), a hyperspectral sensor operating across the Ultraviolet (UV), Visible (UVIS), Near-Infrared (NIR), and Short-Wave Infrared (SWIR) spectrometers. Each spectrometer represents two distinct spectral bands, resulting in eight bands (BD1-BD8) with almost $500$ spectral channels each \cite{s5p_products}. Despite S5P's high spectral resolution, it has a native spatial resolution of approximately $3.5 \times 5.5 \text{ km}^2$ which limits applications requiring fine spatial detail, such as localised emission monitoring and HR air quality assessments in complex urban environments or industrial clusters \cite{s5p_mission, TROPOMI_L01b_Specification, camps2011remote}. This limitation is further compounded by the band-dependent spatial responses and SNR characteristics, together with the absence of true HR observations for direct validation, making S5P SR particularly challenging.

Deep learning (DL)-based SR has become an effective approach for enhancing the spatial resolution of remote sensing imagery. In hyperspectral (HS) imaging, recent advances have demonstrated improvements in recovering spatial details while preserving spectral information \cite{wang2023hyperspectral}. For S5P, however, SR research remains limited, and the existing approaches rely on supervised learning (SL) with paired LR and HR training data. Since true HR observations do not exist for S5P, these methods construct synthetic LR-HR pairs by degrading the original observations using a sensor degradation model \cite{carbone2024model,ali2025depth}. Although these methods demonstrate encouraging reconstruction performance, they remain dependent on synthetic degradations and cannot be trained directly from native S5P observations without HR references.

Self-supervised learning (SSL) has emerged as a powerful framework for learning from data without requiring paired images. In remote sensing, SSL has attracted increasing attention due to the abundance of unlabeled Earth observation data and the difficulty of obtaining reliable GT \cite{wang2022self}. More recently, self-supervised reconstruction methods have demonstrated that image formation models and self-consistency constraints can be exploited to recover high-quality images without HR references \cite{scanvic2026scale}. These developments make SSL particularly attractive for S5P. However, self-supervised SR has not yet been investigated for S5P HS imagery.

To address this gap, we introduce the first self-supervised SR framework specifically designed for S5P HS imagery. Our framework adapts a self-supervised reconstruction strategy based on Stein's Unbiased Risk Estimator (SURE) and Equivariant Imaging (EI). Specifically, it incorporates a sensor-aware degradation model together with band-specific noise statistics derived from the instrument's SNR metadata, allowing the model to learn directly from native S5P observations while remaining consistent with the underlying image formation process. To efficiently handle the high spectral dimensionality of S5P data, we further investigate efficient architectures based on Depthwise Separable Convolutions (DSC), including multi-scale Unet and recursive DSC-based designs, providing an effective balance between reconstruction quality and computational efficiency. Our main contributions of this work are summarised as follows:
\begin{itemize}
\item We enable self-supervised hyperspectral SR of S5P observations directly from native data, without requiring paired LR-HR training data.
\item We adapt SURE and EI self-supervised reconstruction to the S5P acquisition process through a sensor-aware degradation model, band-specific noise estimation derived from the instrument's SNR metadata, and efficient DSC-based architectures tailored for S5P HS data.
\item We evaluate the framework in both LR-HR and GT-SHR settings, demonstrating competitive performance with supervised counterparts and providing complementary evidence of reconstruction stability and spatial plausibility through uncertainty analysis and coincident EMIT observations.
\end{itemize}

\section{Related Work}\label{LR_section}
SR has been extensively studied in remote sensing to overcome the spatial resolution limitations of satellite sensors. While traditional interpolation and regression methods are computationally efficient, they often struggle to recover fine spatial details and vulnerable to noise amplification, particularly for coarse-resolution imagery \cite{park2003super, wang2022review, qi2026advancing}. The emergence of DL fundamentally transformed SR by enabling data-driven models to learn complex LR–HR mappings. Convolutional neural networks (CNNs) established the foundation for modern single-image super-resolution (SISR) and consistently outperformed traditional methods \cite{dong2015image, wang2020deep}.

The success of DL has naturally extended to HS image SR, where the objective is to enhance spatial resolution while preserving the spectral information. Compared with RGB imagery, HS image SR is considerably more challenging due to its high spectral dimensionality and strong inter-band correlations, requiring models to jointly exploit spatial and spectral information \cite{he2023spectral, aburaed2023review}. 
Consequently, recent research has increasingly explored attention mechanisms, transformer architectures, and diffusion-based networks to better model spatial-spectral dependencies and improve reconstruction fidelity \cite{chen2023review, wang2023hyperspectral}. Despite these architectural advances, effective learning strategies for HS image SR remain an active research challenge.

The availability of suitable training data is of the main challenges in HS images SR. While supervised methods have achieved remarkable performance when paired LR-HR data are available, acquiring true HR HS observations is generally impractical for remote sensing sensors. This limitation has motivated increasing interest in unsupervised and SSL strategies that learn directly from observed data by exploiting image priors, measurement consistency, or the underlying imaging acquisition process to reduce or eliminate the dependence on paired training data.

Early self-supervised SR methods, including Deep Image Prior (DIP) \cite{ulyanov2018deep} and Zero-Shot Super-Resolution (ZSSR) \cite{shocher2018zero}, demonstrated that high-quality image reconstruction can be learned directly from a single LR image by exploiting internal image statistics and cross-scale recurrence, without external training datasets. However, these methods were originally developed for RGB images and do not explicitly model the high-dimensional spectral characteristics of HS imagery. 

Enhanced Deep Image Prior \cite{li2025enhanced} extends DIP to HS image SR through a multi-stage framework combining degradation estimation, interactive spectral learning, double U-Net reconstruction, and decision fusion. Similarly, M2U-Net \cite{li2024model} incorporates model-informed degradation constraints within a multi-stage optimisation framework to jointly estimate degradation and reconstruct HS images. However, these formulations rely on degradation estimation and additional spectral or multimodal components that are not directly applicable to the S5P SR. Deep Spectral Prior \cite{cheng2025deep} performs unsupervised image reconstruction in the Fourier domain, improving implicit regularisation without requiring manual early stopping; however, it targets generic image reconstruction rather than HS image SR. A related line of research formulates HS image SR as a hyperspectral-multispectral (HS-MS) fusion problem. uSDN \cite{qu2018unsupervised} introduced an unsupervised framework for HS-MS fusion, while subsequent methods exploited spectral-spatial mapping \cite{chen2021hyperspectral}, material consistency \cite{hong2023decoupled}, modality decoupling \cite{du2026unsupervised}, and spectral response function guidance \cite{shah2025spectralift}. Although these approaches achieve strong performance, they require complementary LR HS and HR multispectral observations, which are unavailable for S5P data.

More recently, image formation-aware reconstruction has emerged as a promising direction by incorporating measurement consistency and acquisition models into the learning objective rather than relying solely on image statistics. Representative examples include blind reconstruction frameworks based on measurement consistency \cite{bell2019blind, gu2019blind} and the recently proposed  Robust Equivariant Imaging (REI) framework \cite{chen2022robust}. However, these methods have not been adapted to the imaging characteristics, degradation process, and noise properties of the S5P.

Research on S5P SR remains relatively limited. S5Net introduced the first DL framework specifically designed for S5P by employing a CNN together with a spectral fine-tuning cascade \cite{carbone2024efficient}. While effective, S5Net reconstructs each spectral channel independently and requires geographically region-specific models, limiting its ability to exploit inter-channel dependencies and generalise across geographically diverse regions. More recently, S5-DSCR models \cite{ali2025depth} addressed these limitations by jointly reconstructing all spectral channels within a unified global model while substantially reducing computational complexity through DSC \cite{chollet2017xception}. Although these models improve both efficiency and reconstruction accuracy, they remain fully supervised and therefore depend on synthetically LR data.

In addition to the learning paradigm, computational efficiency remains an important consideration for HS images SR due to its high spectral dimensionality. Architectures based on DSC, recursive parameter sharing, and encoder-decoder designs have demonstrated favourable trade-offs between reconstruction accuracy and computational cost \cite{hung2019real, hussain2023depth, sun2024enhanced, muhammad2021multi, jiang2020single}. U-Net and its variants have also been widely adopted in HS image analysis because of their ability to capture multi-scale spatial representations \cite{bidari2024enhancing, moustafa2021hyperspectral, han2025dca, li2022semantic, gao2023high}. However, their application to HS SR remains relatively limited, particularly for S5P.

Collectively, the existing literature reveals three key limitations. First, current S5P SR methods remain fully supervised despite the absence of true HR observations. Second, existing self-supervised HS images SR methods are primarily designed for image-specific reconstruction or HS-MS fusion rather than native S5P. Third, although image-formation-aware reconstruction has recently shown considerable promise, these advances have not yet been adapted to the unique degradation characteristics and noise properties of S5P. These gaps motivate the proposed self-supervised framework for S5P SR, which combines a degradation model, noise-guided learning derived from sensor SNR metadata,  and an efficient DSC-based architecture.

\section{Proposed Method}\label{PM_section}
The objective of this study is to develop a self-supervised HS SR framework for S5P that learns spatial enhancement directly from degraded observations without relying on HR GT. For each spectral band, we consider two experimental settings. First, LR-HR setting, where the SL baseline maps LR inputs to HR GT, while the corresponding SSL model is trained using only LR observations. Second, GT-SHR setting, the available GT images are treated as degraded observations, and the SSL model learns to reconstruct enhanced spatial detail beyond the sensor's native resolution.

\subsection{SR Degradation Model and Noise Estimation} \label{DM_section} 
Let $ \bx\in \mathbb{R}^{H\times W\times C} $ denote an HR image with $C$ spectral channels, and $\by\in \mathbb{R}^{h\times w\times C}$ the corresponding LR image, where $h=H/s$ and $w=W/s$. A scaling factor $s=4$ was used for downsampling. The S5P image formation process is modelled separately for each spectral band as
\begin{equation}\label{eq:main}
\by = \mathcal{A}(\bx)+ \mathbf{n}
\end{equation}
where  $\mathcal{A}(.)$ is a known band-specific degradation operator representing the sensor’s spatial response and sampling process, and $\mathbf{n}$ denotes additive measurement noise.

The degradation operator $\mathcal{A}$ is adopted from the physically derived S5P imaging model proposed in \cite{carbone2024model}. It approximates the TROPOMI spatial response using a band-dependent asymmetric Gaussian PSF whose parameters are derived from the estimated detector modulation transfer function (MTF). Accordingly, the degradation operator is expressed as
\begin{equation}
\mathcal{A}(\bx)= (\bx*k)\downarrow s,
\end{equation}
where $k$ is the corresponding band-dependent asymmetric Gaussian blur kernel with different standard deviations in the along-track and cross-track directions. We use these operators throughout this work to generate degraded observations and to enforce data fidelity within the SSL framework by mapping the reconstructed images back to the measurement space during SSL training. 

S5P Level-1B (L1B) products provide \textit{radiance\_noise} metadata as SNR in decibels (dB). According to the official L1B specification \cite{KNMI2022TROPOMIL1BIODS}, this quantity is defined as $\mathrm{SNR}_{\mathrm{dB}}=10\log_{10}\left(\frac{L}{\sigma_n}\right)$ where $L$ is the measured radiance and $\sigma_n$ is its one-standard-deviation random error. We therefore convert the metadata to the linear domain $\mathrm{SNR}_{\mathrm{linear}}=10^{\frac{\mathrm{SNR}_{\mathrm{dB}}}{10}}$
and estimate the band-specific noise standard deviation as $\sigma=\frac{\mu}{\mathrm{SNR}_{\mathrm{linear}}}$ where $\mu$ is the global mean radiance over the training data. The resulting band-specific Gaussian noise level is used for LR simulation and SSL training.

While traditional SR relies on a supervised mapping $f_\theta:\by \mapsto \hat{\bx}$, the unavailability of $\bx$ for S5P necessitates an SSL framework. We train a reconstruction function $f_\theta$ using a loss function depending on $\by$, the known $\mathcal{A}$, and the statistical properties of the noise.

\subsection{Self-supervised Loss}\label{Loss_section}
Training a SR model without access to HR GT requires a loss function that learns only from the LR images. In this work, we adopt a self-supervised loss derived from the scale-equivariant imaging framework proposed in \cite{scanvic2026scale}, which builds upon the REI formulation in \cite{chen2022robust}. REI combines SURE, which provides an unbiased estimate of the measurement error under Gaussian noise, with an equivariance-based regularisation strategy inspired by Equivariant Imaging. REI exploits both the known degradation operator and the noise characteristics.

In the supervised SR, model parameters are learned by minimising the MSE between the reconstruction $f_\theta(\by)$ and the known HR image $\bx$, where $\by$ is the observed LR image:
$$ \mathcal{L}_{\text{MSE}} = \left\| f_{\theta}(\by) - \bx \right\|_2^{2}$$
Since $\bx$ is unavailable in the SSL framework, SURE provides an unbiased estimator of the MSE in the measurement space $$\mathbb{E}_{\bx,\by}\!\left[\left\| \mathcal{A}(f_{\theta}(\by)) - \mathcal{A}(\bx) \right\|_2^{2}\right]$$ using only the observed noisy measurements $\by$ rather than direct comparison with the GT. For $\hat{\bx}=f_\theta(\by)$, the SURE loss is given by:
\begin{equation} \label{eq:1}
\mathcal{L}_{\text{SURE}} = \left\| \mathcal{A}(\hat{\bx}) - \by \right\|_2^{2} - N\sigma^{2} + 2\sigma^{2} \mathrm{div}_{\by}\!\left(\mathcal{A}(f_{\theta}(\by))\right)
\end{equation}
where $N=hwC$ denotes the dimension of $\by$ and $\mathrm{div}(\cdot)$ is the divergence of the reconstruction with respect to the input. In practice, the divergence term is approximated using Monte Carlo perturbations \cite{chen2022robust}. By explicitly incorporating $\mathcal{A}$, the SURE loss enforces consistency with the degradation model while accounting for the measurement noise. 

While SURE enforces measurement consistency,  the framework in \cite{scanvic2026scale} introduces an additional equivariance constraint to regularise the reconstruction. Specifically, a spatial scaling transformation $\tau$ (zoom operator) is applied to the reconstructed image $\bx' = \tau(\hat{\bx})$, which is then degraded using the degradation operator $\by' = \mathcal{A}(\bx')$. The model reconstructs $\hat{\bx}' = f_\theta(\by')$, and equivariance is enforced by encouraging consistency between $\hat{\bx}'$ and $\bx'$. 
\begin{equation}
\mathcal{L}_{\text{EQ}} = \left\| f_\theta\big(\mathcal{A}(\tau(\hat{\bx}))\big) - \tau(\hat{\bx}) \right\|_2^2
\end{equation}
This regularisation encourages the model to produce reconstructions that remain stable under scale transformations \cite{scanvic2026scale}.

The resulting self-supervised loss combines the SURE data-consistency term with the equivariant regularisation:
\begin{equation} \label{eq:3}\mathcal{L}_{\text{SSL}} = \mathcal{L}_{\text{SURE}} + \lambda \mathcal{L}_{\text{EQ}}
\end{equation}
where $\lambda$ controls the strength of the equivariance regularisation relative to the SURE data-consistency term.

\subsection{Network Architectures}
To evaluate the proposed SSL framework across different architectural designs, we consider two CNN architecture families tailored to S5P SR: the Unet-S5P and the recursive DSC-based. The Unet-S5P family employs a multi-level encoder-decoder structure to learn multi-scale spatial representations, whereas the recursive DSC-based family performs progressive residual refinement through successive recursive blocks. 

Both architecture families follow two common design principles. First, all models employ DSC to reduce the number of parameters and computational cost while efficiently processing the high spectral dimensionality of S5P data. Each DSC module consists of a depthwise convolution, which processes each feature channel independently, followed by a pointwise $1\times1$ convolution that combines information across channels and controls the feature dimensionality. Second, all models adopt a pre-upsampling residual learning strategy, in which the LR input is first upsampled using bicubic interpolation and the network learns a residual correction representing the missing high-frequency spatial details.

\subsubsection{Unet-S5P Variants}
\begin{figure}[pos=h,width=\linewidth]
    \centering
    \subfloat[Overall Unet-S5P architecture]{
        \includegraphics[width=0.9\linewidth]{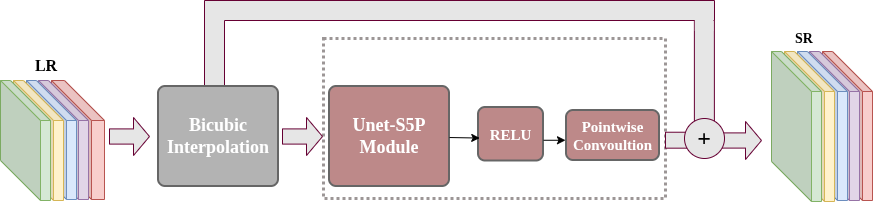}
        \label{global_fig}
    }
    \hfill
    \subfloat[Unet-S5P module showing the encoder-decoder and the internal blocks composed of successive DSC modules and ReLU activations]{
        \includegraphics[width=0.95\linewidth]{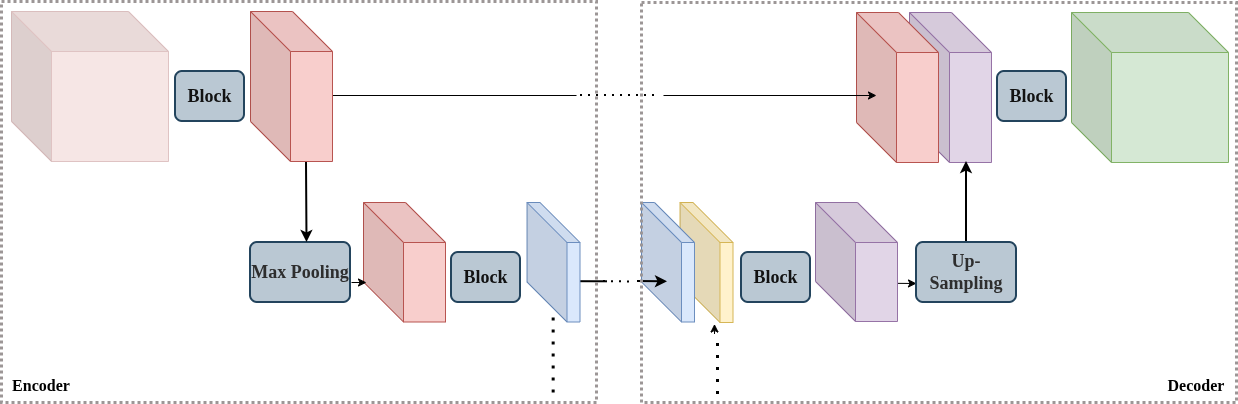}
        \label{unet_fig}
    }
    \caption{Overview of the proposed Unet-S5P architecture and its building blocks, illustrating the residual learning strategy for refining the interpolated input}
    \label{arch_fig}
\end{figure}

The Unet-S5P architecture is illustrated in Fig.\ref{arch_fig}. The overall network is shown in Fig.\ref{global_fig}, while Fig.\ref{unet_fig} details the encoder-decoder module and its internal blocks with skip connections between corresponding resolutions that facilitate the recovery of fine spatial details.

Unlike the classical U-Net architecture \cite{ronneberger2015u}, Unet-S5P incorporates several modifications to accommodate the characteristics of S5P hyperspectral data. The encoder progressively reduces the spatial resolution while compressing the feature representation, whereas the decoder progressively restores the spatial resolution and expands the feature representation. This inverted channel progression provides compact intermediate representations while retaining sufficient capacity for spatial reconstruction. Each encoder and decoder stage consists of successive DSC modules followed by ReLU activations. The channel configuration is progressively adjusted across stages to accommodate the different numbers of spectral channels across S5P bands, while maintaining the same scaling strategy across all evaluated bands to ensure architectural consistency and control the computational cost of intermediate representations. Detailed channel configurations for each spectral band are provided in \ref{App_unet}.

We consider two variants that differ in the number of encoder-decoder levels and the depth of the DSC blocks:
\begin{itemize}
    \item Unet-S5P-800k: A lightweight architecture with three encoder-decoder levels, providing multi-scale feature extraction with reduced computational cost.
    \item Unet-S5P-1M: A deeper architecture with four encoder-decoder levels, providing greater modelling capacity through richer hierarchical feature representations.
\end{itemize}
These variants enable assessment of the trade-off between reconstruction capacity and computational complexity. Both use 3$\times$3 depthwise kernels to balance receptive field size and computational efficiency.

\subsubsection{DSC-based Variants}
This architecture family consists of recursive DSC-based networks adapted from our previous supervised S5P SR architecture \cite{ali2025depth}, with minor modifications introduced for stable SSL training. Unlike the Unet-S5P, these models perform successive residual
refinement at a constant spatial resolution through recursive DSC blocks. To improve training stability within the SSL framework, batch normalisation layers and the final ReLU activation used in the original supervised architecture are removed. Two variants are considered:
\begin{itemize}
    \item S5-DSCR: A deeper architecture comprising five recursive DSC blocks separated by ReLU activations, providing hierarchical feature extraction and increased modelling capacity.
    \item S5-DSCR-S: A lightweight architecture consisting of a single DSC block without a ReLU activation. This results in an affine model that provides a low-complexity baseline for analysing the behaviour of the proposed noise-aware SSL framework.
\end{itemize}

Together, the four architectures span a range of model capacities and computational costs, enabling the proposed SSL framework to be evaluated under both multi-scale encoder-decoder and recursive refinement designs.

\subsection{Training Configuration}\label{section_config}
To ensure a fair comparison across the evaluated architectures and learning frameworks, all models were trained using the same data partitions and common training configuration, with the corresponding SL or SSL loss function applied to each framework.

Before training and evaluation, each spectral band was normalised using a channel-wise strategy, with each channel independently standardised using the training dataset's mean and standard deviation. This replaces the global normalisation strategy used in our previous work \cite{ali2025depth} and was found to provide more stable SSL optimisation and improved spectral fidelity. The resulting data were divided into non-overlapping training, validation, and testing subsets with proportions of 65\%, 20\%, and 15\%, respectively. Following the orbit-level split and image segmentation described in Section \ref{section_data}, the image segments were divided into non-overlapping patches. Patches of size $112\times112$ were used for most spectral bands, while $52\times52$ patches were used for BD7 and BD8.

Training was conducted independently for each spectral band to account for the band-dependent imaging characteristics of S5P. This allows the degradation model and noise estimation to be tailored to the characteristics of each band while maintaining a computationally efficient training procedure. Although this strategy does not explicitly exploit spectral correlations between bands, it avoids imposing a common degradation and noise model across bands with different acquisition characteristics.

The band-specific noise parameters used in the SSL experiments are summarised in Table \ref{SNR_table}, which reports the linear SNR derived from the S5P metadata and the corresponding global mean radiance $\mu$ computed over the training data.
\begin{table*}[!htbp] 
\centering
\caption{Linear SNR and global mean radiance ($\mu$) for the S5P bands used in this study}
\label{SNR_table}
\begin{tabular}{@{}l c c c c c c c@{}}
\hline
\textbf{Band} & BD2 & BD3 & BD4 & BD5 & BD6 & BD7 & BD8 \\
\hline
\textbf{$\text{SNR}_{\text{linear}}$}  & $239$ & $909$ & $1344$ & $1219$  & $1255$ & $285$ & $229$ \\
\hline
\textbf{$\mu$} 
& $7.88\times10^{-8}$ 
& $2.31\times10^{-7}$ 
& $4.25\times10^{-7}$ 
& $4.29\times10^{-7}$ 
& $4.10\times10^{-7}$ 
& $3.25\times10^{-8}$ 
& $2.23\times10^{-8}$ \\
\hline
\end{tabular}
\end{table*}

A batch size of 1 was used for all experiments for training the proposed models based on preliminary experiments. All models were trained for 40 epochs using the Adam optimiser with the default PyTorch parameters. Preliminary experiments with training schedules of up to 100 epochs showed that the training process generally stabilised after approximately 30 epochs; therefore, 40 epochs were selected without early stopping. The initial learning rate was set to $10^{-3}$ and reduced by a factor of $0.1$ when the validation loss did not improve for three consecutive epochs. The supervised models were trained using MSE loss, whereas the self-supervised models were trained using the proposed loss (Eq.\eqref{eq:3}) with $\lambda=1$, based on the sensitivity analysis in Section \ref{section_EI}. The same SSL configuration was used for the GT-SHR experiments. All experiments were conducted on an NVIDIA A100 80GB PCIe GPU using PyTorch 2.9.1 with CUDA 12.8.

In the original equivariant formulation \cite{scanvic2026scale}, mirror padding is used to preserve spatial dimensions during downscaling. However, mirror padding might introduce boundary repetitions that the model could unintentionally exploit. Circular padding was also discarded, as it would artificially wrap spatial content across image borders and may bias the SSL objective. Therefore, we adopt a no-padding strategy in all our experiments to preserve natural spatial statistics and avoid boundary-induced artefacts.

\section{Experiments}
\subsection{Datasets}\label{section_data}
All experiments were conducted using S5P L1B radiance data, freely accessible from Copernicus’ official platform as Level-1B radiance products \cite{CopernicusDataSpace}. The dataset comprises 9 complete S5P orbits acquired on 4 January 2023 and 7 September 2023, covering diverse geographical regions and atmospheric conditions to provide broad spatial and spectral variability. 

For all S5P spectral bands except BD1, the complete orbit products contain 4172-3735 scanlines in the along-track direction and 450-215 ground pixels in the across-track direction, depending on the band and geographical region. To make the data computationally manageable while preserving the complete across-track swath, each orbit was divided along the along-track direction into image segments of 512 scanlines, while retaining the full across-track width. The polar portions at the beginning and end of each orbit were subsequently discarded because their very low radiance values resulted in a high frequency of negative or outlier measurements.

In contrast to our previous work \cite{ali2025depth}, where the across-track width was cropped, the present strategy preserves the full swath width to maintain cross-track spatial continuity. This maximises the use of the available spatial information while providing suitable input dimensions for optimisation. Preserving the full across-track extent is also advantageous for enforcing equivariant consistency across spatial scales, which is a key component of the proposed SSL framework. BD1 is excluded because its full-swath across-track dimension contains only 77 pixels \cite{s5p_mission}, compared with 215 pixels for BD7-BD8 and 450 pixels for BD2-BD6. Consequently, this narrower swath provides insufficient spatial extent for the preprocessing and SR experiment.

\begin{figure}[pos=h,width=\linewidth]
    \centering
    \includegraphics[width=\linewidth]{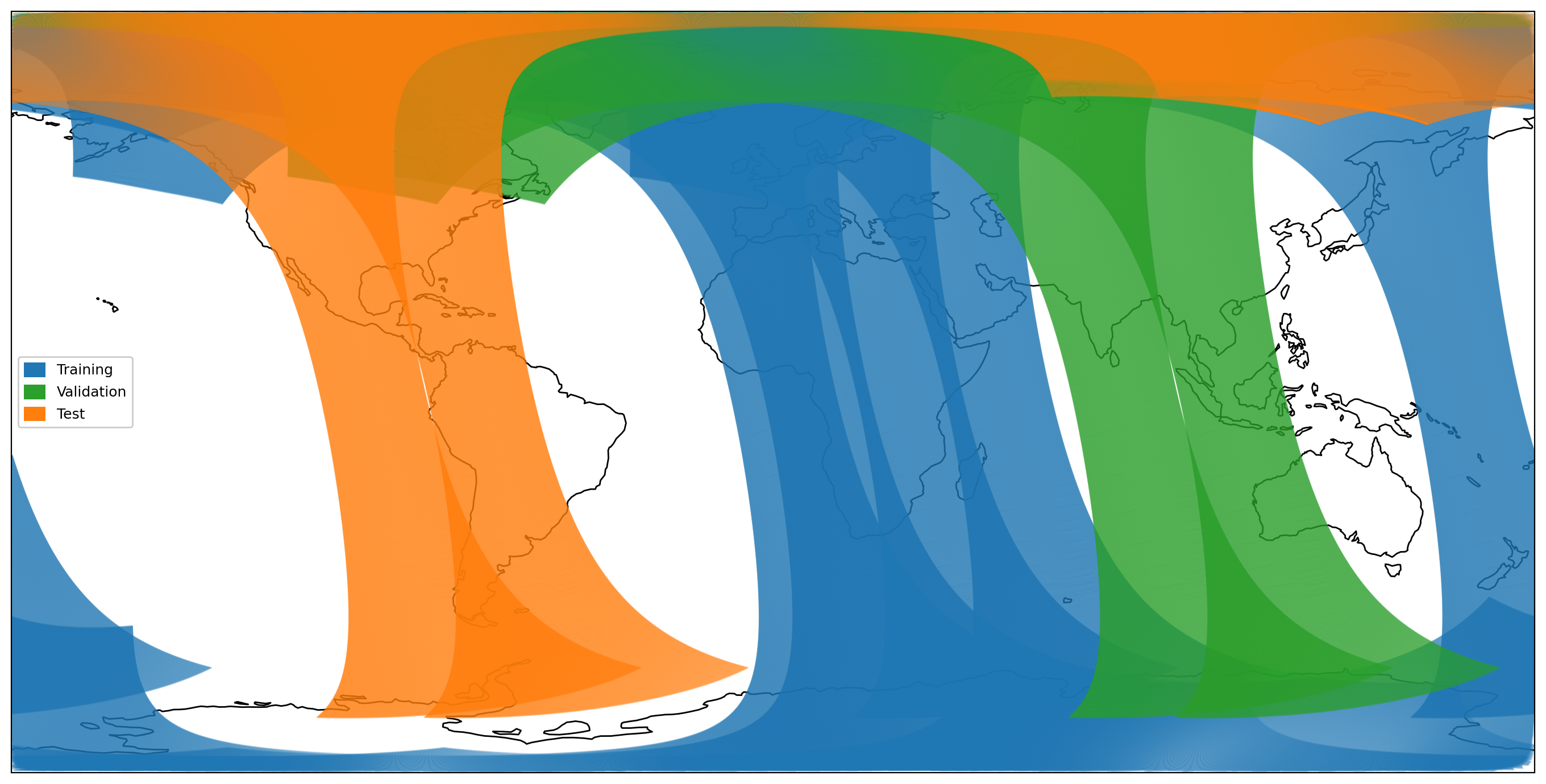}
    \caption{Geographical distribution and dataset assignment of the 9 S5P orbits used in this study}
    \label{data_split}
\end{figure}

To prevent spatial and orbital correlations between the training and evaluation data, the 9 complete orbits were partitioned into non-overlapping training, validation, and test sets before any image segmentation or patch extraction. Each complete orbit was assigned exclusively to one subset. Image segments of 512 scanlines and the corresponding LR patches were then generated independently within each subset. Consequently, no image segment or patch in the validation or test sets originates from an orbit used for training. The geographical distribution and subset assignment of the 9 orbits are shown in Fig.\ref{data_split}.

To remove corrupted pixels and ensure numerical stability during training, we implemented a data cleaning step before training and evaluation. We define a threshold $t = 10^{-2}$ based on the radiance range, which typically remains below  $10^{-4}$ but occasionally contains extreme outliers exceeding $10^{30}$. Negative radiance values greater than $-t$ were clipped to zero, and remaining outliers exceeding $t$ were replaced with the median of valid neighbouring pixels within a $3\times3$ window. Applying the median replacement preserves local spatial structure while preventing extreme values from biasing the learning and evaluation process.

For all the experiments and for each band, the simulated LR images were generated once and fixed across all models and training frameworks to ensure consistency and reproducibility. 

\subsection{Evaluation metrics}\label{section_M}
The evaluation protocol depends on the availability of HR GT. We employ two categories of metrics:
\begin{inparaenum}[(i)]
    \item Reference-based metrics: When HR GT is available, reconstruction fidelity is evaluated using four complementary metrics: Peak Signal-to-Noise Ratio (PSNR) and Structural Similarity Index (SSIM), Spatial Correlation Coefficient (SCC), and Learned Perceptual Image Patch Similarity (LPIPS) \cite{an2014orientation, wang2004image, zhang2018unreasonable}. PSNR and SSIM assess pixel‑level and structural similarity, SCC measures spectral correlation between the reconstructed image $\hat{\bx}$ and $\bx$, while LPIPS measures perceptual similarity using deep feature representations.
    
    \item No-reference metrics: When HR GT is unavailable, reconstruction quality is assessed using measurement consistency and sharpness. Measurement consistency quantifies the agreement between the degraded reconstruction $\mathcal{A}(\hat{\bx})$ and the observed LR image $\by$, that is how closely the reconstructed image reproduces the original observation after applying the degradation operator $\mathcal{A}$. Higher values indicate greater compatibility with the observed S5P measurement. Sharpness is measured using the mean magnitude of the Sobel gradient, which quantifies local spatial intensity variations and provides an indicator of fine spatial detail, as sharper structures and boundaries produce stronger gradient responses \cite{herrmann2020learning}.
    \end{inparaenum} 

The proposed framework is evaluated in two complementary settings:
\textbf{LR-HR} where synthetic LR images are generated from the available HR GT and used as input to the models. The reconstructed images are compared with the corresponding GT using both categories of metrics. \textbf{GT-SHR} where the available native S5P observations are treated as degraded observations, and the SSL model learns to reconstruct enhanced spatial detail beyond the sensor's native resolution. Since no HR reference exists for these outputs, evaluation relies on the no-reference metrics. Qualitative comparisons with bicubic interpolation and alternative SR methods are additionally used to assess spatial detail and potential reconstruction artefacts.

\subsection{Ablation Analysis} \label{section_denoise}
All ablation analysis are conducted on four representative spectral bands (BD2, BD4, BD5, and BD7), with one band selected from each spectrometer. The analyses use the Unet-S5P architectures, which demonstrated stronger overall performance than the DSC-based architectures in the main evaluation. Results are reported on the
test dataset.

\subsubsection{Equivariant Regularization}\label{section_EI}
We first assess the contribution of the equivariant regularisation by comparing the full proposed loss ($\lambda=1$) with SURE alone ($\lambda=0$), where the equivariant term is removed. The results are reported in Table \ref{tab:eq_ablation}.

Compared with the SURE-only formulation, incorporating the equivariant regularisation consistently improves the reconstruction metrics (PSNR, SCC and SSIM) across the evaluated bands, demonstrating that the equivariant constraint provides complementary information beyond the measurement-consistency enforced by the SURE. In contrast, the SURE-only model ($\lambda=0$) achieves higher measurement consistency and, for some bands, slightly higher sharpness scores. This behaviour is expected since the SURE loss primarily enforces fidelity to the observed LR measurements, whereas the equivariant regularisation introduces an additional constraint that encourages stable reconstructions under scale transformations. Consequently, introducing the equivariant term introduces a trade-off between measurement consistency and reconstruction fidelity. 
\begin{table*}[!htbp]
\centering
\caption{Contribution of the equivariant regularisation. The best results are highlighted in bold. Bicubic is included as a reference}
\label{tab:eq_ablation}
\begin{tabular}{@{}lc|cc|cc|cc|cc|cc|cc@{}}
\hline
Band & \multicolumn{1}{c|}{Model} & \multicolumn{2}{c|}{PSNR$\uparrow$}      & \multicolumn{2}{c|}{SCC$\uparrow$}       & \multicolumn{2}{c|}{SSIM$\uparrow$}      & \multicolumn{2}{c|}{LPIPS$\downarrow$}     & \multicolumn{2}{c|}{Consistency$\uparrow$} & \multicolumn{2}{c}{Sharpness$\uparrow$} \\
\multicolumn{1}{l}{}     &    & $\lambda=0$   & $\lambda=1$ & $\lambda=0$ & $\lambda=1$  & $\lambda=0$  & $\lambda=1$ & $\lambda=0$ & $\lambda=1$      & $\lambda=0$ & $\lambda=1$   & $\lambda=0$  & $\lambda=1$ \\ \hline
\multirow{3}{*}{BD2} & Bicubic       & \multicolumn{1}{l|}{47.08}  & 47.08           & \multicolumn{1}{l|}{0.825} & 0.825          & \multicolumn{1}{l|}{0.940} & 0.943     & \multicolumn{1}{l|}{0.143}           & 0.143  & \multicolumn{1}{l|}{52.78}          & 52.78 & \multicolumn{1}{l|}{0.380}          & 0.378 \\ 
                               & Unet-S5P-800k & \multicolumn{1}{l|}{46.31} & \textbf{47.42} & \multicolumn{1}{l|}{0.748} & \textbf{0.821} & \multicolumn{1}{l|}{0.941} & \textbf{0.954} & \multicolumn{1}{l|}{\textbf{0.070}} & 0.097 & \multicolumn{1}{l|}{\textbf{66.71}} & 61.44 & \multicolumn{1}{l|}{\textbf{0.628}} & 0.504 \\ 
                               & Unet-S5P-1M   & \multicolumn{1}{l|}{45.96}   & \textbf{47.28} & \multicolumn{1}{l|}{0.732} & \textbf{0.810} & \multicolumn{1}{l|}{0.939}  & \textbf{0.953} & \multicolumn{1}{l|}{\textbf{0.072}} & 0.103  & \multicolumn{1}{l|}{\textbf{66.34}} & 59.51 & \multicolumn{1}{l|}{\textbf{0.630}} & 0.493 \\ \hline
\multirow{3}{*}{BD4} & Bicubic       & \multicolumn{1}{l|}{27.52} & 27.52    & \multicolumn{1}{l|}{0.824} & 0.824          & \multicolumn{1}{l|}{0.740} & 0.740          & \multicolumn{1}{l|}{0.327}           & 0.327  & \multicolumn{1}{l|}{36.33}          & 36.33 & \multicolumn{1}{l|}{0.385}          & 0.385 \\
                               & Unet-S5P-800k & \multicolumn{1}{l|}{27.86} & \textbf{28.25} & \multicolumn{1}{l|}{0.819} & \textbf{0.845} & \multicolumn{1}{l|}{0.762} & \textbf{0.782} & \multicolumn{1}{l|}{\textbf{0.214}}  & 0.249 & \multicolumn{1}{l|}{\textbf{58.61}} & 50.76 & \multicolumn{1}{l|}{\textbf{0.550}} & 0.499 \\ 
                               & Unet-S5P-1M   & \multicolumn{1}{l|}{27.84} & \textbf{28.32} & \multicolumn{1}{l|}{0.819} & \textbf{0.847} & \multicolumn{1}{l|}{0.764} & \textbf{0.785} & \multicolumn{1}{l|}{\textbf{0.230}}  & 0.241  & \multicolumn{1}{l|}{\textbf{59.82}} & 52.73  & \multicolumn{1}{l|}{\textbf{0.549}} & 0.503 \\ \hline
\multirow{3}{*}{BD5} & Bicubic       & \multicolumn{1}{l|}{24.66} & 24.66     & \multicolumn{1}{l|}{0.818} & 0.818         & \multicolumn{1}{l|}{0.665} & 0.665          & \multicolumn{1}{l|}{0.373}      & 0.373   & \multicolumn{1}{l|}{34.20}          & 34.20 & \multicolumn{1}{l|}{0.391}          & 0.391 \\ 
                               & Unet-S5P-800k & \multicolumn{1}{l|}{25.06} & \textbf{25.41} & \multicolumn{1}{l|}{0.816}  & \textbf{0.840} & \multicolumn{1}{l|}{0.697} & \textbf{0.716} & \multicolumn{1}{l|}{\textbf{0.244}}  & 0.291  & \multicolumn{1}{l|}{\textbf{56.65}} & 48.17  & \multicolumn{1}{l|}{\textbf{0.558}} & 0.508 \\ 
                               & Unet-S5P-1M   & \multicolumn{1}{l|}{25.09} & \textbf{25.48} & \multicolumn{1}{l|}{0.821} & \textbf{0.841} & \multicolumn{1}{l|}{0.697} & \textbf{0.721} & \multicolumn{1}{l|}{\textbf{0.247}}  & 0.284  & \multicolumn{1}{l|}{\textbf{57.08}} & 50.51 & \multicolumn{1}{l|}{\textbf{0.557}} & 0.509  \\ \hline
\multirow{3}{*}{BD7} & Bicubic       & \multicolumn{1}{l|}{24.48} & 24.48   & \multicolumn{1}{l|}{0.738} & 0.738        & \multicolumn{1}{l|}{0.565} & 0.565          & \multicolumn{1}{l|}{0.405}           & 0.405  & \multicolumn{1}{l|}{32.79}          & 32.79 & \multicolumn{1}{l|}{0.330}          & 0.330 \\ 
                               & Unet-S5P-800k & \multicolumn{1}{l|}{21.84} & \textbf{24.12}  & \multicolumn{1}{l|}{0.591} & \textbf{0.663} & \multicolumn{1}{l|}{0.563} & \textbf{0.623}  & \multicolumn{1}{l|}{\textbf{0.211}}  & 0.325   & \multicolumn{1}{l|}{\textbf{48.28}} & 40.89 & \multicolumn{1}{l|}{\textbf{1.02}} & 0.554 \\ 
                               & Unet-S5P-1M   & \multicolumn{1}{l|}{22.64} & \textbf{23.97} & \multicolumn{1}{l|}{0.626} & \textbf{0.677} & \multicolumn{1}{l|}{0.595} & \textbf{0.635} & \multicolumn{1}{l|}{\textbf{0.207}}  & 0.311  & \multicolumn{1}{l|}{\textbf{50.89}} & 42.97 & \multicolumn{1}{l|}{\textbf{0.884}} & 0.575  \\ \hline
\end{tabular}
\end{table*}

Having established the contribution of the equivariant regularisation, we further investigate the influence of its weighting parameter over several orders of magnitude. We evaluate $\lambda\in\{0.01,0.1,1,10,100\}$, with the results reported in Table \ref{tab:lambda_ablation}. For compactness, only PSNR and SSIM are reported, as they provide complementary measures of pixel-level fidelity and structural similarity. The remaining evaluation metrics were also computed for all $\lambda$ values but are omitted from the table due to space limitations.

The results show that very small values of $\lambda$ generally lead to lower reconstruction performance. Increasing $\lambda$ from $0.01$ to $1$ generally improves both PSNR and SSIM, indicating that the equivariant constraint provides a meaningful contribution when sufficiently weighted. However, increasing $\lambda$ beyond $1$ does not produce a consistent improvement. Thus, no single $\lambda$ consistently maximises PSNR or SSIM across bands and architectures. Rather than performing band and architecture-specific tuning, we retain $\lambda=1$ throughout this work, consistent with the original framework \cite{scanvic2026scale}.
\begin{table*}[pos=h]
\centering
\caption{Sensitivity analysis of the weighting parameter $\lambda$}
\label{tab:lambda_ablation}
\begin{tabular}{@{}lc|ccccc|ccccc@{}}
\hline
\multirow{2}{*}{Band} & \multirow{2}{*}{Model} & \multicolumn{5}{c|}{PSNR$\uparrow$} & \multicolumn{5}{c}{SSIM$\uparrow$} \\ 

& & \multicolumn{1}{c}{$\lambda=0.01$} & \multicolumn{1}{c}{$0.1$} & \multicolumn{1}{c}{$1$} & \multicolumn{1}{c}{$10$} & $100$ 
& \multicolumn{1}{c}{$\lambda=0.01$} & \multicolumn{1}{c}{$0.1$} & \multicolumn{1}{c}{$1$} & \multicolumn{1}{c}{$10$} & $100$ \\ \hline

\multirow{3}{*}{BD2} & Bicubic
& \multicolumn{1}{c|}{47.08} & \multicolumn{1}{c|}{47.08} & \multicolumn{1}{c|}{47.08} & \multicolumn{1}{c|}{47.08} & 47.08
& \multicolumn{1}{c|}{0.943} & \multicolumn{1}{c|}{0.943} & \multicolumn{1}{c|}{0.943} & \multicolumn{1}{c|}{0.943} & 0.943 \\ 

& Unet-S5P-800k
& \multicolumn{1}{c|}{47.02} & \multicolumn{1}{c|}{47.39} & \multicolumn{1}{c|}{\underline{47.42}} & \multicolumn{1}{c|}{\textbf{47.56}} & 47.41
& \multicolumn{1}{c|}{0.951} & \multicolumn{1}{c|}{\textbf{0.955}} & \multicolumn{1}{c|}{0.954} & \multicolumn{1}{c|}{0.955} & \underline{0.955} \\ 

& Unet-S5P-1M
& \multicolumn{1}{c|}{46.82} & \multicolumn{1}{c|}{47.04} & \multicolumn{1}{c|}{47.28} & \multicolumn{1}{c|}{\textbf{47.59}} & \underline{47.35}
& \multicolumn{1}{c|}{0.948} & \multicolumn{1}{c|}{0.952} & \multicolumn{1}{c|}{0.953} & \multicolumn{1}{c|}{\textbf{0.956}} & \underline{0.956} \\ \hline

\multirow{3}{*}{BD4} & Bicubic
& \multicolumn{1}{c|}{27.52} & \multicolumn{1}{c|}{27.52} & \multicolumn{1}{c|}{27.52} & \multicolumn{1}{c|}{27.52} & 27.52
& \multicolumn{1}{c|}{0.740} & \multicolumn{1}{c|}{0.740} & \multicolumn{1}{c|}{0.740} & \multicolumn{1}{c|}{0.740} & 0.740 \\ 

& Unet-S5P-800k
& \multicolumn{1}{c|}{28.10} & \multicolumn{1}{c|}{28.24} & \multicolumn{1}{c|}{\underline{28.25}} & \multicolumn{1}{c|}{28.23} & \textbf{28.27}
& \multicolumn{1}{c|}{0.774} & \multicolumn{1}{c|}{0.781} & \multicolumn{1}{c|}{\underline{0.782}} & \multicolumn{1}{c|}{0.780} & \textbf{0.783} \\ 

& Unet-S5P-1M
& \multicolumn{1}{c|}{28.01} & \multicolumn{1}{c|}{28.22} & \multicolumn{1}{c|}{\textbf{28.32}} & \multicolumn{1}{c|}{28.28} & \underline{28.29}
& \multicolumn{1}{c|}{0.770} & \multicolumn{1}{c|}{0.780} & \multicolumn{1}{c|}{\textbf{0.785}} & \multicolumn{1}{c|}{0.782} & \underline{0.783} \\ \hline

\multirow{3}{*}{BD5} & Bicubic
& \multicolumn{1}{c|}{24.66} & \multicolumn{1}{c|}{24.66} & \multicolumn{1}{c|}{24.66} & \multicolumn{1}{c|}{24.66} & 24.66
& \multicolumn{1}{c|}{0.665} & \multicolumn{1}{c|}{0.665} & \multicolumn{1}{c|}{0.665} & \multicolumn{1}{c|}{0.665} & 0.665 \\ 

& Unet-S5P-800k
& \multicolumn{1}{c|}{25.31} & \multicolumn{1}{c|}{25.41} & \multicolumn{1}{c|}{25.41} & \multicolumn{1}{c|}{\underline{25.45}} & \textbf{25.46}
& \multicolumn{1}{c|}{0.711} & \multicolumn{1}{c|}{0.717} & \multicolumn{1}{c|}{0.716} & \multicolumn{1}{c|}{\textbf{0.719}} & \underline{0.719} \\  

& Unet-S5P-1M
& \multicolumn{1}{c|}{25.25} & \multicolumn{1}{c|}{25.37} & \multicolumn{1}{c|}{\underline{25.48}} & \multicolumn{1}{c|}{\textbf{25.49}} & 25.45
& \multicolumn{1}{c|}{0.707} & \multicolumn{1}{c|}{0.715} & \multicolumn{1}{c|}{\textbf{0.721}} & \multicolumn{1}{c|}{\underline{0.721}} & 0.719 \\ \hline

\multirow{3}{*}{BD7} & Bicubic
& \multicolumn{1}{c|}{24.48} & \multicolumn{1}{c|}{24.48} & \multicolumn{1}{c|}{24.48} & \multicolumn{1}{c|}{24.48} & 24.48
& \multicolumn{1}{c|}{0.565} & \multicolumn{1}{c|}{0.565} & \multicolumn{1}{c|}{0.565} & \multicolumn{1}{c|}{0.565} & 0.565 \\ 

& Unet-S5P-800k
& \multicolumn{1}{c|}{\underline{23.92}} & \multicolumn{1}{c|}{23.89} & \multicolumn{1}{c|}{\textbf{24.12}} & \multicolumn{1}{c|}{23.79} & 23.83
& \multicolumn{1}{c|}{0.653} & \multicolumn{1}{c|}{0.659} & \multicolumn{1}{c|}{0.623} & \multicolumn{1}{c|}{\underline{0.662}} & \textbf{0.664} \\ 

& Unet-S5P-1M
& \multicolumn{1}{c|}{23.50} & \multicolumn{1}{c|}{\underline{23.55}} & \multicolumn{1}{c|}{\textbf{23.97}} & \multicolumn{1}{c|}{22.74} & 23.06
& \multicolumn{1}{c|}{0.664} & \multicolumn{1}{c|}{0.664} & \multicolumn{1}{c|}{0.635} & \multicolumn{1}{c|}{\underline{0.665}} & \textbf{0.666} \\ \hline

\end{tabular}
\end{table*}

\subsubsection{Impact of the SURE Denoising Term}\label{section_sure}
The proposed SSL frameworks are based on the REI discussed in Section \ref{Loss_section}, where the loss combines a SURE-based data consistency term with an equivariant regularisation. The SURE term inherently introduces a denoising effect, which depends on the magnitude of the noise $\sigma$. For the S5P data considered in this study, the noise levels are small (Section \ref{section_config}). As a result, the SURE term is effectively dominated by the measurement fidelity component, with a limited contribution from denoising.

Because the S5P noise level is small, the denoising component is expected to have a limited effect. Nevertheless, its contribution is evaluated because jointly performing denoising and SR may affect reconstruction behaviour and reference-based metrics. To investigate these effects, we compare the REI framework with a simplified variant where $\sigma=0$ in the SURE term. This removes the denoising effect while preserving the overall SSL pipeline and the measurement consistency constraint.

Table \ref{Denoising_SSL} summarises the results. The differences between the two variants are consistently small across all bands, models, and metrics. Also, no systematic degradation or improvement was observed. These results confirm that in the proposed SSL framework, the optimisation is primarily driven by the measurement fidelity term. This behaviour is consistent with the low noise magnitude in the data. Although the denoising term has a limited effect on the S5P observations considered here, it is retained to account for measurement uncertainty and noise variations, and remains relevant to other sensors where noise levels may be more significant.

Finally, we highlight that challenging bands such as BD7, characterised by relatively lower SNR  and reduced reconstruction performance, do not show a consistent improvement when the denoising term is removed. The ablation results therefore indicate that the performance limitations are not caused by the denoising term, but are instead governed by the intrinsic characteristics of the data, as further discussed in the next section.

\begin{table*}[pos=h]
\centering
\caption{Impact of the denoising term in the SSL framework in the LR-HR setting. REI denotes the proposed framework, while $\sigma=0$ represents the variant without the denoising term. Bold values indicate the best performance between the two variants for each model and metric}
\label{Denoising_SSL}
\begin{tabular}{@{}lc|cc|cc|cc|cc|cc|cc@{}}
\hline
Band & \multicolumn{1}{c|}{Model} & \multicolumn{2}{c|}{PSNR$\uparrow$}      & \multicolumn{2}{c|}{SCC$\uparrow$}       & \multicolumn{2}{c|}{SSIM$\uparrow$}      & \multicolumn{2}{c|}{LPIPS$\downarrow$}     & \multicolumn{2}{c|}{Consistency$\uparrow$} & \multicolumn{2}{c}{Sharpness$\uparrow$} \\
\multicolumn{1}{l}{}     &                            & REI            & $\sigma=0$             & REI            & $\sigma=0$             & REI            & $\sigma=0$             & REI            & $\sigma=0$             & REI             & $\sigma=0$              & REI            & $\sigma=0$             \\ \hline
\multirow{3}{*}{BD2}    & Bicubic                    & 47.08          & 47.08          & 0.825          & 0.825          & 0.943          & 0.943          & 0.143          & 0.143          & 52.78           & 52.78           & 0.378          & 0.378          \\
                         & Unet-S5P-800k              & \textbf{47.42} & 47.25          & \textbf{0.821} & 0.808          & \textbf{0.954} & 0.953          & \textbf{0.097} & 0.103          & \textbf{61.44}  & 60.08           & \textbf{0.504} & 0.491          \\
                         & Unet-S5P-1M                & 47.28          & \textbf{47.35} & 0.801          & \textbf{0.813} & 0.953          & \textbf{0.954} & 0.103          & \textbf{0.099} & 59.51           & \textbf{60.66}  & 0.493          & \textbf{0.503} \\ \hline
\multirow{3}{*}{BD4}    & Bicubic                    & 27.52          & 27.52          & 0.824          & 0.824          & 0.740          & 0.740          & 0.327          & 0.327          & 36.33           & 36.33           & 0.385          & 0.385          \\
                         & Unet-S5P-800k              & 28.25          & \textbf{28.25} & \textbf{0.845} & \textbf{0.845} & \textbf{0.782} & \textbf{0.782} & 0.249          & \textbf{0.249} & \textbf{50.76}  & 50.35           & \textbf{0.499} & 0.498          \\
                         & Unet-S5P-1M                & 28.32          & \textbf{28.33} & 0.847          & \textbf{0.848} & \textbf{0.785} & 0.784          & \textbf{0.241} & 0.244          & \textbf{52.73}  & 52.45           & \textbf{0.503} & 0.502          \\ \hline
\multirow{3}{*}{BD5}    & Bicubic                    & 24.66          & 24.66          & 0.818          & 0.818          & 0.665          & 0.665          & 0.373          & 0.373          & 34.20           & 34.20           & 0.391          & 0.391          \\
                         & Unet-S5P-800k              & 25.41          & \textbf{25.44} & 0.840          & \textbf{0.840} & 0.716          & \textbf{0.719} & 0.291          & \textbf{0.290} & 48.17           & \textbf{48.62}  & \textbf{0.508} & 0.505          \\
                         & Unet-S5P-1M                & \textbf{25.48} & 25.46          & \textbf{0.842} & 0.841          & \textbf{0.721} & 0.719          & \textbf{0.284} & 0.288          & \textbf{50.51}  & 49.91           & \textbf{0.509} & 0.508          \\ \hline
\multirow{3}{*}{BD7}    & Bicubic                    & 24.48          & 24.48          & 0.738          & 0.738          & 0.565          & 0.565          & 0.405          & 0.405          & 32.79           & 32.79           & 0.330          & 0.330          \\
                         & Unet-S5P-800k              & \textbf{24.14} & 23.94          & \textbf{0.688} & 0.678          & \textbf{0.658} & 0.660          & 0.286          & \textbf{0.275} & 45.76           & \textbf{47.21}  & 0.623          & \textbf{0.645} \\
                         & Unet-S5P-1M                & \textbf{23.86} & 22.74          & \textbf{0.671} & 0.573          & 0.662          & \textbf{0.666} & 0.270          & \textbf{0.267} & 47.59           & \textbf{48.66}  & 0.654          & \textbf{0.698} \\ \hline
\end{tabular}
\end{table*}

\section{Results}
\subsection{Quantitative Comparison for the LR-HR Setting}
We report the quantitative performance of the proposed SR frameworks in the LR-HR setting using the reference-based metrics described in Section \ref{section_M}. Both SL and SSL frameworks are evaluated under identical conditions to assess the impact of the learning strategy while controlling for network architecture. S5Net serves as the existing SL baseline, while the proposed architectures are evaluated under both SL and SSL. Bicubic interpolation is included as a reference baseline, reflecting the native spatial resolution of the S5P observations and providing an indication of the intrinsic SR difficulty of each spectral band. DIP and ZSSR are additionally included as generic self-supervised SR baselines.

Table \ref{LRGT_table} summarises the results across BD2-BD8, including bicubic interpolation, S5Net, DIP, and ZSSR as reference baselines. The proposed SL models substantially outperform bicubic interpolation across the evaluated bands. The SSL models exhibit a moderate and band-dependent reduction in most reference-based metrics compared with their SL counterparts, while maintaining competitive perceptual quality. Measurement consistency remains high for both SL and SSL, with the SSL models achieving comparable or higher consistency than their corresponding SL models for most bands. 

For comparison with an existing S5P supervised approach, we evaluate S5Net, which was trained using region-specific models for Egypt, India, and the US \cite{carbone2024efficient}. We selected the US model based on its best average PSNR on the validation dataset. S5Net improves upon bicubic interpolation for some bands but remains below the proposed models under the SL framework. Notably, for several bands, the supervised S5Net model achieves performance comparable to or lower than that of the proposed SSL models. BD7 and BD8 are excluded from this comparison because S5Net applies an initial interpolation step before SR, resulting in a different upscaling factor from that used in our evaluation.

The comparison with DIP and ZSSR provides an additional assessment against generic self-supervised SR methods. DIP and ZSSR were originally developed for natural RGB images and are here adapted to the S5P setting using the same degradation operator $\mathcal{A}$. As shown in Table \ref{LRGT_table}, DIP consistently achieves lower PSNR, SCC, and SSIM than the proposed SSL models across all bands. ZSSR achieves competitive PSNR for BD2-BD6 but generally remains below the proposed models, except for BD8, where it achieves higher PSNR. In terms of LPIPS, the proposed models outperform both DIP and ZSSR for BD2-BD4 and BD7, while DIP achieves lower LPIPS for BD5, BD6, and BD8. In contrast, the proposed models consistently achieve higher measurement consistency than both baselines across all bands. Sharpness shows a different behaviour; DIP achieves higher values for several bands, whereas ZSSR remains consistently lower than the proposed models. The higher sharpness of DIP should be interpreted with caution, as its spectral bias can progressively fit high-frequency components, including noise \cite{shi2022measuring}. Thus, higher sharpness does not necessarily imply better reconstruction fidelity, as high-frequency artefacts can also increase the sharpness measure. Overall, the proposed framework provides more consistent reconstruction behaviour, particularly in terms of measurement consistency.

\begin{table*}[!htbp]
\centering
\caption{LR-HR quantitative results across all spectral bands. For each band and each metric, the best results for each training setting (SL and SSL) are shown in bold and the second-best are underlined}\label{LRGT_table}
\resizebox{\textwidth}{!}{
\begin{tabular}{@{}lc|cc|cc|cc|cc|cc|cc@{}}
\hline
Band & \multicolumn{1}{c|}{Model} & \multicolumn{2}{c|}{PSNR$\uparrow$}      & \multicolumn{2}{c|}{SCC$\uparrow$}       & \multicolumn{2}{c|}{SSIM$\uparrow$}      & \multicolumn{2}{c|}{LPIPS$\downarrow$}     & \multicolumn{2}{c|}{Consistency$\uparrow$} & \multicolumn{2}{c}{Sharpness$\uparrow$} \\ 
\multicolumn{1}{l}{}     &                            & SL             & SSL            & SL             & SSL            & SL             & SSL            & SL             & SSL            & SL              & SSL             & SL             & SSL            \\ \hline
\multirow{8}{*}{BD2}  & Bicubic                    & 47.08          & 47.08          & 0.825          & \textbf{0.825} & 0.943          & 0.943          & 0.143          & 0.143          & 52.78           & 52.78           & 0.378          & 0.378          \\
                         & S5Net \cite{carbone2024efficient}                      & 41.84          & -              & 0.723          & -              & 0.923          & -              & 0.107          & -              & 45.86           & -               & \textbf{0.547} & -              \\
                         & DIP \cite{ulyanov2018deep}     & - & 42.02 & - & 0.719 & - &0.837 & - & 0.253 & - & 49.14 & - & \textbf{0.689} \\
                         & ZSSR \cite{shocher2018zero}   & -  & 46.14   & -  & 0.798  & - & 0.934 & - & 0.160 &    -  &  51.17  & - & 0.340 \\                       
                         & S5-DSCR-S                  & 47.45          & 47.14          & 0.824          & 0.817          & 0.946          & 0.944          & 0.134          & 0.140          & 53.80           & 53.20           & 0.398          & 0.390          \\
                         & S5-DSCR                    & 48.65          & \underline{47.33}    & 0.869          & 0.805          & 0.958          & \underline{0.953}    & 0.094          & \textbf{0.096} & 55.69           & 59.16           & 0.482          & 0.494    \\
                         & Unet-S5P-800k              & \underline{49.13}    & \textbf{47.42} & \underline{0.879}    & \underline{0.821}    & \underline{0.962}    & \textbf{0.954} & \textbf{0.079} & \underline{0.097}    & \underline{56.07}     & \textbf{61.44}  & 0.509          & \underline{0.504} \\
                         & Unet-S5P-1M                & \textbf{49.16} & 47.28          & \textbf{0.881} & 0.801          & \textbf{0.962} & 0.953          & \underline{0.080}    & 0.103          & \textbf{56.42}  & \underline{59.51}     & \underline{0.511}    & 0.493          \\ \hline
\multirow{8}{*}{BD3}  & Bicubic                    & 37.90          & 37.90          & 0.831          & 0.831          & 0.884          & 0.884          & 0.206          & 0.206          & 45.19           & 45.19           & 0.382          & 0.382          \\
                         & S5Net \cite{carbone2024efficient}                      & 36.07          & -              & 0.783          & -              & 0.873          & -              & 0.171          & -              & 42.15           & -               & \textbf{0.518} & -              \\
                         & DIP \cite{ulyanov2018deep}     & -  &  34.29  & -  & 0.762  & - & 0.772 & - & 0.262 &   -   & 40.30   & - & \textbf{0.490} \\
                         & ZSSR \cite{shocher2018zero}   & -  &  36.62  & -  & 0.801  & - & 0.864 & - & 0.221 &    -  & 41.59   & - & 0.331 \\                            
                         & S5-DSCR-S                  & 38.36          & \underline{38.25}    & 0.834          & 0.832          & 0.892          & 0.890          & 0.187          & 0.194          & 47.95           & 47.53           & 0.416          & 0.410          \\
                         & S5-DSCR                    & 38.81          & 38.11          & 0.854          & 0.831          & 0.902          & 0.897          & 0.152          & \underline{0.152}    & 49.80           & \underline{54.04}     & 0.463          & 0.479          \\
                         & Unet-S5P-800k              & \underline{38.99}    & 38.18          & \underline{0.858}    & \underline{0.836}    & \underline{0.905}    & \underline{0.897}    & \underline{0.136}    & 0.154          & \underline{50.07}     & 53.62           & 0.478          & 0.481    \\
                         & Unet-S5P-1M                & \textbf{39.05} & \textbf{38.29} & \textbf{0.861} & \textbf{0.844} & \textbf{0.906} & \textbf{0.901} & \textbf{0.134} & \textbf{0.145} & \textbf{50.30}  & \textbf{57.10}  & \underline{0.481}    & \underline{0.489} \\ \hline
\multirow{8}{*}{BD4}  & Bicubic                    & 27.52          & 27.52          & 0.824          & 0.824          & 0.740          & 0.740          & 0.327          & 0.327          & 36.33           & 36.33           & 0.385          & 0.385          \\
                         & S5Net \cite{carbone2024efficient}                      & 27.46          & -              & 0.810          & -              & 0.747          & -              & 0.263          & -              & 36.90           & -               & \textbf{0.508} & -              \\
                         & DIP \cite{ulyanov2018deep}     & -  &  25.79 & -  & 0.779 & - & 0.609 & - & 0.312 & - & 32.17 & - & 0.428  \\
                         & ZSSR \cite{shocher2018zero}   & -  &  26.37  & -  & 0.798  & - & 0.699 & - & 0.334 &  -  & 32.12 & - & 0.326 \\                            
                         & S5-DSCR-S                  & 28.01          & 27.99          & 0.827          & 0.827          & 0.758          & 0.758          & 0.293          & 0.296          & 40.76           & 40.51           & 0.431          & 0.427          \\
                         & S5-DSCR                    & 28.41          & 28.17          & 0.849          & 0.842          & 0.778          & 0.774          & 0.257          & \underline{0.248}    & 43.06           & 47.10           & 0.466          & 0.485          \\
                         & Unet-S5P-800k              & \underline{28.60}    & \underline{28.25}    & \underline{0.854}    & \underline{0.845}    & \underline{0.790}    & \underline{0.782}    & \underline{0.232}    & 0.249          & \underline{44.70}     & \underline{50.76}     & 0.494          & \underline{0.499}    \\
                         & Unet-S5P-1M                & \textbf{28.63} & \textbf{28.32} & \textbf{0.854} & \textbf{0.847} & \textbf{0.791} & \textbf{0.785} & \textbf{0.230} & \textbf{0.241} & \textbf{44.77}  & \textbf{52.73}  & \underline{0.498}    & \textbf{0.503} \\ \hline
\multirow{8}{*}{BD5}  & Bicubic                    & 24.66          & 24.66          & 0.818          & 0.818          & 0.665          & 0.665          & 0.373          & 0.373          & 34.20           & 34.20           & 0.391          & 0.391          \\
                         & S5Net \cite{carbone2024efficient}                      & 24.76          & -              & 0.802          & -              & 0.677          & -              & 0.302          & -              & 35.65           & -               & 0.497          & -              \\
                         & DIP \cite{ulyanov2018deep}     & -  &  23.18  & -  & 0.773  & - & 0.526 & - & \textbf{0.277} &   -   &  30.35  & - & 0.446 \\
                         & ZSSR \cite{shocher2018zero}   & -  & 23.50 & -  & 0.790  & - & 0.605 & - & 0.394 & -  & 29.65  & - & 0.329 \\                            
                         & S5-DSCR-S                  & 25.13          & 25.09          & 0.824          & 0.825          & 0.688          & 0.684          & 0.333          & 0.331          & 39.16           & 38.17           & 0.439          & 0.428          \\
                         & S5-DSCR                    & 25.48          & 25.34          & 0.842          & 0.838          & 0.711          & 0.708          & 0.297          & 0.289  & 42.40           & 44.95           & 0.479          & 0.489          \\
                         & Unet-S5P-800k              & \underline{25.66}    & \underline{25.41}    & \underline{0.845}    & \underline{0.840}    & \underline{0.726}    & \underline{0.716}    & \underline{0.278}    & 0.291          & \textbf{44.79}  & \underline{48.17}     & \underline{0.506}    & \underline{0.508}    \\
                         & Unet-S5P-1M                & \textbf{25.66} & \textbf{25.48} & \textbf{0.846} & \textbf{0.842} & \textbf{0.728} & \textbf{0.721} & \textbf{0.273} & \underline{0.284} & \underline{44.38}     & \textbf{50.51}  & \textbf{0.513} & \textbf{0.509} \\ \hline
\multirow{8}{*}{BD6}  & Bicubic                    & 25.45          & 25.45          & 0.817          & 0.817          & 0.672          & 0.672          & 0.355          & 0.355          & 34.87           & 34.87           & 0.389          & 0.389          \\
                         & S5Net \cite{carbone2024efficient}                      & 25.47          & -              & 0.799          & -              & 0.682          & -              & 0.289          & -              & 35.97           & -               & 0.501          & -              \\
                         & DIP \cite{ulyanov2018deep}     & -  & 23.85  & -  & 0.696  & - & 0.482 & - & \textbf{0.241} &  -   & 34.51  & - & \textbf{0.641}  \\
                         & ZSSR \cite{shocher2018zero}   & -  & 24.29  & -  & 0.787  & - & 0.619 & - & 0.367 &    -  &  30.35  & - & 0.326 \\                            
                         & S5-DSCR-S                  & 25.92          & 25.86          & 0.823          & 0.824          & 0.695          & 0.691          & 0.319          & 0.316          & 39.58           & 38.47           & 0.436          & 0.424          \\
                         & S5-DSCR                    & 26.24          & 26.10          & 0.840          & 0.834          & 0.715          & 0.712          & 0.285          & 0.282          & 41.92           & 44.56           & 0.469          & 0.479          \\
                         & Unet-S5P-800k              & \underline{26.49}    & \underline{26.13}    & \underline{0.846}    & \underline{0.836}    & \underline{0.732}    & \underline{0.719}    & \underline{0.262}    & 0.277    & \underline{43.01}     & \underline{47.88}     & \underline{0.506}    & \underline{0.507} \\
                         & Unet-S5P-1M                & \textbf{26.50} & \textbf{26.19} & \textbf{0.846} & \textbf{0.837} & \textbf{0.734} & \textbf{0.721} & \textbf{0.257} & \underline{0.276} & \textbf{43.24}  & \textbf{49.12}  & \textbf{0.513} & 0.503    \\ \hline
\multirow{8}{*}{BD7}  & Bicubic                    & 24.48          & \underline{24.48}    & 0.738          & \textbf{0.738} & 0.565          & 0.565          & 0.405          & 0.405          & 32.79           & 32.79           & 0.330          & 0.330          \\
                         & S5Net \cite{carbone2024efficient}                      & -              & -              & -              & -              & -              & -              & -              & -              & -               & -               & -              & -              \\
                         & DIP \cite{ulyanov2018deep}     & -  & 23.38   & -  & 0.700  & - & 0.460 & - & 0.276 & - & 33.78 & - & \textbf{0.664}\\
                         & ZSSR \cite{shocher2018zero}   & -  & 23.91  & -  &  0.730 & - & 0.537 & - & 0.412 &   -  &  31.03  & - & 0.301 \\                             
                         & S5-DSCR-S                  & 24.73          & \textbf{24.63} & 0.734          & \underline{0.734}    & 0.576          & 0.574          & 0.389          & 0.385          & 33.94           & 33.58           & 0.367          & 0.364          \\
                         & S5-DSCR                    & 26.10          & 23.73          & \textbf{0.822} & 0.608          & 0.658          & 0.635          & 0.272          & \underline{0.274}         & 35.64           & 41.89           & 0.510          & 0.596          \\
                         & Unet-S5P-800k              & \underline{26.37}    & 24.14          & 0.810          & 0.688          & \textbf{0.697} & \underline{0.658}    & \textbf{0.224} & 0.286    & \textbf{35.94}  & \underline{45.76}     & \textbf{0.646} & 0.623    \\
                         & Unet-S5P-1M                & \textbf{26.41} & 23.86          & \underline{0.811}    & 0.671          & \underline{0.696}    & \textbf{0.662} & \underline{0.233}    & \textbf{0.270} & \underline{35.92}     & \textbf{47.59}  & \underline{0.622}    & \underline{0.654} \\ \hline
\multirow{8}{*}{BD8}  & Bicubic                    & 24.79          & \underline{24.79}    & 0.729          & \textbf{0.729} & 0.583          & 0.583          & 0.404          & 0.404          & 33.25           & 33.25           & 0.330          & 0.330          \\
                         & S5Net \cite{carbone2024efficient}                      & -              & -              & -              & -              & -              & -              & -              & -              & -               & -               & -              & -              \\
                         & DIP \cite{ulyanov2018deep}     & -  & 23.98 & -  & 0.698 & - & 0.494 & - & \textbf{0.232} &  -  & 34.51 & - &  0.627\\
                         & ZSSR \cite{shocher2018zero}   & -  & 24.14   & -  & 0.714  & - & 0.550 & - & 0.416 & -  & 31.40  & - & 0.296  \\                            
                         & S5-DSCR-S                  & 25.04          & \textbf{24.95} & 0.726          & \underline{0.724}    & 0.595          & 0.593          & 0.387          & 0.387          & 34.64           & 34.29           & 0.371          & 0.364          \\
                         & S5-DSCR                    & 26.42          & 24.04          & \textbf{0.831} & 0.607          & 0.675          & 0.652          & 0.267          & 0.272          & 36.25           & 42.14           & 0.509          & 0.592          \\
                         & Unet-S5P-800k              & \underline{26.64}    & 23.77          & 0.770          & 0.612          & \underline{0.711}    & \underline{0.673}    & \underline{0.230}    & 0.251 & \underline{36.46}     & \underline{46.16}     & \underline{0.600}    & \underline{0.655}    \\
                         & Unet-S5P-1M                & \textbf{26.69} & 23.69          & \underline{0.772}    & 0.612          & \textbf{0.714} & \textbf{0.680} & \textbf{0.226} & \underline{0.246} & \textbf{36.65}  & \textbf{47.99}  & \textbf{0.610} & \textbf{0.677} \\ \hline
\end{tabular}
}
\end{table*}

From an architectural perspective, performance depends on both model capacity and network design. S5-DSCR provides strong performance with 0.757M parameters, while its lightweight affine variant, S5-DSCR-S, exhibits noticeably lower performance due to its substantially reduced capacity. The Unet-S5P architectures achieve competitive performance, with Unet-S5P-1M providing the highest capacity among the evaluated models. These results indicate that reconstruction quality depends not only on parameter count but also on how DSC blocks, recursive refinement, and multi-scale pathways are organised.

\begin{figure}[pos=h,width=\linewidth]
    \centering
    \includegraphics[width=\linewidth]{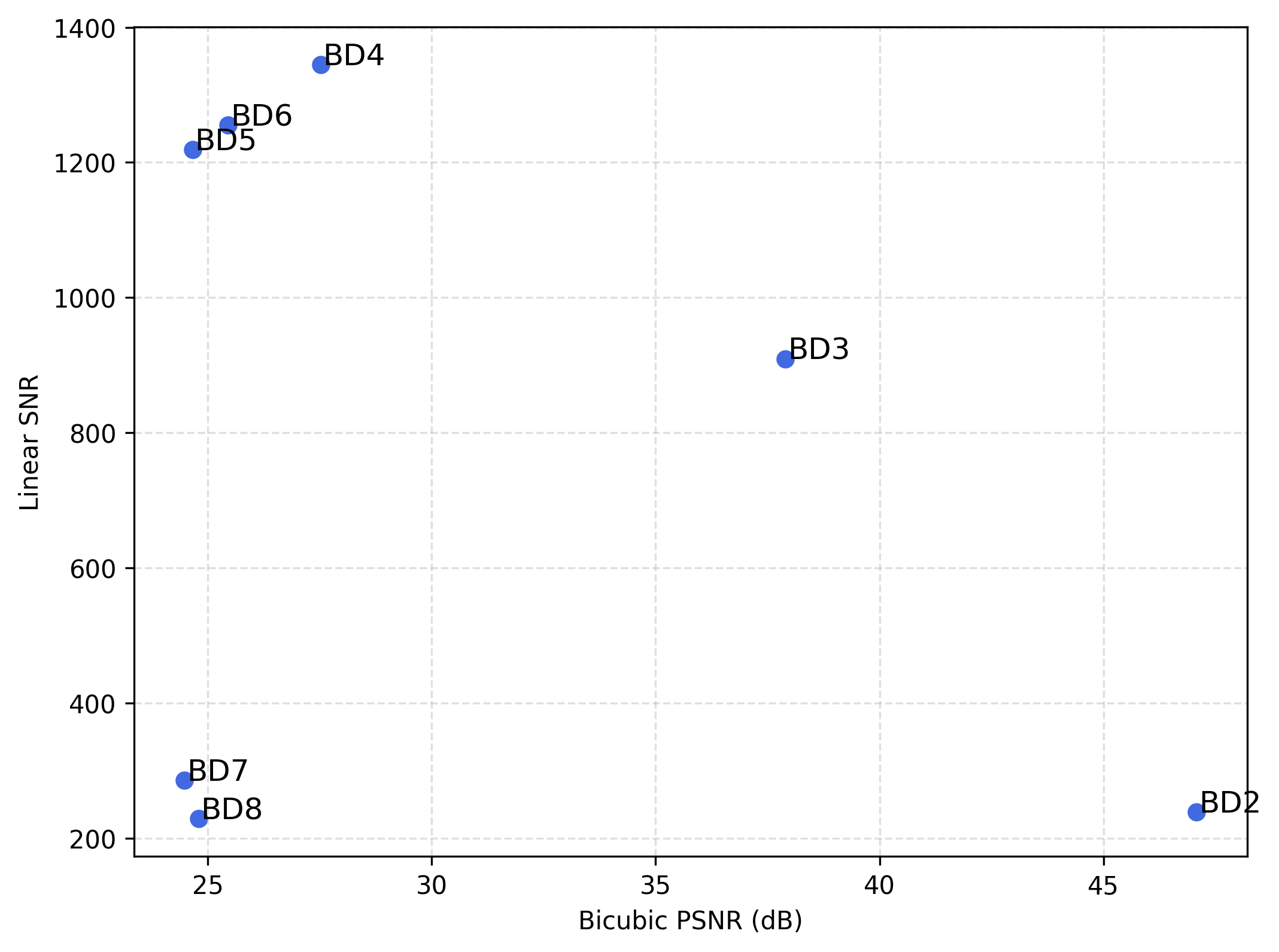}
    \caption{Relationship between SNR and bicubic PSNR across S5P spectral bands. Bicubic PSNR serves as an indicator of reconstruction difficulty. The trend highlights that performance is influenced by both signal quality and spatial characteristics}
    \label{psnr_fig}
\end{figure}

As shown in Table \ref{LRGT_table}, the proposed SL models generally achieve the highest reference-based performance, while the gap between SL and SSL remains moderate and band-dependent. This motivates a band-wise analysis in terms of SNR and bicubic PSNR, shown in Fig. \ref{psnr_fig}. Although higher SNR generally corresponds to better reconstruction performance, the relationship is not monotonic, indicating that SNR alone does not determine reconstruction difficulty. The spatial characteristics of each band therefore also play an important role.

In particular, BD2 represents a notable case, exhibiting one of the lowest SNR values while having the highest bicubic PSNR among the evaluated bands. It nevertheless achieves stable performance across the different models. This indicates that lower noise quality does not necessarily translate into greater SR difficulty when the underlying spatial structures are relatively favourable for reconstruction. Both SL and SSL models therefore maintain strong performance on BD2 with a relatively small gap between the two learning settings.

BD3 and BD4, corresponding to the UVIS spectrometer, represent favourable cases for SR. BD3 combines high bicubic PSNR with moderate SNR, while BD4 has the highest SNR among the evaluated bands. Both bands achieve strong reconstruction performance and exhibit relatively small differences between SL and SSL models. These results further indicate that signal quality contributes to reconstruction performance, although the behaviour of BD2 demonstrates that it is not the sole determining factor.

BD5 and BD6 provide an intermediate reconstruction scenario, combining relatively high SNR with lower bicubic PSNR than BD2-BD4. The two bands contain more challenging spatial structures while retaining relatively good signal quality. BD6 benefits from slightly higher SNR and bicubic PSNR than BD5 and consequently achieves slightly stronger performance across most metrics. Nevertheless, both bands remain effectively reconstructed under the SSL setting, with only a moderate gap relative to SL.

In contrast, BD7 and BD8, corresponding to the SWIR spectrometer, constitute the most challenging reconstruction scenario. They exhibit the lowest SNR and lowest bicubic PSNR among the bands, together with the largest gap between SL and SSL models. In several cases, the SSL models achieve lower PSNR than bicubic interpolation. However, they maintain competitive perceptual quality, measurement consistency, and sharpness. This indicates that pixel-wise reconstruction accuracy becomes increasingly difficult under the degraded SWIR observations, while the reconstructed images can still remain consistent with the S5P measurement process and preserve spatial detail.

To better understand the origin of the reduced performance observed for BD7 and BD8, a series of diagnostic experiments was conducted using BD5 as a reference band. First, to evaluate the sensitivity of the proposed framework to degradation-model mismatch, BD5 LR observations were generated using the degradation operator of BD7 while keeping all other settings unchanged. Second, the influence of measurement noise was investigated by degrading BD5 to match the SNR of BD7 before training and evaluation. Finally, the effect of the training configuration was examined by reducing the patch size to $52\times52$, matching the spatial dimensions used for BD7 and BD8. None of these experiments reproduced the degradation observed for the SWIR bands, indicating that the reduced performance cannot be attributed solely to the degradation model mismatch, increased noise level, or the adopted training configuration. Instead, the results suggest that the SWIR bands combine lower signal quality with distinct spatial imaging characteristics that make high-frequency reconstruction more challenging.

To further verify that the SSL formulation itself does not cause this behaviour, the ablation study presented in Section \ref{section_sure} evaluates the contribution of the denoising term. Removing the denoising component does not produce a consistent improvement for BD7 and BD8, indicating that the denoising objective does not limit the observed performance. Together with the diagnostic experiments above, these results support the interpretation that the reconstruction difficulty of BD7 and BD8 is primarily associated with their intrinsic imaging characteristics rather than with a specific modelling or training choice.

Overall, the results show that reconstruction performance is governed by the joint effect of signal quality and spatial image characteristics rather than by SNR alone. While SL provides a useful reference through access to GT during training, the proposed SSL framework achieves competitive reconstruction quality across most spectral bands and remains consistent with the S5P measurement model even under the challenging imaging conditions of the SWIR spectrometer.

\subsection{GT-SHR (Self-Supervised Evaluation)}
We next evaluate the core objective of the proposed framework: super-resolving native S5P observations beyond the sensor's native spatial resolution without access to HR GT. Since no HR reference exists for these observations, the evaluation relies on the no-reference metrics described in Section \ref{section_M}, together with qualitative comparisons.

\begin{figure*}[pos=h]
\centering
\setlength{\tabcolsep}{3pt} 
\renewcommand{\arraystretch}{1.0} 
\begin{tabular}{cccccc}
    & \textbf{GT} & \textbf{Bicubic} & \textbf{DIP} & \textbf{ZSSR} & \textbf{Ours} \\[1mm]

    \raisebox{15mm}{\rotatebox{90}{\textbf{BD2}}}&
    \includegraphics[width=0.20\linewidth]{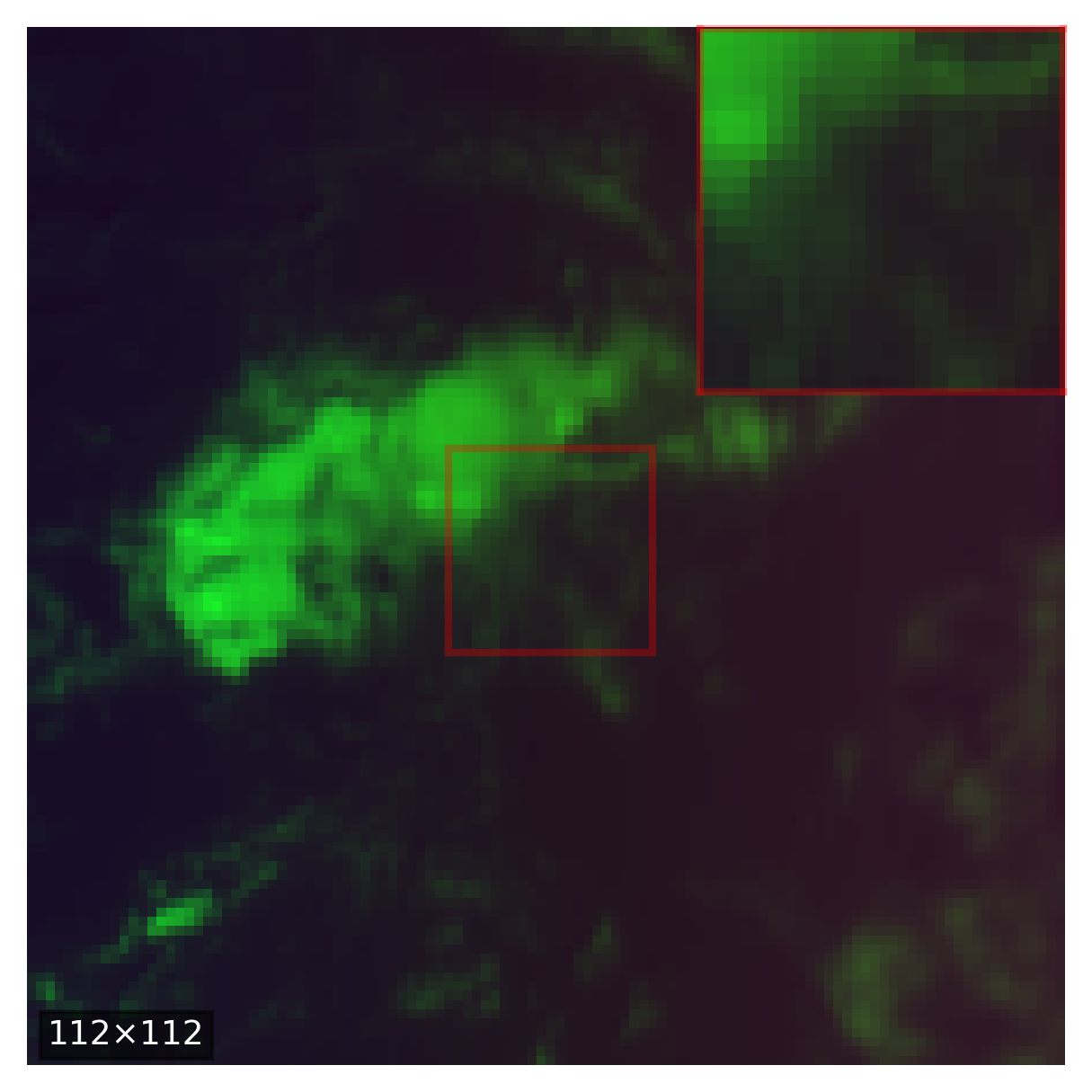} &
    \includegraphics[width=0.20\linewidth]{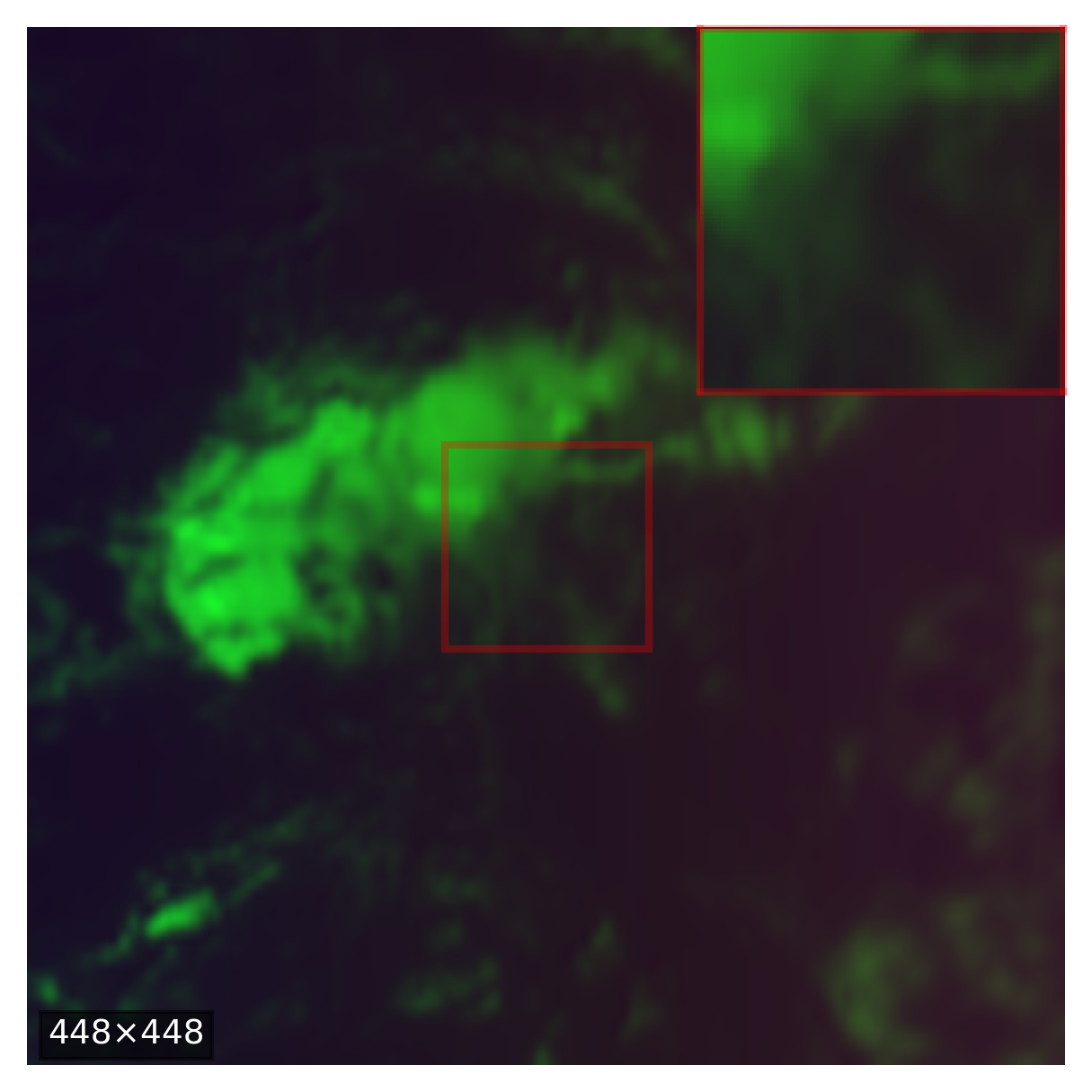} &
    \includegraphics[width=0.20\linewidth]{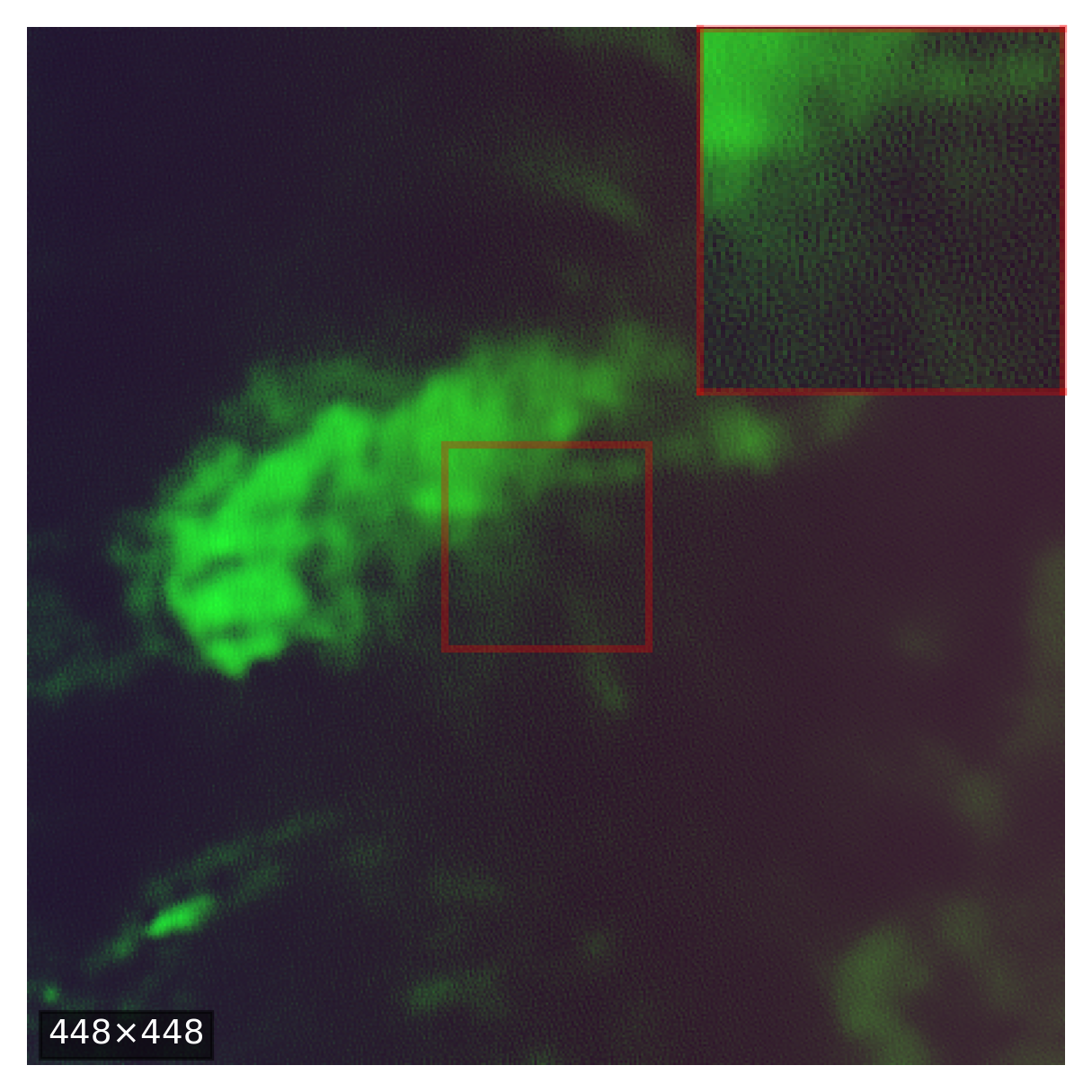} &
    \includegraphics[width=0.20\linewidth]{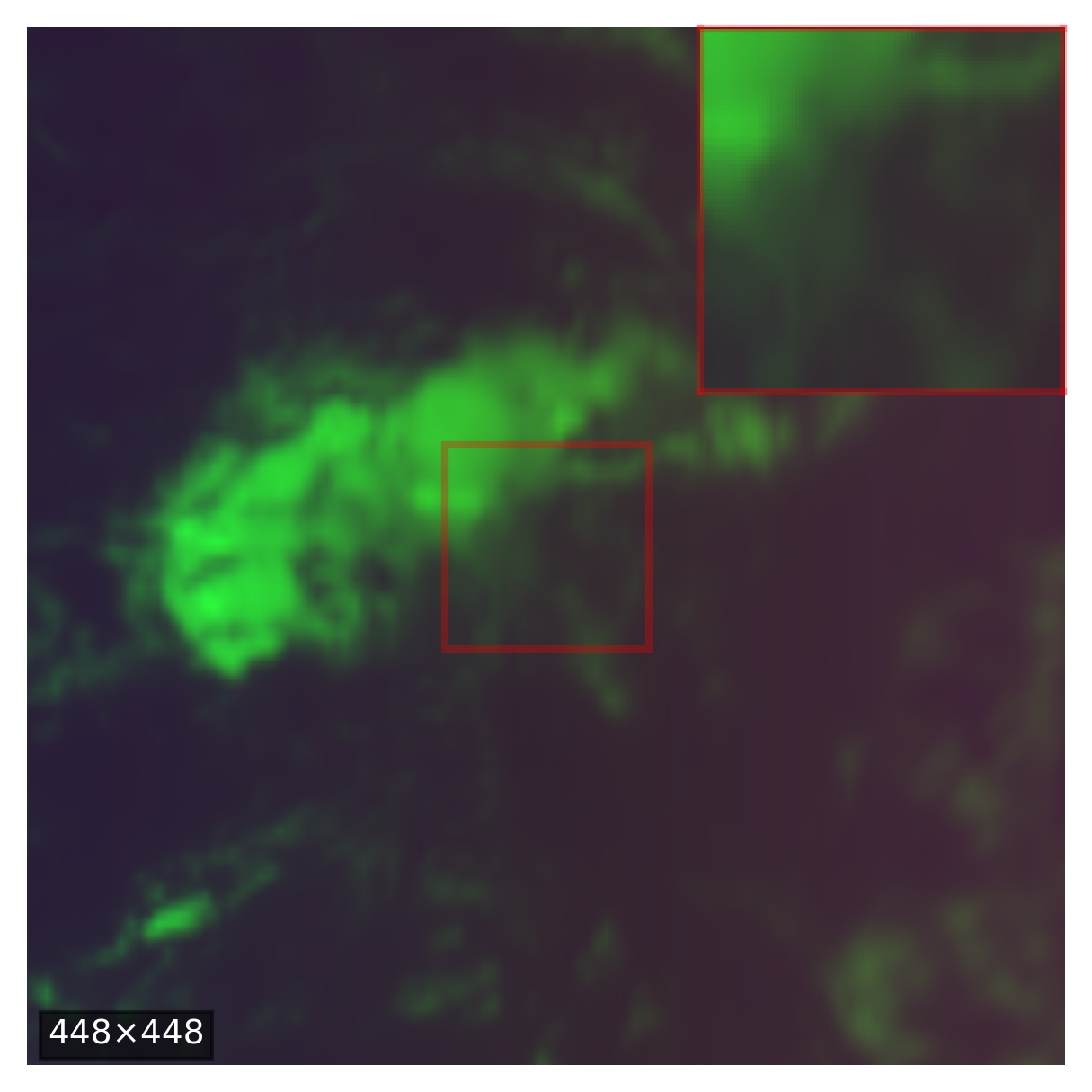} &
    \includegraphics[width=0.20\linewidth]{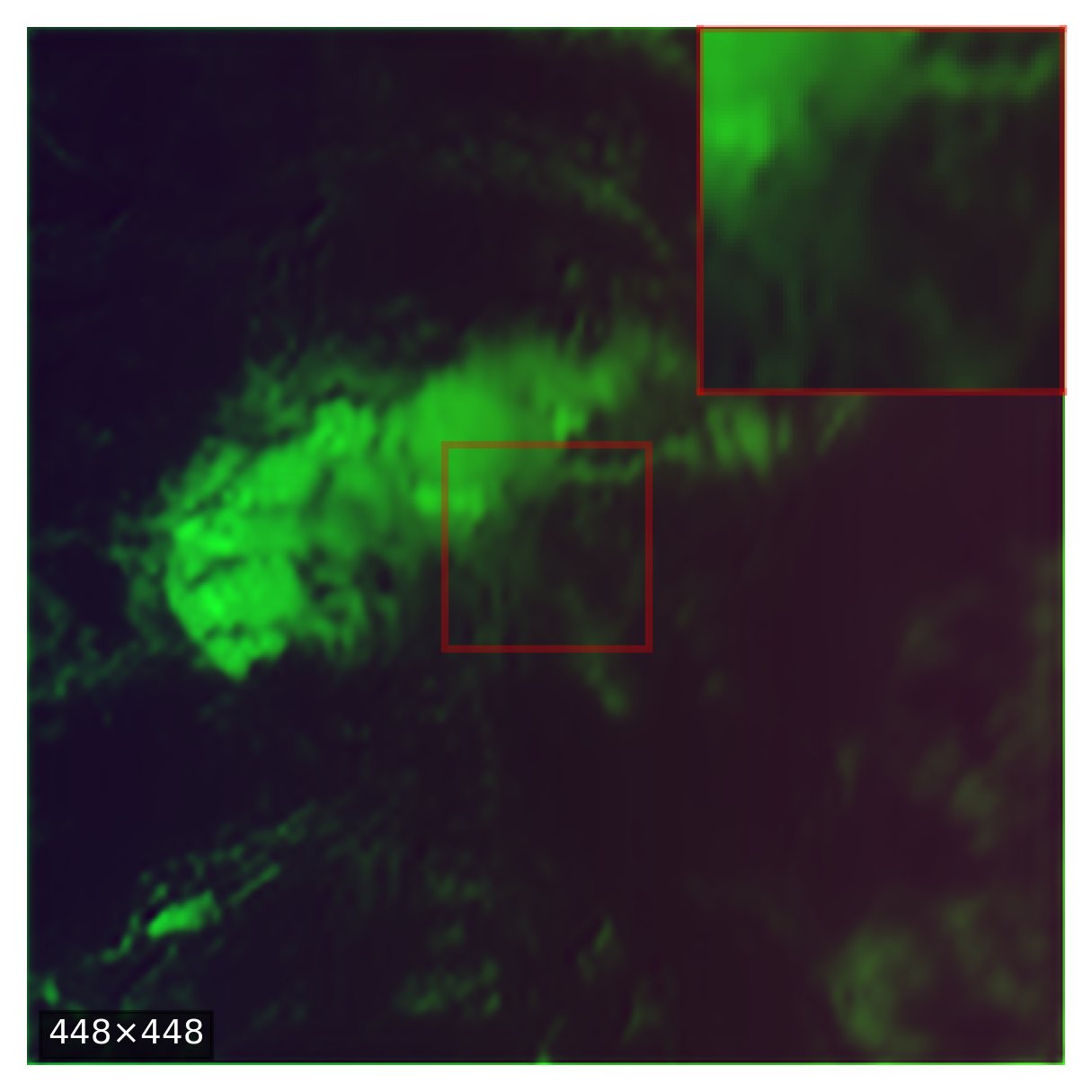}  \\[1mm]

    \raisebox{15mm}{\rotatebox{90}{\textbf{BD5}}} &
    \includegraphics[width=0.20\linewidth]{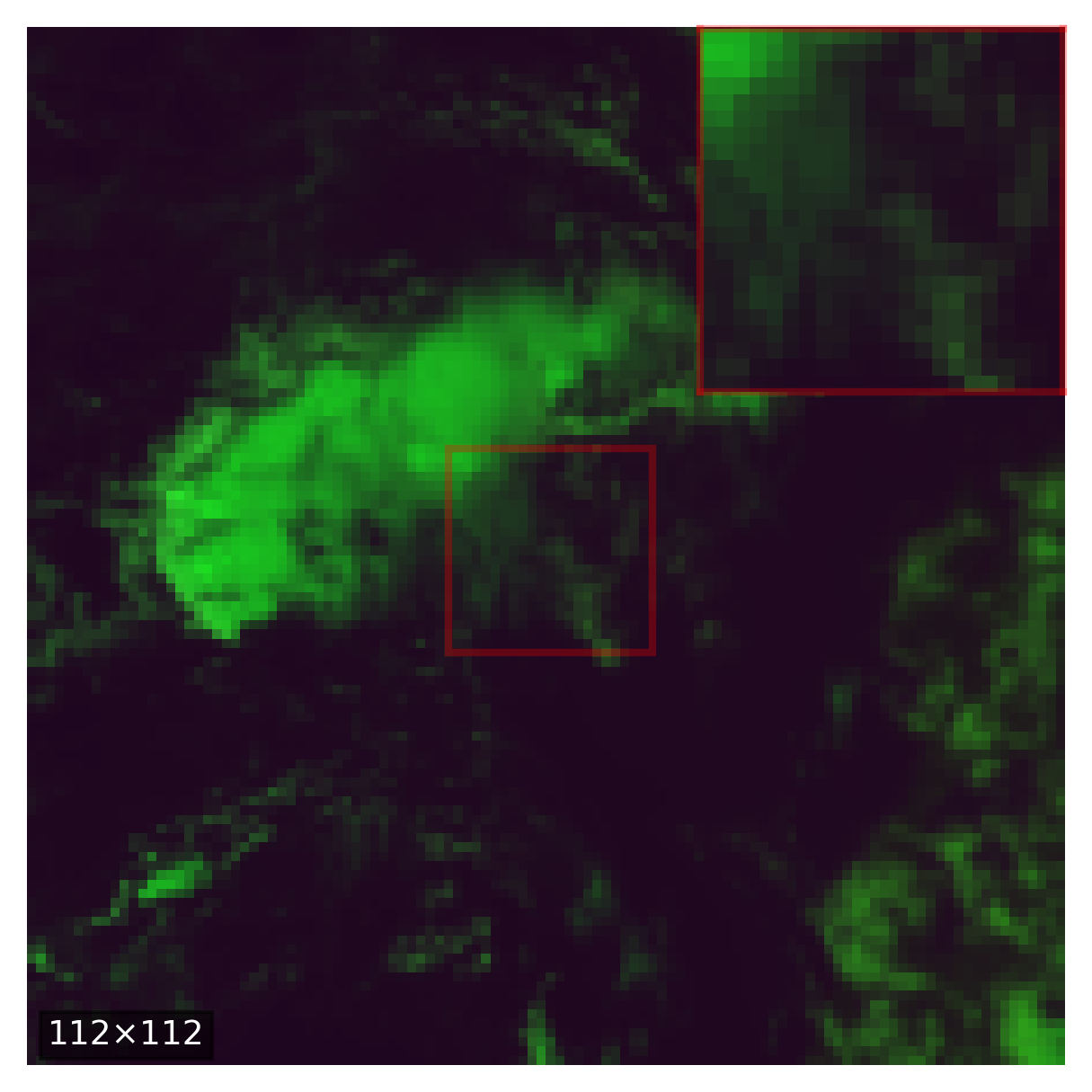} &
    \includegraphics[width=0.20\linewidth]{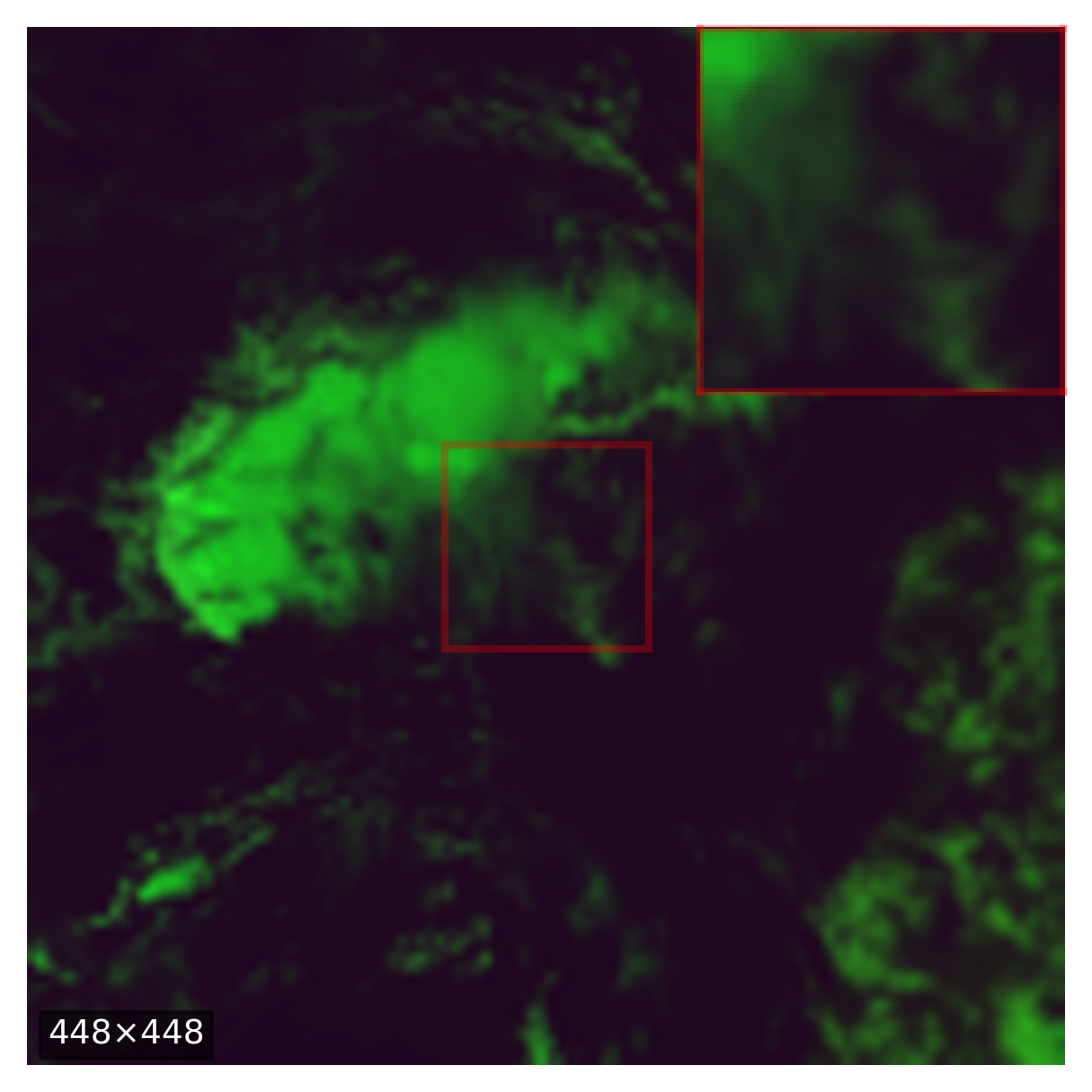} &
    \includegraphics[width=0.20\linewidth]{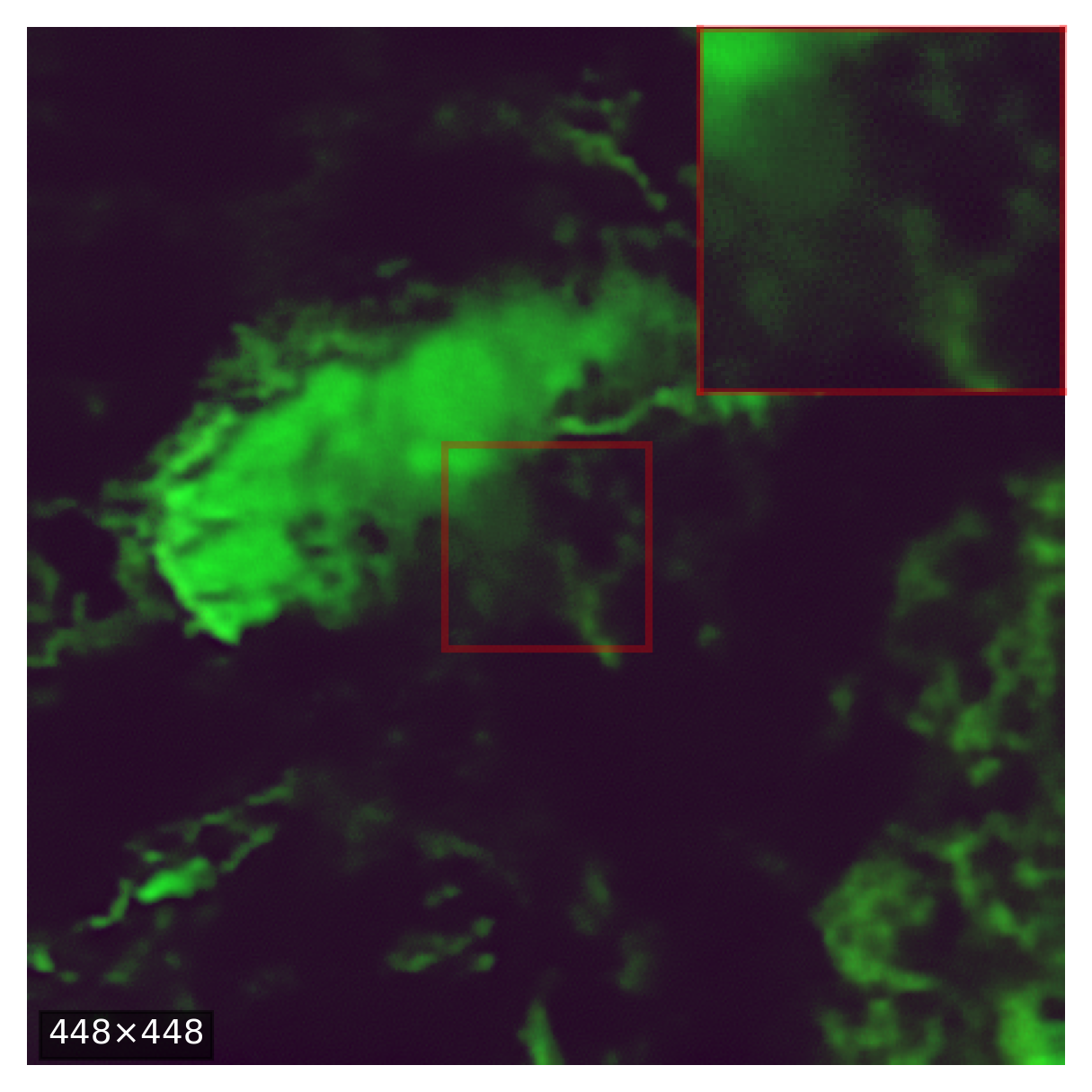} &
    \includegraphics[width=0.20\linewidth]{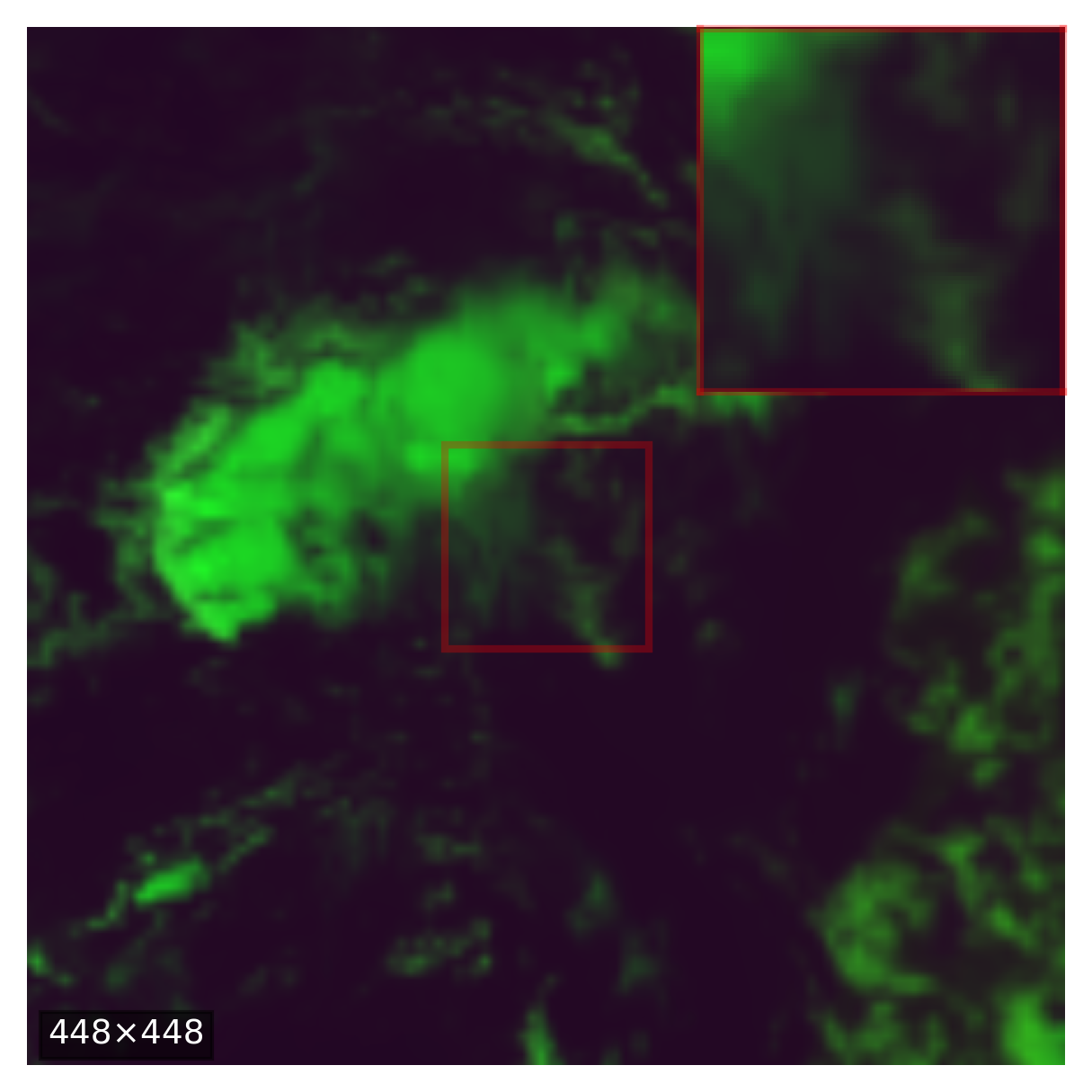} &
    \includegraphics[width=0.20\linewidth]{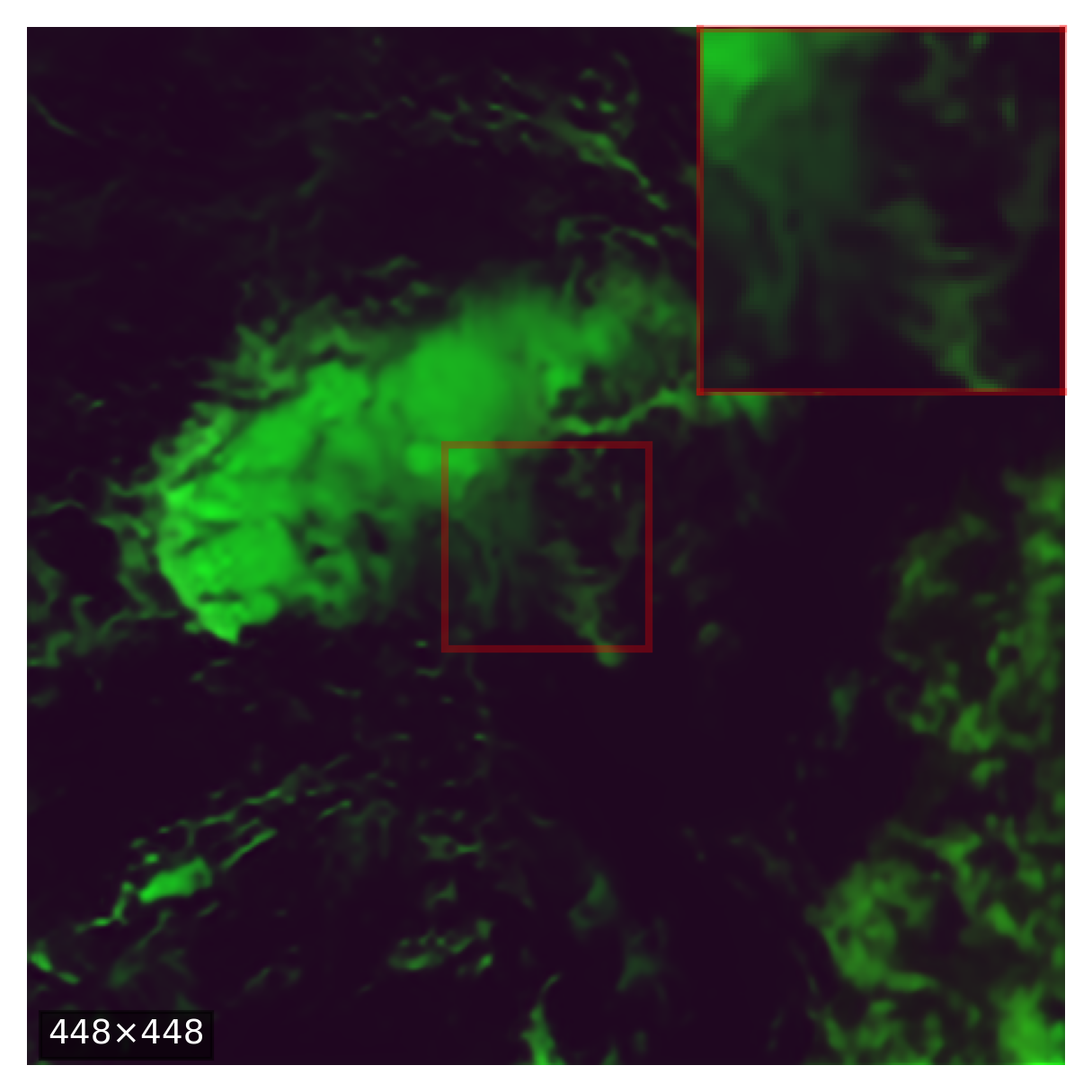} \\[1mm]

    \raisebox{15mm}{\rotatebox{90}{\textbf{BD4}}} &
    \includegraphics[width=0.20\linewidth]{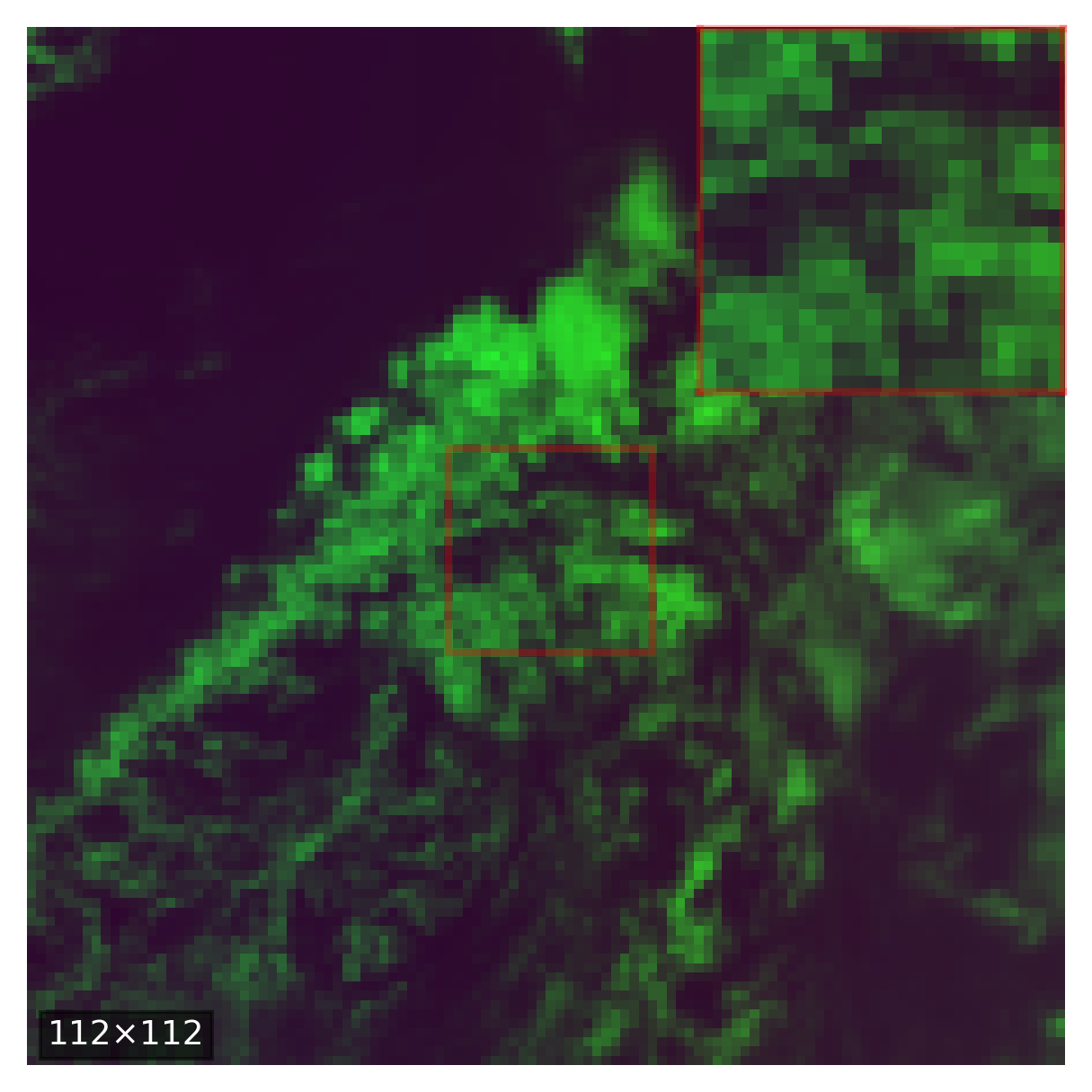} &
    \includegraphics[width=0.20\linewidth]{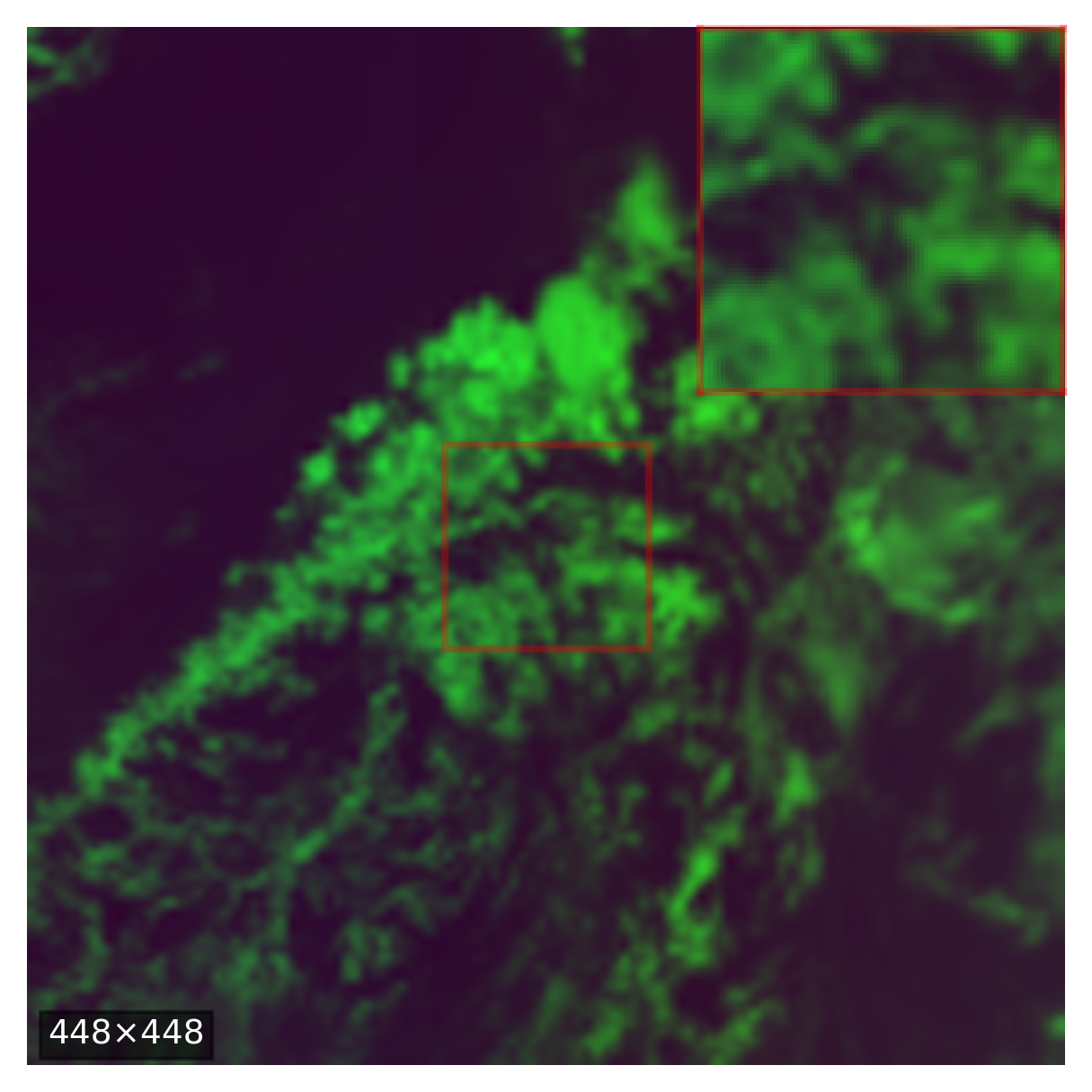} &
    \includegraphics[width=0.20\linewidth]{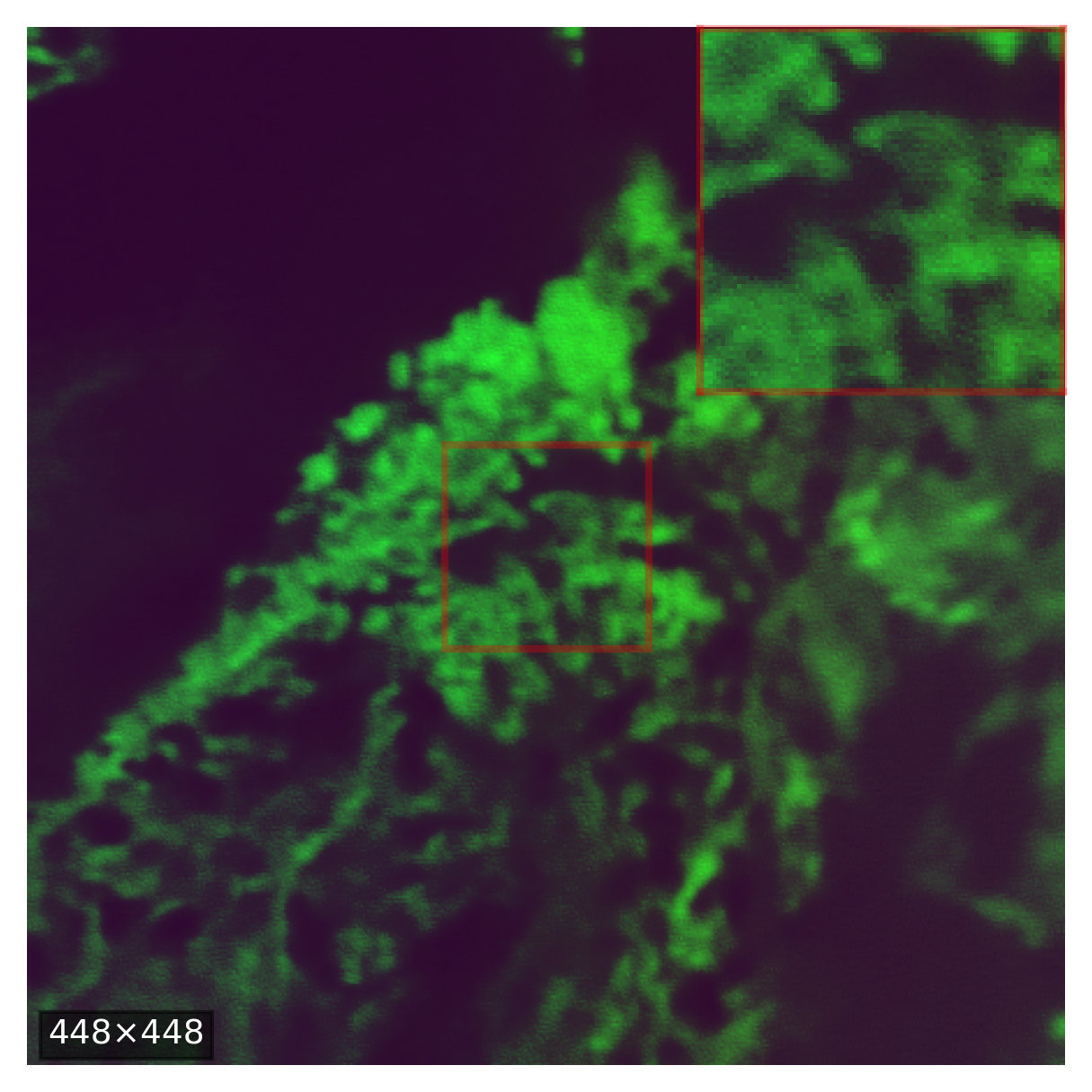} &
    \includegraphics[width=0.20\linewidth]{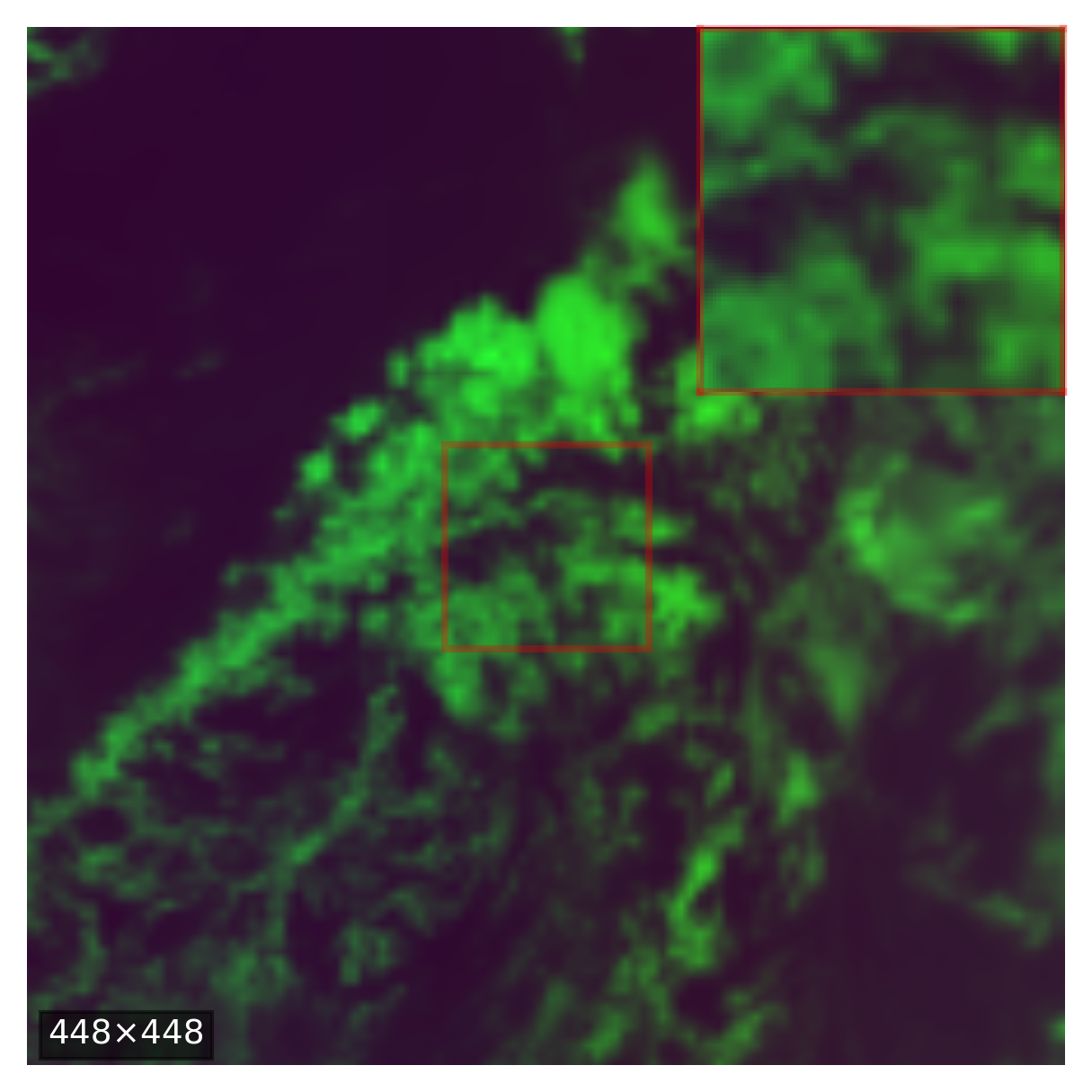} &
    \includegraphics[width=0.20\linewidth]{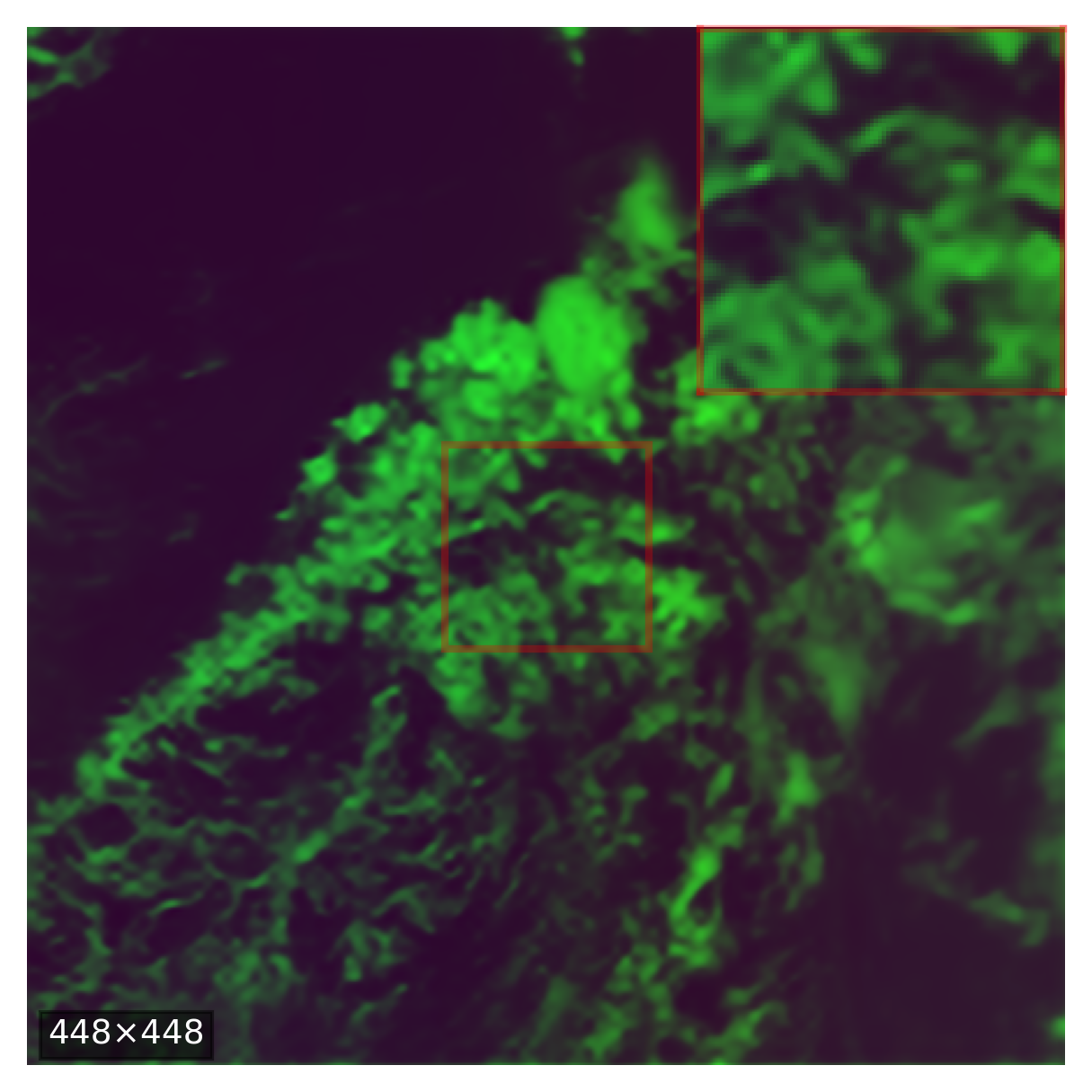}  \\[1mm]

    \raisebox{15mm}{\rotatebox{90}{\textbf{BD7}}} &
    \includegraphics[width=0.20\linewidth]{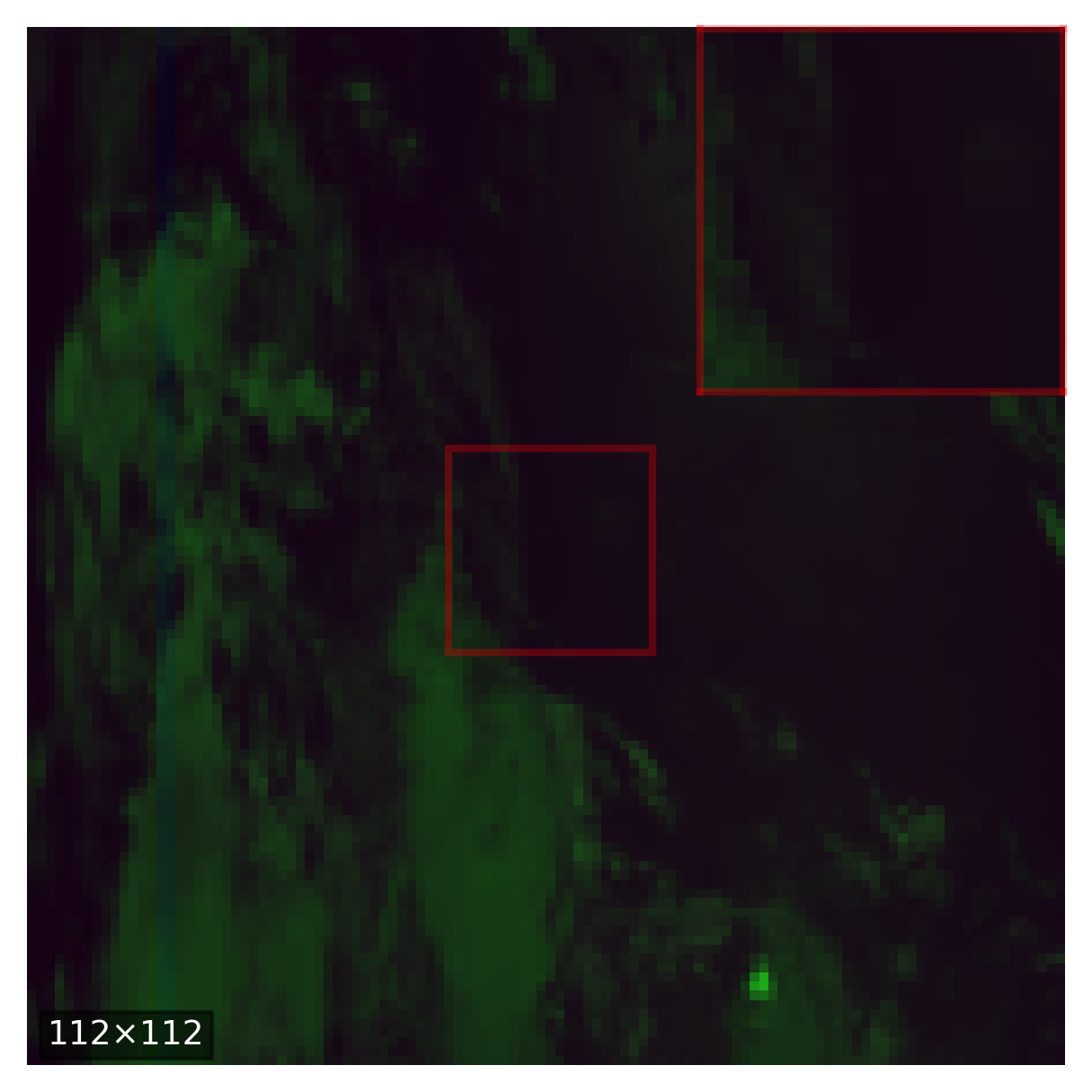} &
    \includegraphics[width=0.20\linewidth]{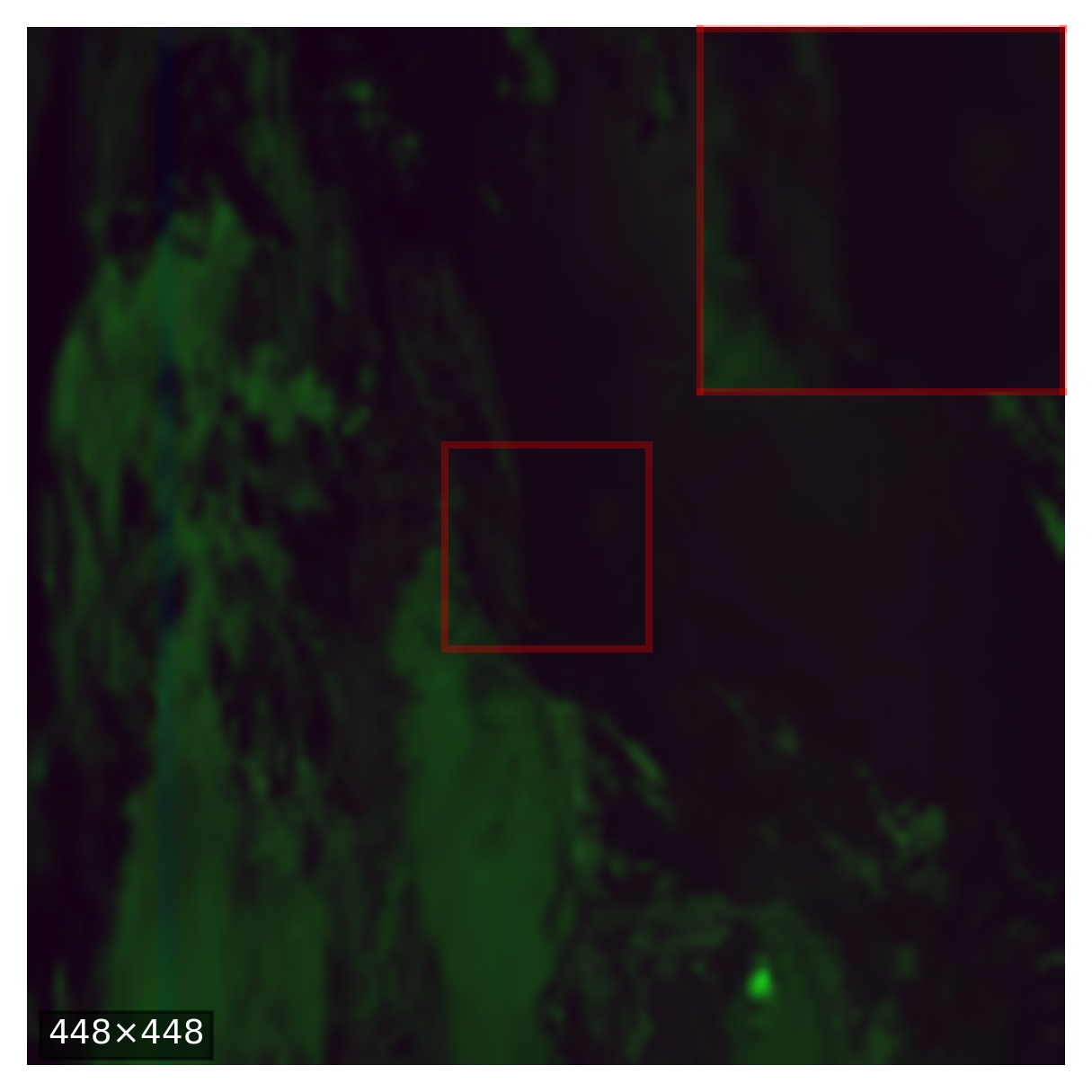} &
    \includegraphics[width=0.20\linewidth]{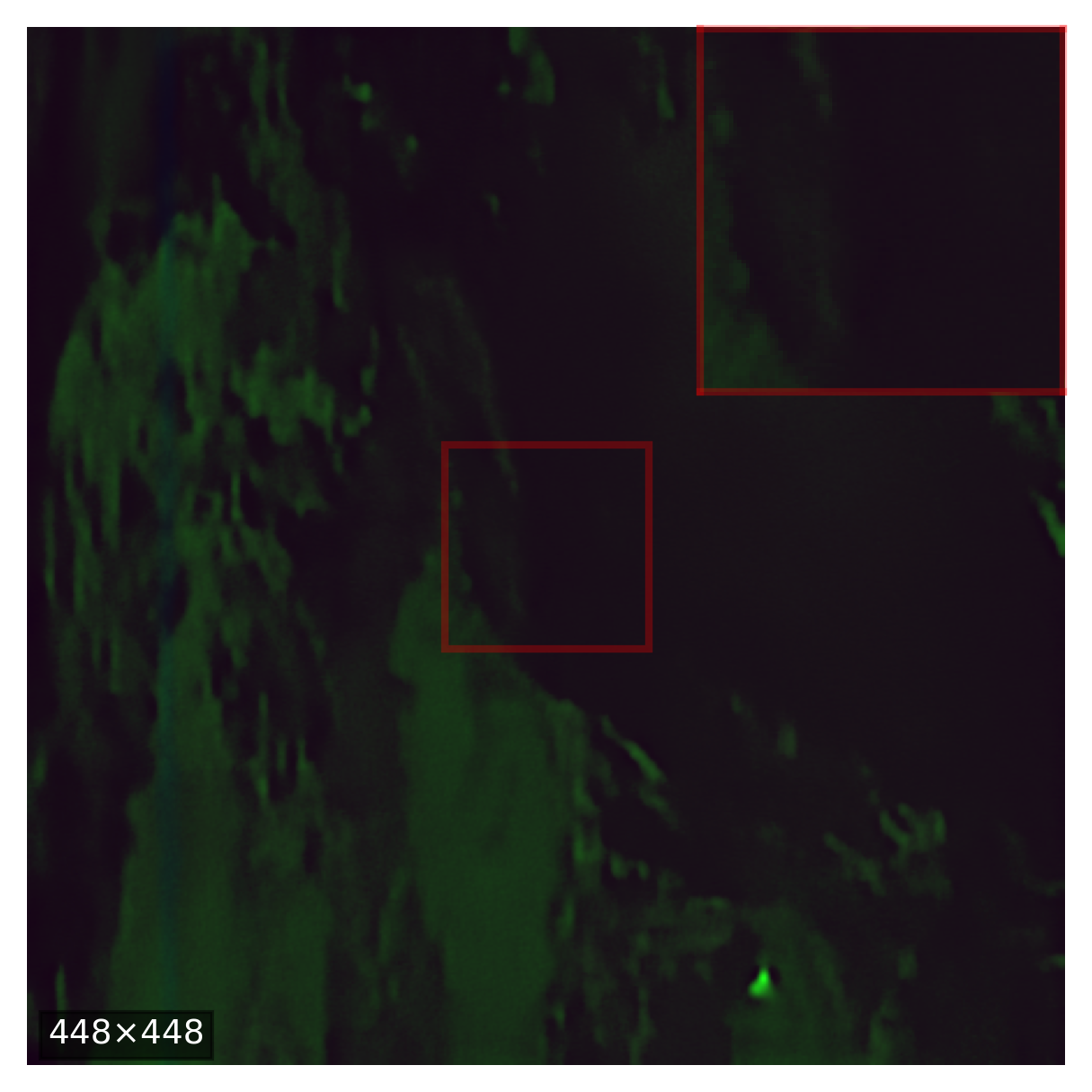} &
    \includegraphics[width=0.20\linewidth]{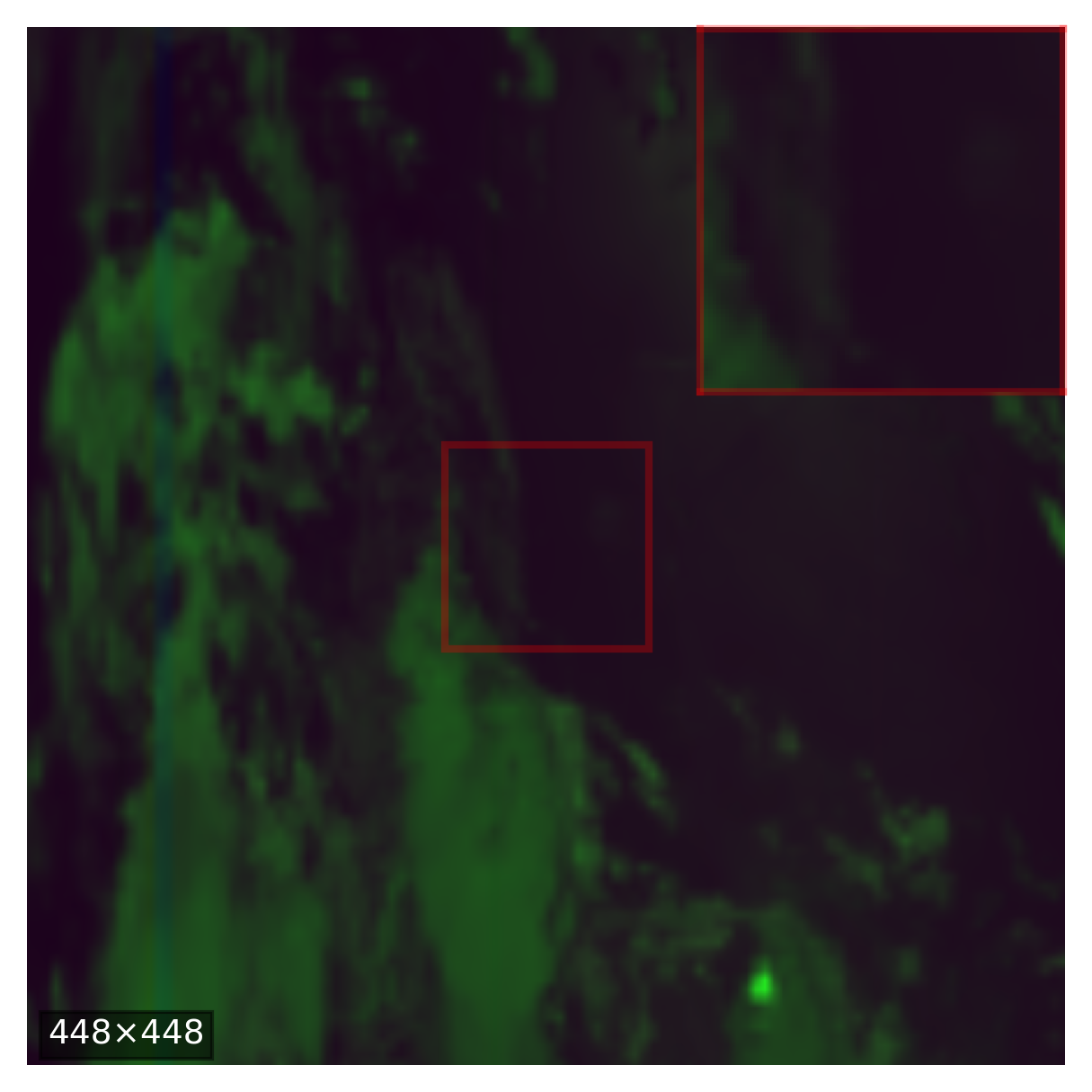} &
    \includegraphics[width=0.20\linewidth]{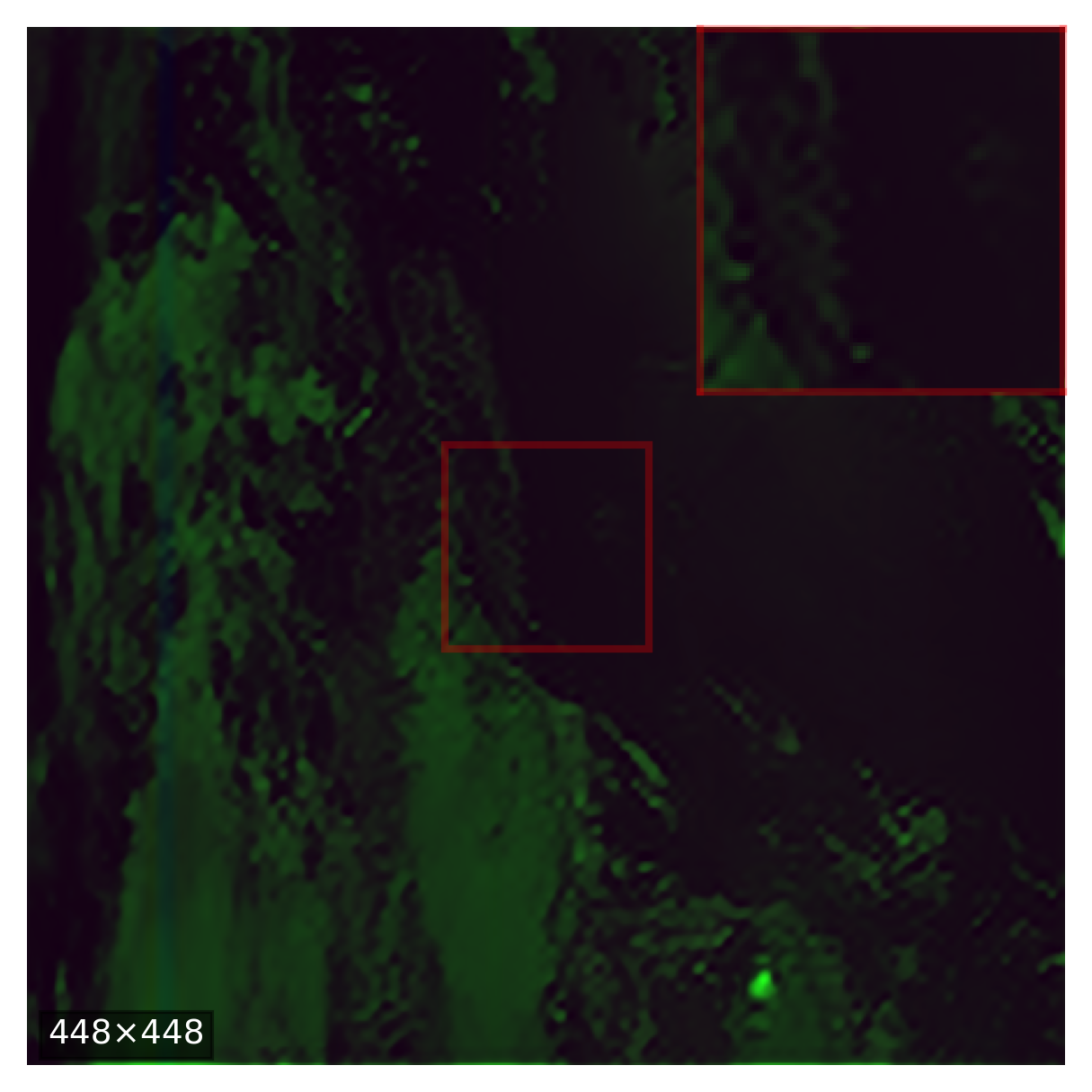}  \\
\end{tabular}

\caption{GT-SHR qualitative evaluation through Unet-S5P-1M. Each spectral band is trained independently. BD2 and BD5 show the same spatial region, illustrating the spatial stability of reconstructed details across bands}
\label{GT_SHR_fig}
\end{figure*}

For visualisation, Principal Component Analysis (PCA) is applied to the HS images, and the first three principal components are used to form an RGB representation. To ensure a consistent and fair visual comparison, the components are determined from the GT image and the same projection is then applied to all corresponding reconstructions. Fig. \ref{GT_SHR_fig} presents representative results for four spectral bands, comparing the original S5P input, bicubic interpolation, DIP, ZSSR, and the proposed SHR reconstruction. Visually, the proposed SHR outputs recover finer spatial details and sharper boundaries than bicubic interpolation. DIP and ZSSR generally produce smoother reconstructions with less spatial enhancement. The proposed models provide enhanced spatial detail while preserving the main structures present in the input observations. 

To further assess the GT-SHR reconstruction quality, the proposed SSL models are compared with DIP and ZSSR using the measurement consistency and sharpness metrics reported in Table \ref{GT_SHR_table}. The measurement consistency results show trends consistent with the LR-HR evaluation in Table \ref{LRGT_table}. The Unet-S5P variants achieve the highest measurement consistency across the evaluated bands. Their consistently high measurement consistency across both settings supports the stability of the proposed reconstruction framework. Sharpness also shows strong performance of the Unet-S5P variants, with the proposed models achieving the highest values for several bands, consistent with the trends observed in the LR-HR results in Table \ref{LRGT_table}. As discussed previously, higher sharpness does not necessarily indicate higher reconstruction fidelity. Notably, for the challenging SWIR bands, Unet-S5P-1M achieves the highest measurement consistency for both BD7 and BD8, together with high sharpness values, ranking among the leading models for both bands.

\begin{table}[!htp]
\centering
\caption{GT-SHR quantitative comparison across all the spectral bands. For each band and metric, the best results are shown in bold and the second-best are underlined}
\label{GT_SHR_table}
\resizebox{0.90\linewidth}{!}{%
\begin{tabular}{@{}cc|c|c@{}}
\hline
\textbf{Band} & \textbf{Model} & \textbf{Consistency} $\uparrow$ & \textbf{Sharpness} $\uparrow$ \\
\hline

\multirow{7}{*}{BD2} 
& Bicubic        & 37.80 & 0.374 \\
& DIP \cite{ulyanov2018deep}  & 49.25  & \textbf{0.943}  \\
& ZSSR \cite{shocher2018zero}  & 50.35 & 0.387  \\
& S5-DSCR-S      & 40.21 & 0.403 \\
& S5-DSCR        & 46.82 & 0.463 \\
& Unet-S5P-800k  & \underline{56.62} & 0.498 \\
& Unet-S5P-1M    & \textbf{56.99} & \underline{0.498} \\
\hline

\multirow{7}{*}{BD3} 
& Bicubic   & 46.06 & 0.390 \\
& DIP \cite{ulyanov2018deep}  & 42.63  & \textbf{0.591} \\
& ZSSR \cite{shocher2018zero}  & 43.51  & 0.345 \\
& S5-DSCR-S  & 47.53 & 0.423 \\
& S5-DSCR  & 53.19 & 0.528 \\
& Unet-S5P-800k  & \underline{61.02} & 0.524 \\
& Unet-S5P-1M & \textbf{63.88} & \underline{0.532} \\
\hline

\multirow{7}{*}{BD4} 
& Bicubic        & 37.78 & 0.374 \\
& DIP \cite{ulyanov2018deep}  & 34.65  & \textbf{0.514} \\
& ZSSR \cite{shocher2018zero}  & 34.27 & 0.324 \\
& S5-DSCR-S      & 40.21 & 0.403 \\
& S5-DSCR        & 46.82 & 0.463 \\
& Unet-S5P-800k  & \underline{56.62} & 0.498 \\
& Unet-S5P-1M    & \textbf{56.99} & \underline{0.498} \\
\hline

\multirow{7}{*}{BD5} 
& Bicubic   & 35.45 & 0.385 \\
& DIP \cite{ulyanov2018deep}  &  32.14 & 0.514 \\
& ZSSR \cite{shocher2018zero}  & 31.49 & 0.327 \\
& S5-DSCR-S  & 38.98  & 0.421 \\
& S5-DSCR  & 45.56 & 0.474 \\
& Unet-S5P-800k  & \underline{52.96} & \textbf{0.520} \\
& Unet-S5P-1M & \textbf{53.43} & \underline{0.517} \\  
\hline

\multirow{7}{*}{BD6} 
& Bicubic   & 36.11 & 0.387 \\
& DIP \cite{ulyanov2018deep}  & 34.68  &  0.521 \\
& ZSSR \cite{shocher2018zero}  & 32.23 & 0.330 \\
& S5-DSCR-S  & 39.50 & 0.425 \\
& S5-DSCR  & 45.28 & 0.494 \\
& Unet-S5P-800k  & \underline{49.70} & \underline{0.561} \\
& Unet-S5P-1M & \textbf{51.46} & \textbf{0.573} \\
\hline

\multirow{7}{*}{BD7} 
& Bicubic   & 39.85 & 0.394 \\
& DIP \cite{ulyanov2018deep}  &  40.45  & \underline{1.080}  \\
& ZSSR \cite{shocher2018zero}  & 37.86 & 0.340  \\
& S5-DSCR-S  & 40.95 & 0.444 \\
& S5-DSCR  & 47.35 & 0.970 \\
& Unet-S5P-800k  & \underline{50.71} & 1.059 \\
& Unet-S5P-1M & \textbf{52.57} &  \textbf{1.181} \\
\hline

\multirow{7}{*}{BD8} 
& Bicubic   & 42.89 & 0.402 \\
& DIP \cite{ulyanov2018deep}  &  40.50 & \underline{1.043} \\
& ZSSR \cite{shocher2018zero}  & 40.87 & 0.350 \\
& S5-DSCR-S  & 44.43 &  0.463 \\
& S5-DSCR  & 50.09 &  0.862 \\
& Unet-S5P-800k  & \underline{52.58} & 1.026 \\
& Unet-S5P-1M & \textbf{53.95} & \textbf{1.162} \\
\hline
\end{tabular}}
\end{table}

Beyond the comparison with generic self-supervised baselines, we examine the spatial stability of reconstructed details across spectral bands. Although each band captures a different spectral range and has distinct radiometric responses, the observations correspond to the same scene and acquisition period. Consequently, spatial structures may exhibit different levels of visibility across bands. Since each band is trained independently, correspondence between reconstructed structures across bands provides complementary evidence of spatial stability. However, such correspondence cannot establish the physical authenticity of a recovered structure, since a feature prominent in one band may be weak or absent in another because of the different spectral responses. This behaviour is illustrated in Fig. \ref{GT_SHR_fig}, where the same region is shown for BD2 and BD5. The bicubic BD5 image exhibits a distinct "C"-like pattern composed of five small round blobs, in the top-right of the central region, while the corresponding structure is barely visible in the bicubic BD2 image. After applying the proposed SR model, the pattern becomes more apparent in the SHR of BD2, indicating spatial correspondence with the structure observed in BD5. This correspondence provides complementary evidence of spatial stability, while the observed cross-band agreement supports the consistency of the reconstructed BD2 structure with spatial features visible in BD5.

The GT-SHR comparison also highlights a substantial difference in reconstruction time. As reported in Table \ref{tab:efficiency}, the image-specific optimisation required by DIP and ZSSR results in considerably longer reconstruction times than the proposed models. This difference is particularly relevant for large-scale processing of native S5P observations, where image-specific optimisation would impose higher computational costs.

\subsection{Uncertainty Analysis}
To evaluate the stability of the proposed SSL framework, we trained five instances of Unet-S5P-1M using different random initialisations while keeping the training, validation, and testing datasets identical. Predictive uncertainty was estimated as the pixel-wise standard deviation across the ensemble predictions, while the ensemble mean was used as the final reconstruction. This experiment was conducted on the four representative bands (BD2, BD4, BD5, and BD7).

For qualitative visualisation, a representative test image was selected for each band. Rather than selecting random images, the test image was selected as the image whose mean uncertainty was closest to the average uncertainty computed over the entire test set. This selection strategy provides a representative visual example while avoiding bias toward either exceptionally easy or difficult samples.

\begin{figure*}[pos=h]
\centering
\setlength{\tabcolsep}{3pt}
\renewcommand{\arraystretch}{1.0}

\begin{tabular}{ccccc}
    & \textbf{Input GT} &
    \textbf{Ensemble Mean} &
    \textbf{Uncertainty} &
    \\[1mm]

    \raisebox{15mm}{\rotatebox{90}{\textbf{BD2}}} &
    \includegraphics[
        width=0.24\linewidth,
        height=0.24\linewidth,
        keepaspectratio
    ]{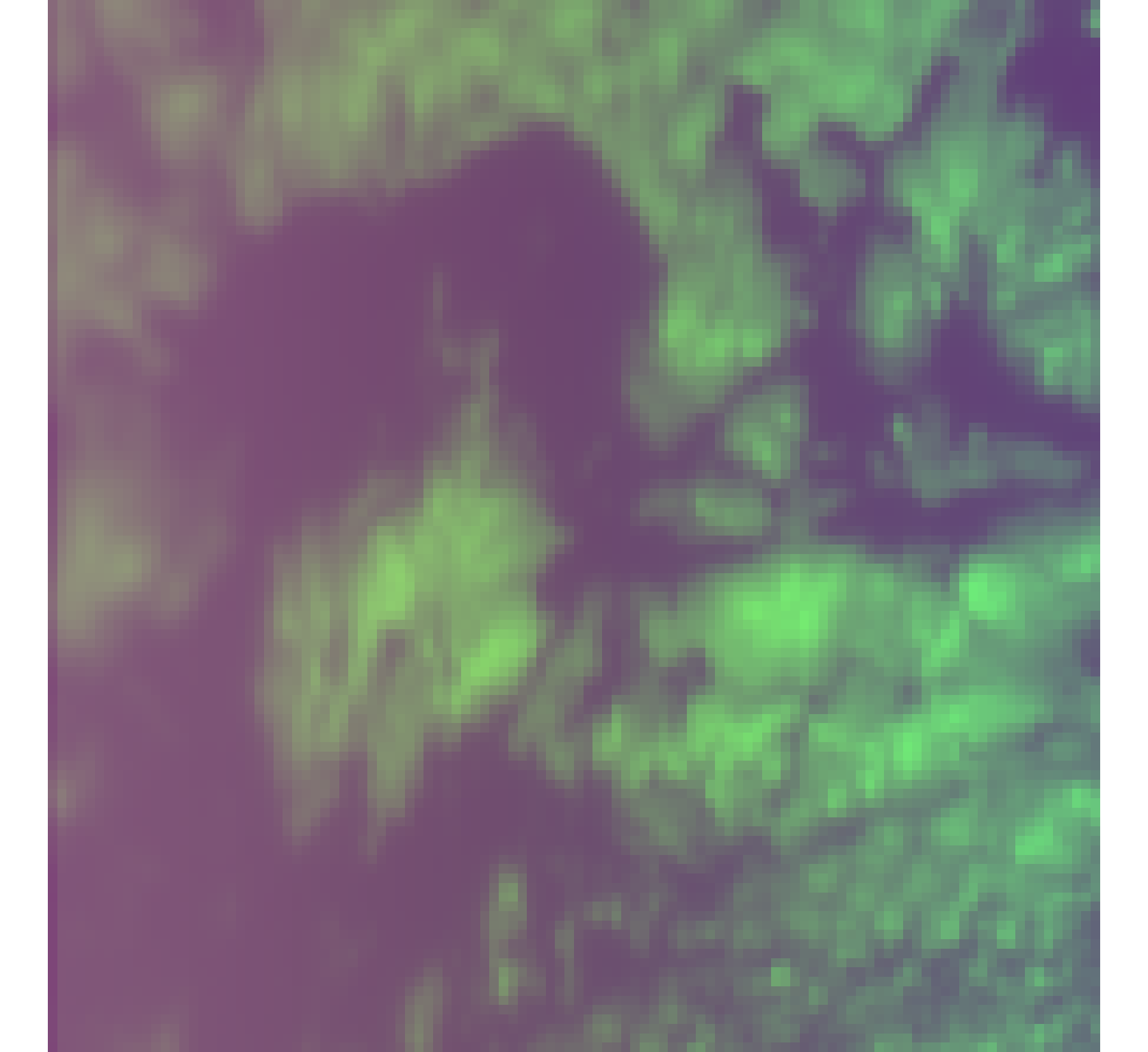} &
    \includegraphics[
        width=0.24\linewidth,
        height=0.24\linewidth,
        keepaspectratio
    ]{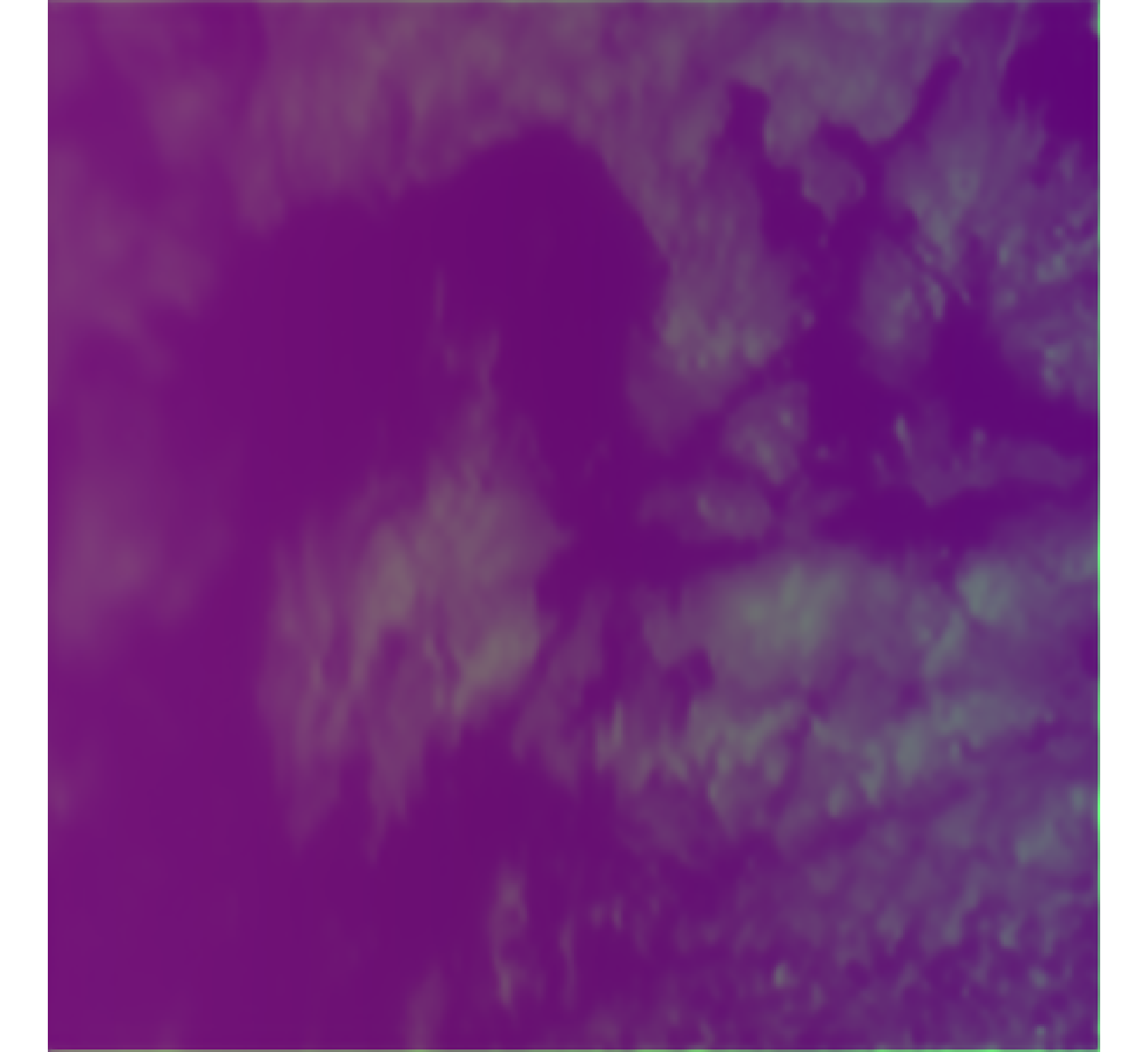} &
    \includegraphics[
        width=0.24\linewidth,
        height=0.24\linewidth,
        keepaspectratio
    ]{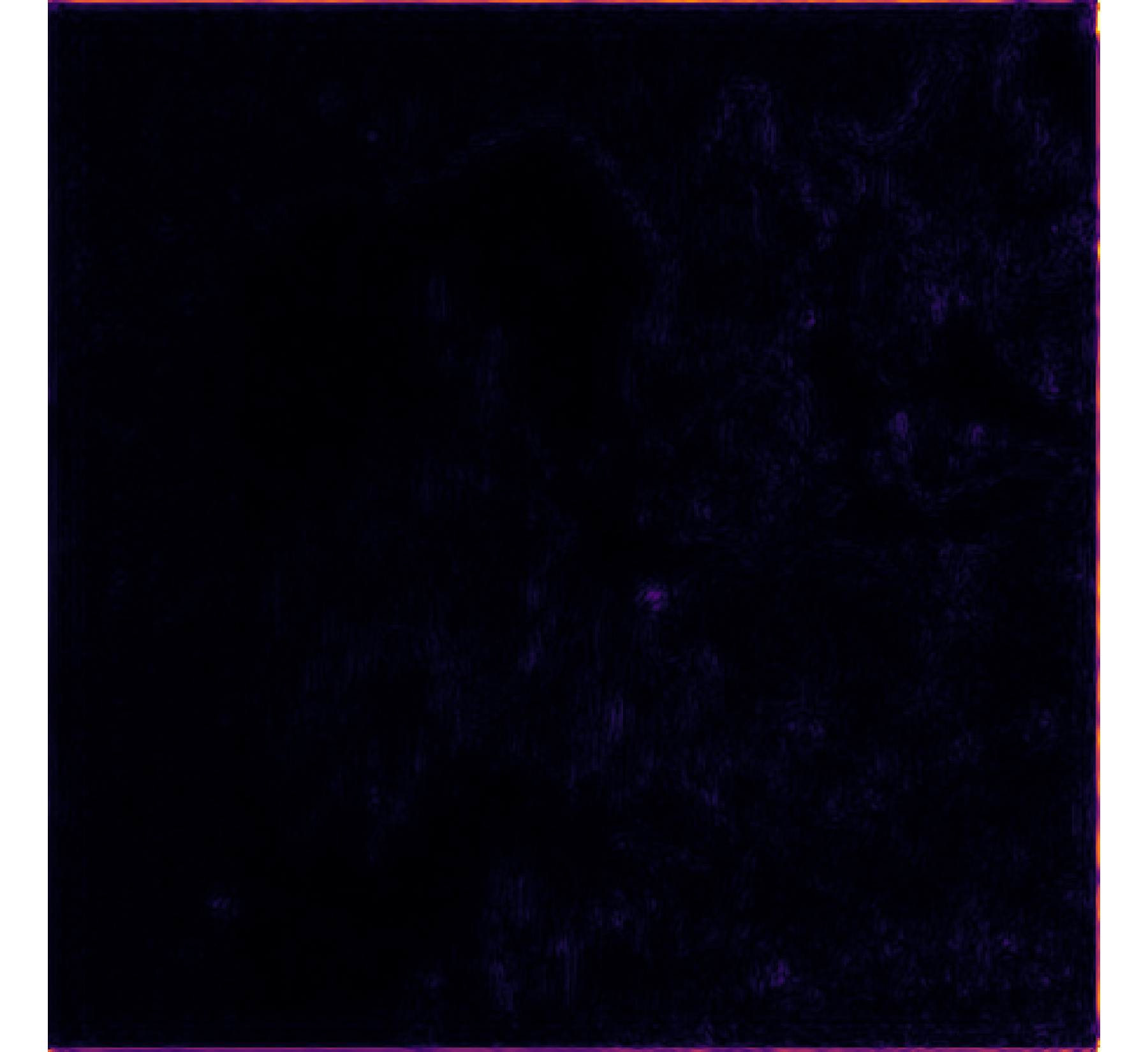} &
    \includegraphics[
        width=0.035\linewidth,
        height=0.24\linewidth,
        keepaspectratio
    ]{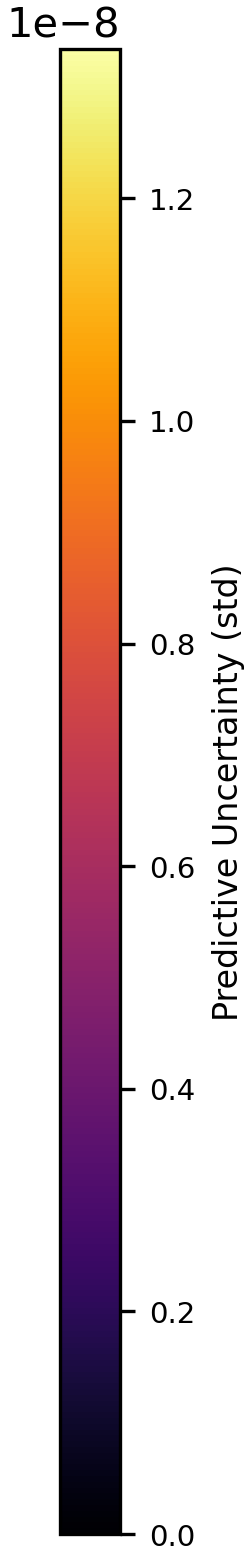}
    \\[1mm]

   \raisebox{15mm}{\rotatebox{90}{\textbf{BD5}}} &
    \includegraphics[
        width=0.24\linewidth,
        height=0.24\linewidth,
        keepaspectratio
    ]{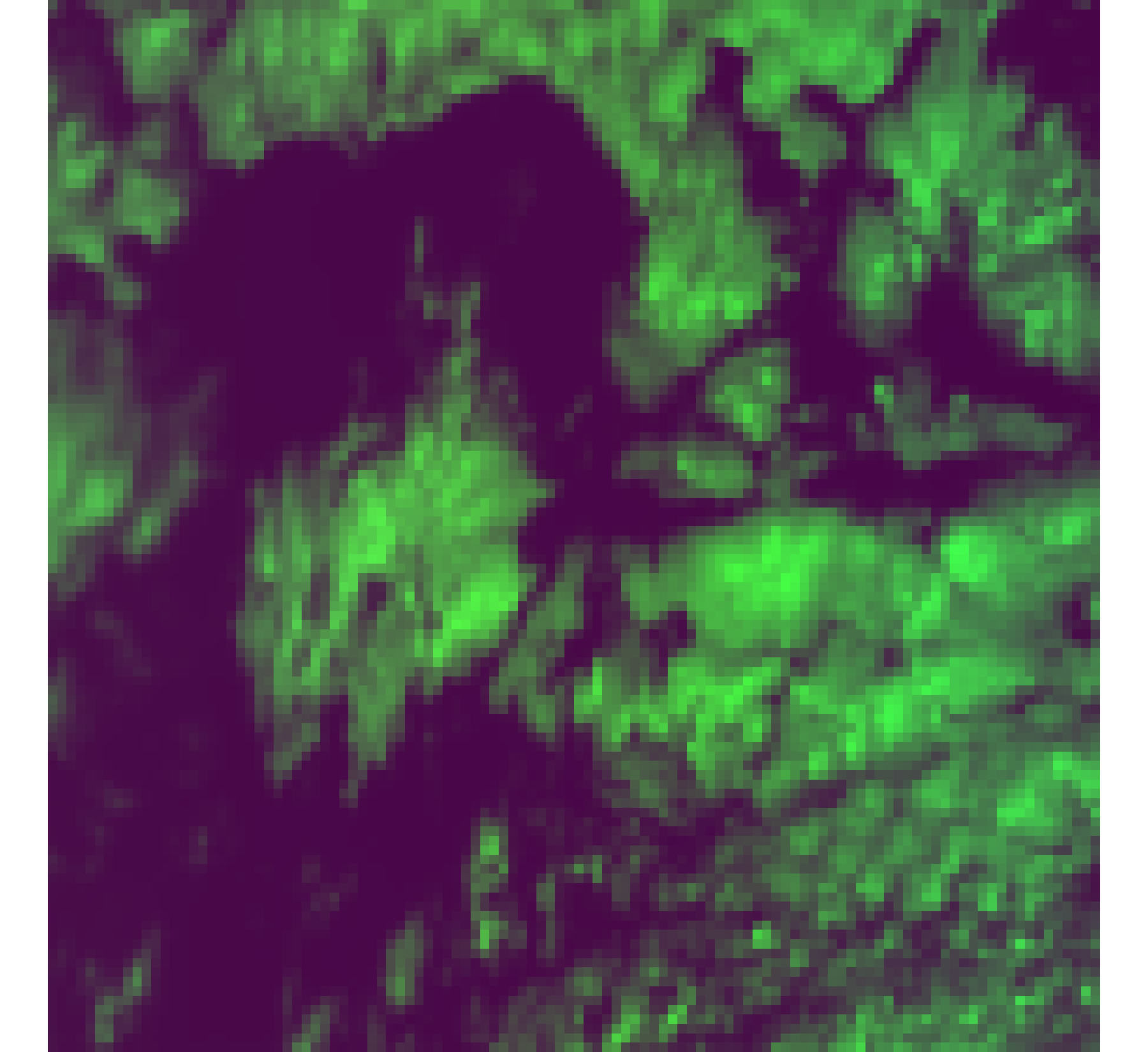} &
    \includegraphics[
        width=0.24\linewidth,
        height=0.24\linewidth,
        keepaspectratio
    ]{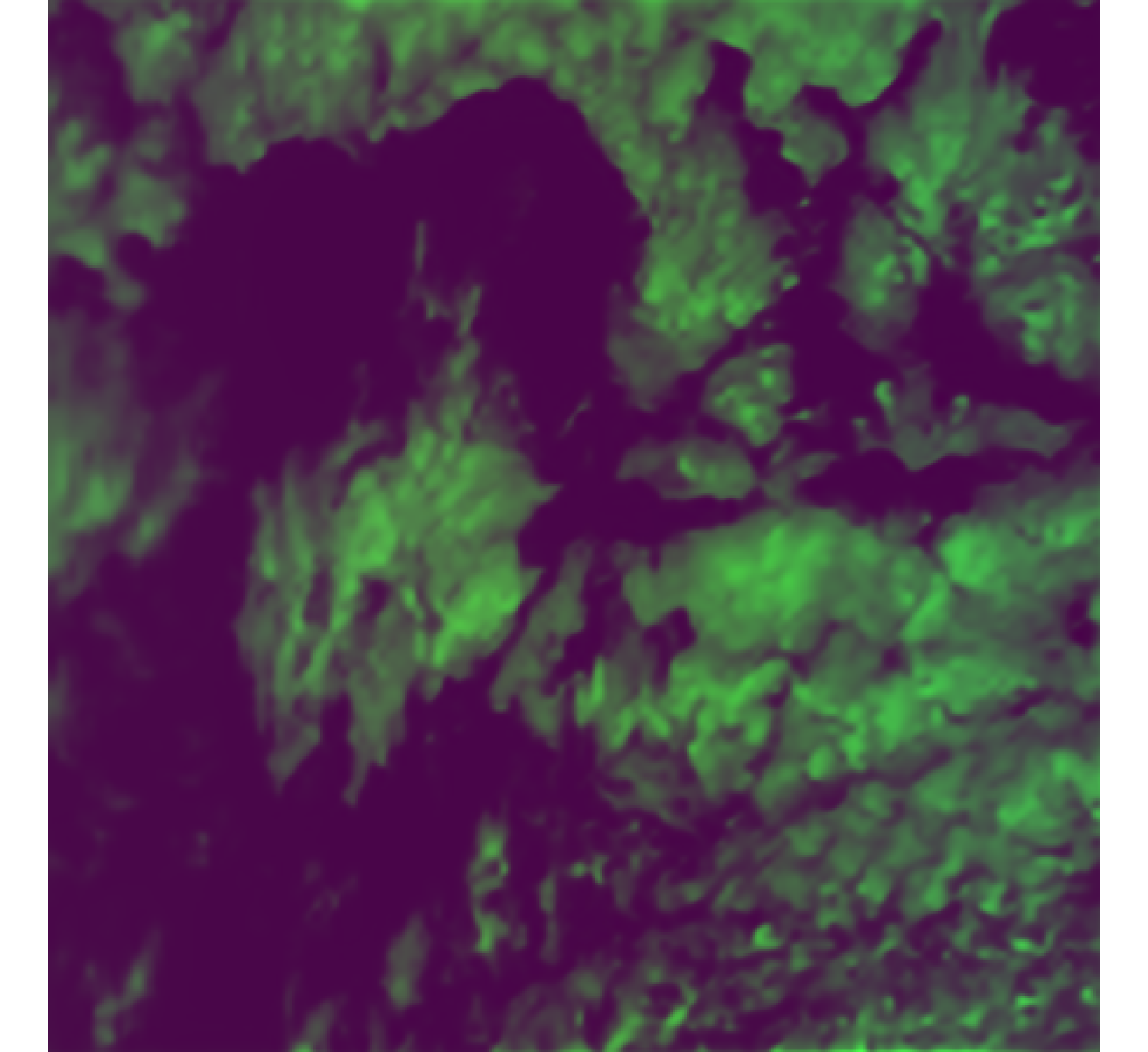} &
    \includegraphics[
        width=0.24\linewidth,
        height=0.24\linewidth,
        keepaspectratio
    ]{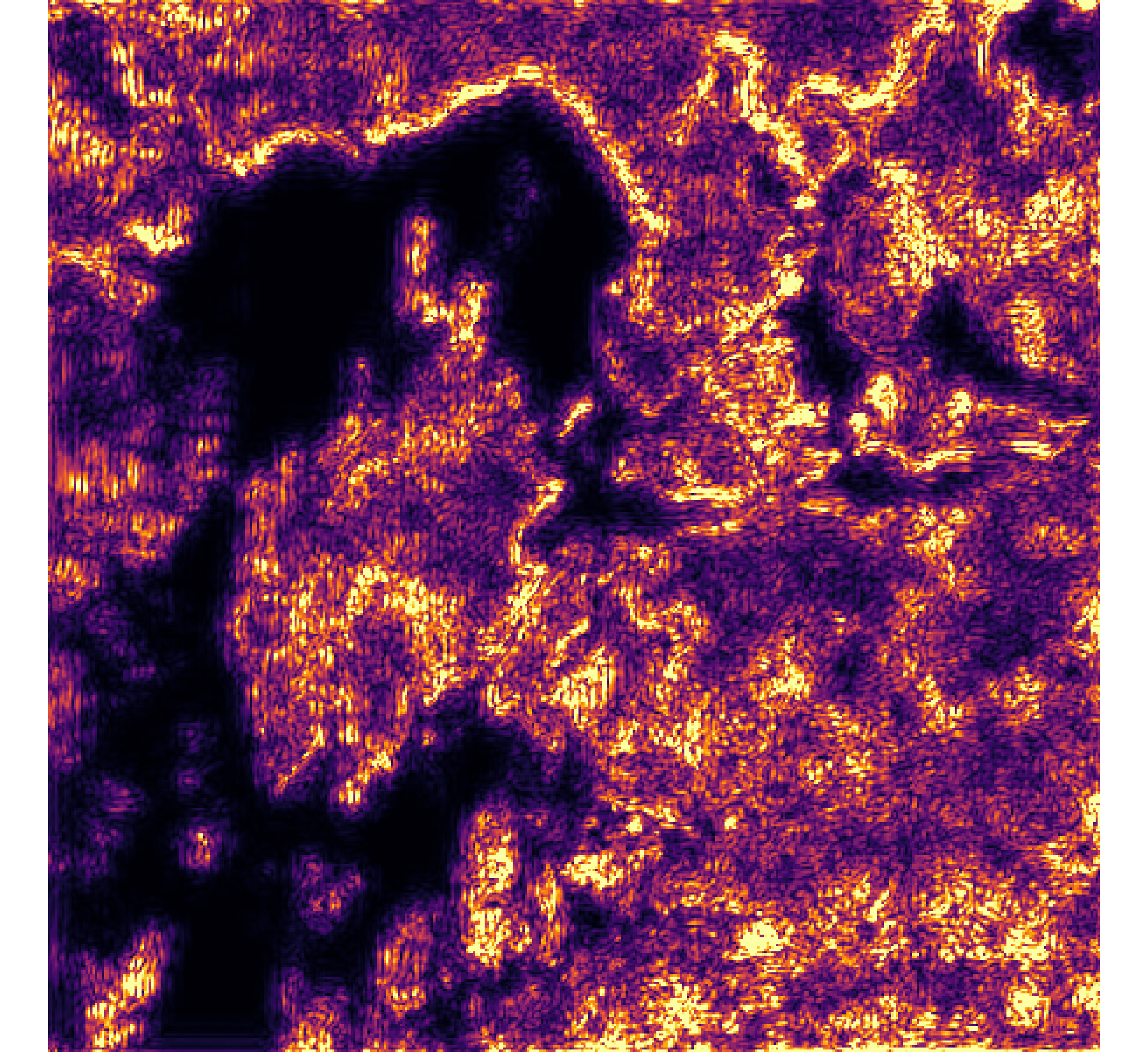} &
    \includegraphics[
        width=0.035\linewidth,
        height=0.24\linewidth,
        keepaspectratio
    ]{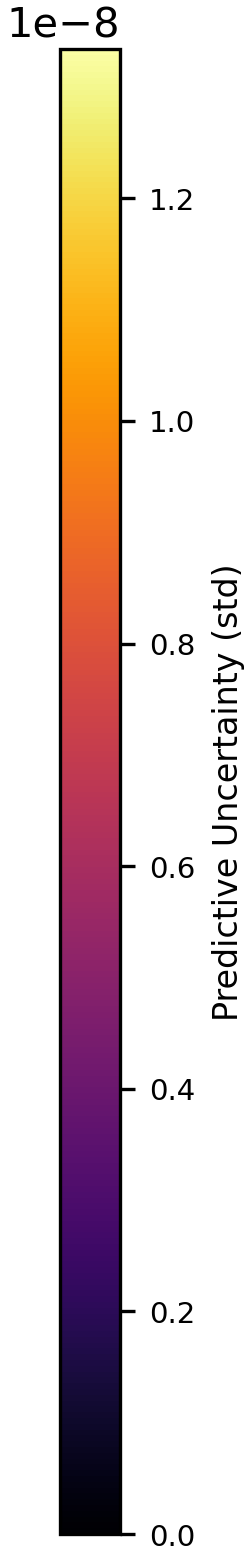}
    \\[1mm]

    \raisebox{15mm}{\rotatebox{90}{\textbf{BD4}}} &
    \includegraphics[
        width=0.24\linewidth,
        height=0.24\linewidth,
        keepaspectratio
    ]{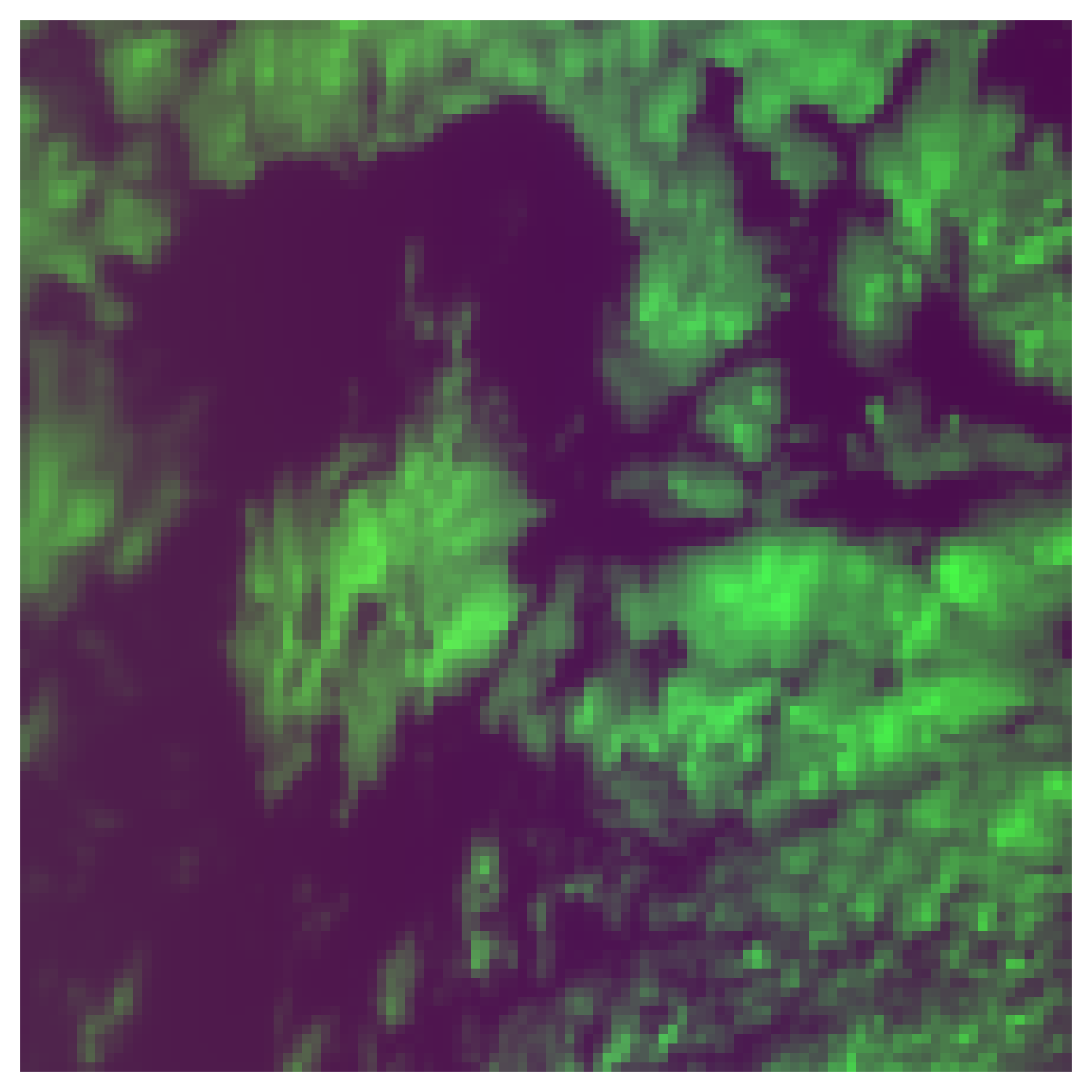} &
    \includegraphics[
        width=0.24\linewidth,
        height=0.24\linewidth,
        keepaspectratio
    ]{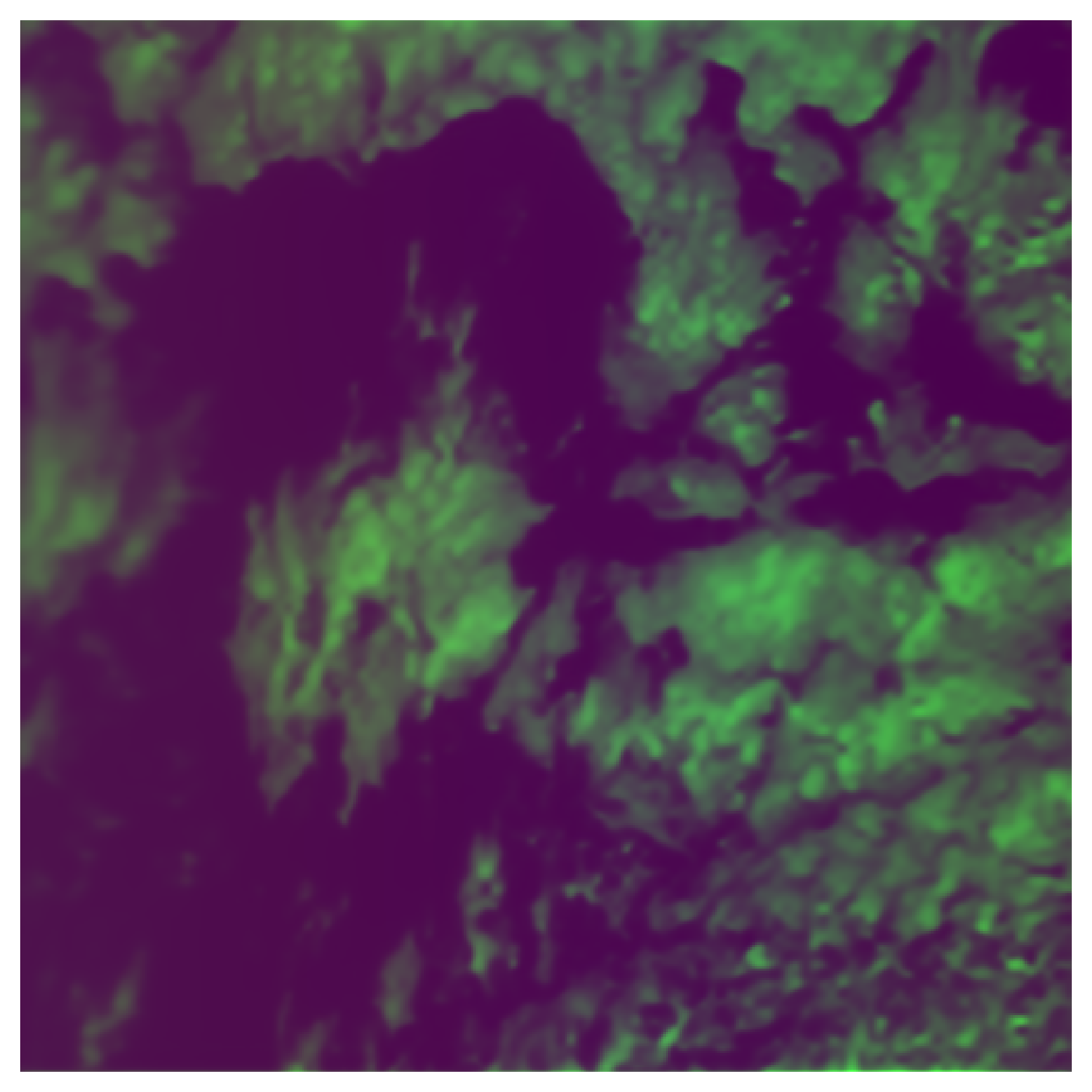} &
    \includegraphics[
        width=0.24\linewidth,
        height=0.24\linewidth,
        keepaspectratio
    ]{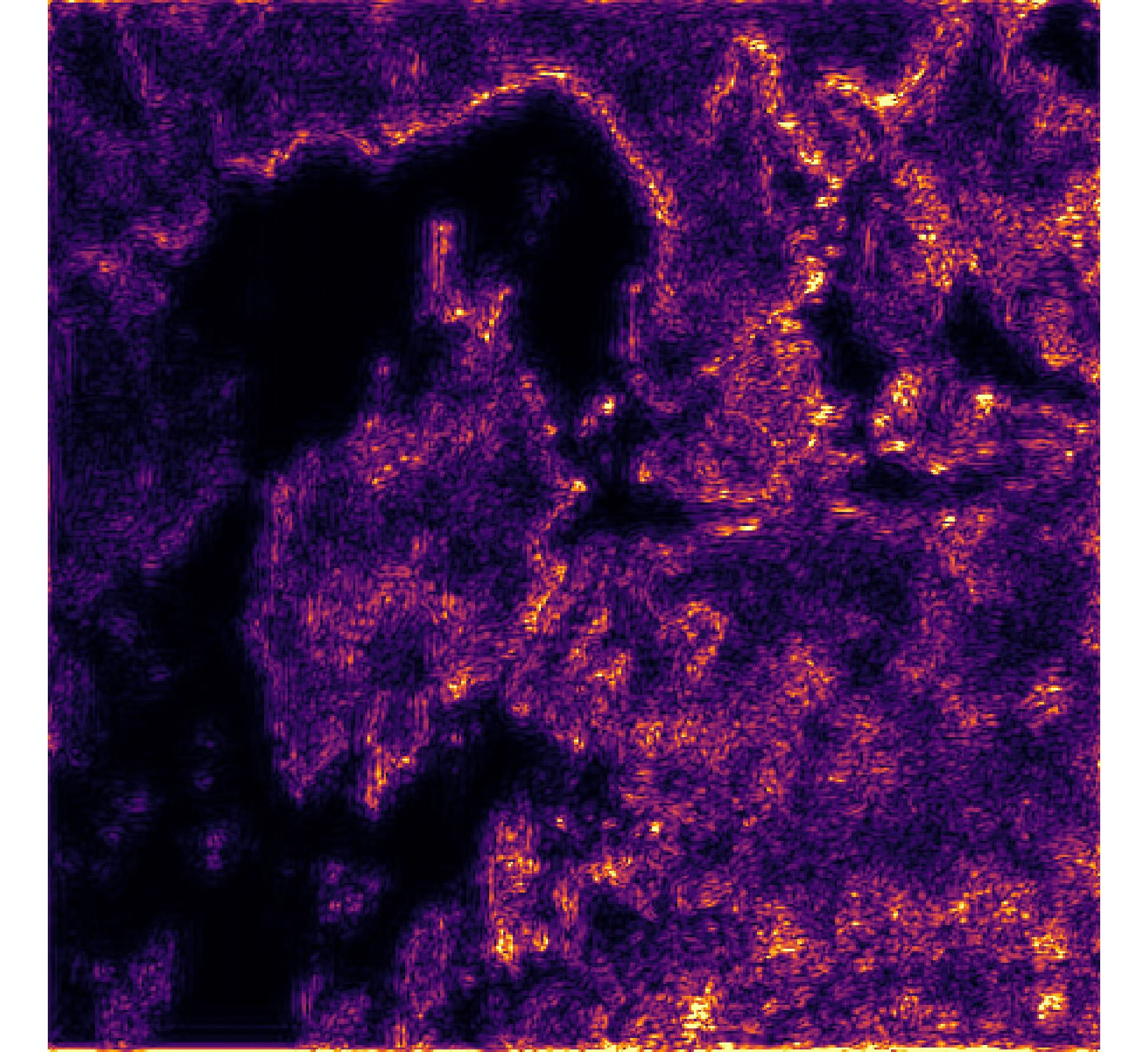} &
    \includegraphics[
        width=0.035\linewidth,
        height=0.24\linewidth,
        keepaspectratio
    ]{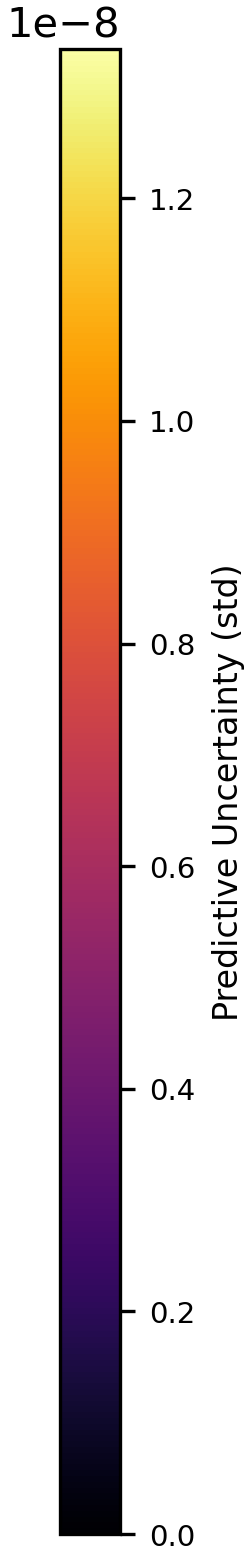}
    \\[1mm]

    \raisebox{15mm}{\rotatebox{90}{\textbf{BD7}}} &
    \includegraphics[
        width=0.24\linewidth,
        height=0.24\linewidth,
        keepaspectratio
    ]{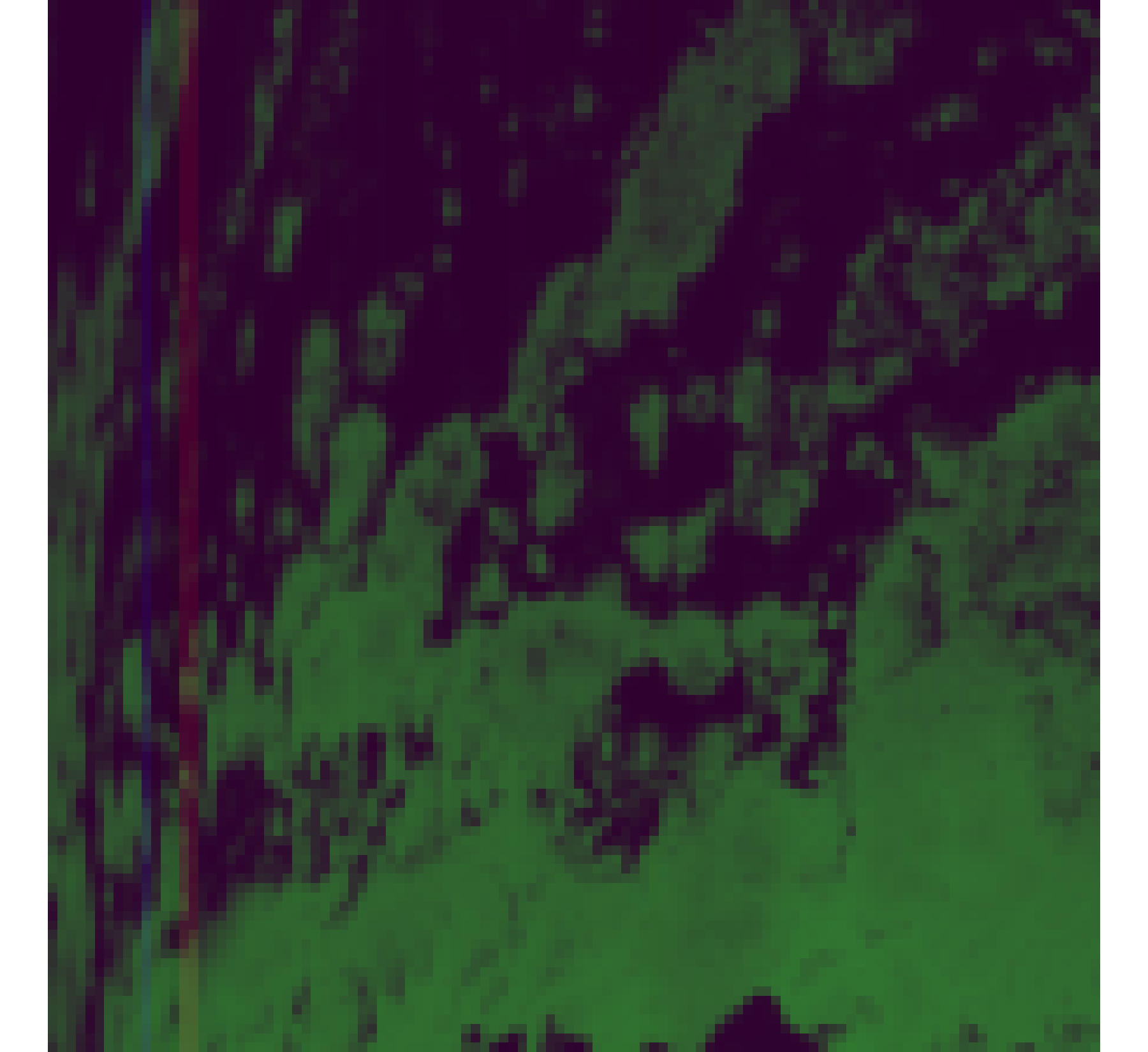} &
    \includegraphics[
        width=0.24\linewidth,
        height=0.24\linewidth,
        keepaspectratio
    ]{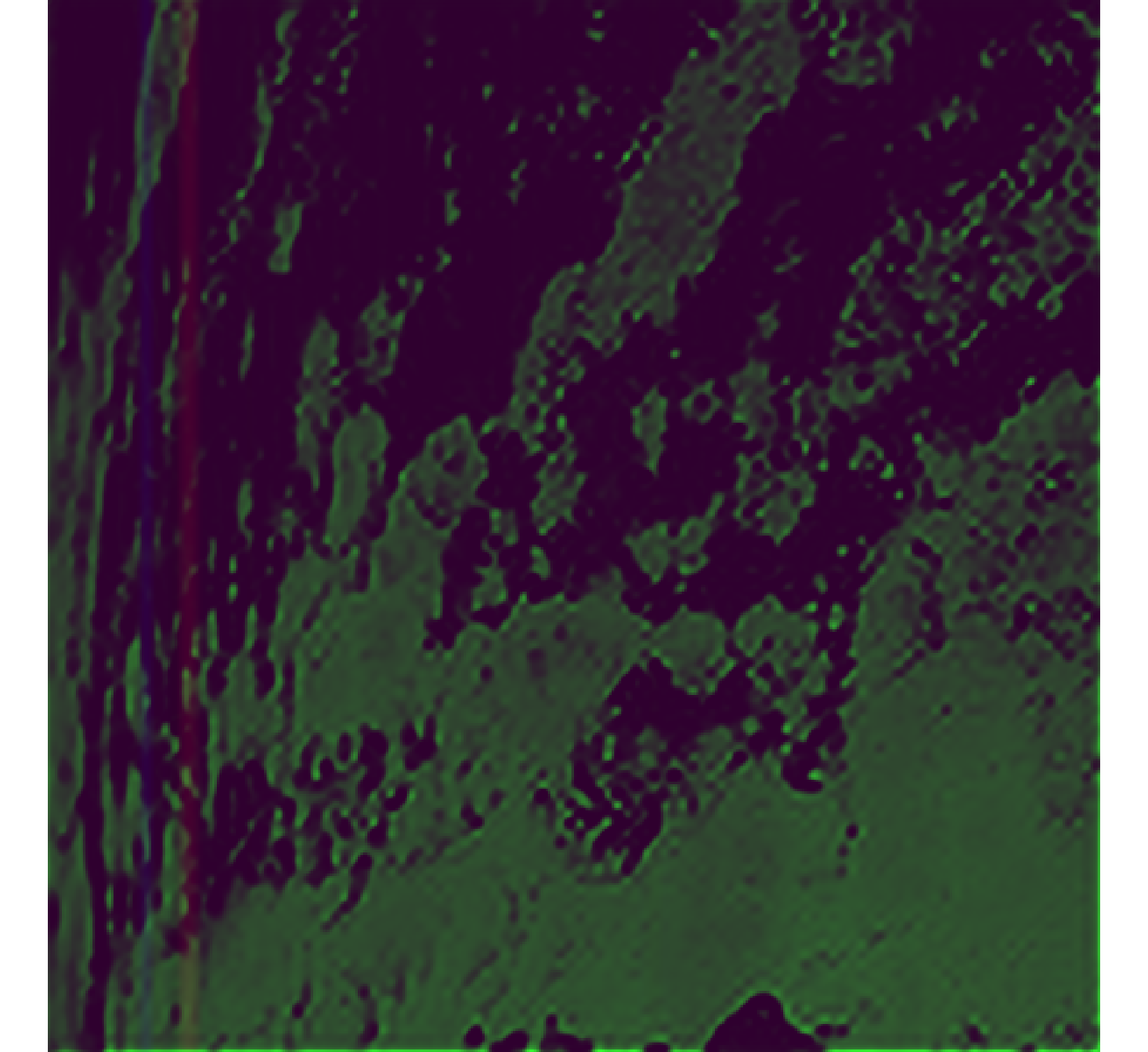} &
    \includegraphics[
        width=0.24\linewidth,
        height=0.24\linewidth,
        keepaspectratio
    ]{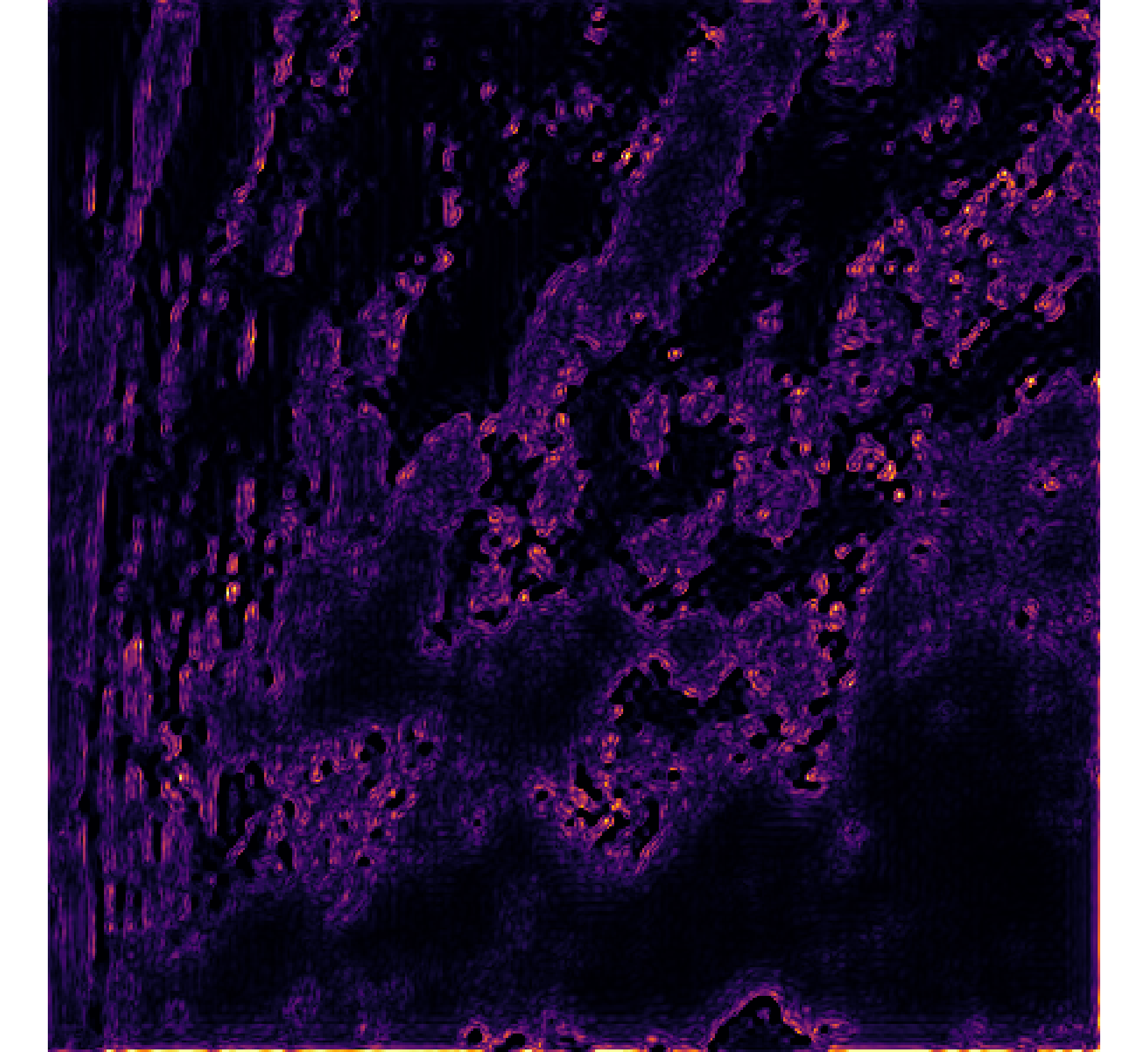} &
    \includegraphics[
        width=0.035\linewidth,
        height=0.24\linewidth,
        keepaspectratio
    ]{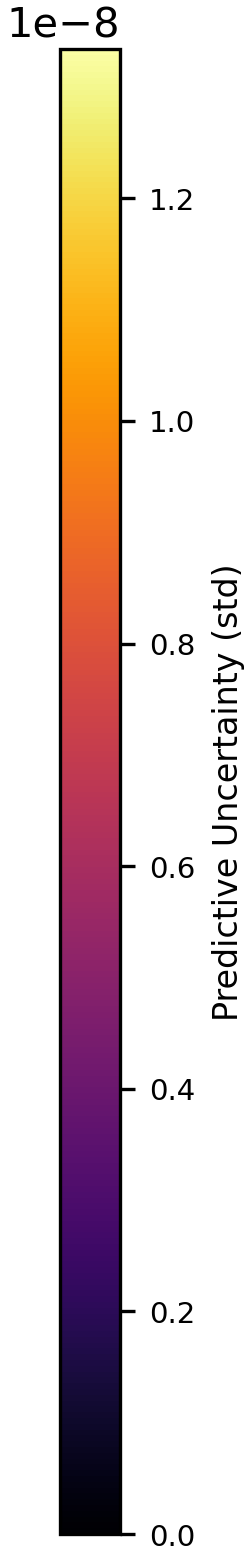}
    \\

\end{tabular}
 \caption{Uncertainty analysis of Unet-S5P-1M using five independently trained models. The ensemble mean is the pixel-wise average of their predictions}
\label{fig:uncertainty}
\end{figure*}

Table \ref{tab:uncertainty} reports the median and 95th-percentile predictive uncertainty over the test set, while Fig. \ref{fig:uncertainty} visualises the representative input images, ensemble-mean reconstructions, and spatial uncertainty maps. Here, the ensemble-mean reconstruction is obtained by taking the pixel-wise average of the predictions from the five independently trained models. The global mean radiance $\mu$ reported in Table \ref{SNR_table} is used as a reference scale for interpreting the magnitude of the predictive uncertainty. The median and 95th-percentile STD values remain below the corresponding $\mu$ values across all evaluated bands, indicating limited variability across the five training runs relative to the characteristic radiance scale. Notably, despite BD7 being one of the more challenging bands, it remains within the same order of magnitude as the other evaluated bands.

\begin{table}[pos=h]
\centering
\caption{Uncertainty of Unet-S5P-1M estimated from five independently trained models}
\label{tab:uncertainty}
\begin{tabular}{@{}lcc@{}}
\hline
\textbf{Band} &
\textbf{Median STD$\downarrow$} &
\textbf{95th-percentile STD$\downarrow$}
\\
\hline

BD2 &
$1.79\times10^{-10}$ &
$1.57\times10^{-9}$
\\

BD4 &
$1.84\times10^{-9}$ &
$8.14\times10^{-9}$
\\

BD5 &
$3.70\times10^{-9}$ &
$1.57\times10^{-8}$
\\

BD7 &
$9.28\times10^{-10}$ &
$3.41\times10^{-9}$
\\

\hline
\end{tabular}
\end{table}

The uncertainty maps further show that the ensemble disagreement is spatially localised. Relatively homogeneous regions generally exhibit low uncertainty, whereas higher uncertainty is concentrated around boundaries and spatially complex structures. This indicates that the variability across independently trained models is primarily associated with regions containing more complex spatial information.

Overall, the results demonstrate stable predictions across independent training runs, while revealing spatially localised uncertainty in regions where reconstruction is more challenging.

\subsection{Model Complexity and Computational Efficiency}
To quantify computational efficiency and reconstruction cost in the GT-SHR setting, we report trainable parameters, computational complexity (GMACs), training time, and reconstruction time. Parameters are obtained from the PyTorch models and reported in millions (M). GMACs are measured with THOP for a single forward pass of one input patch.

Training time for the proposed models denotes the total time required to complete the training procedure, including per-epoch validation but excluding model and result saving. Reconstruction time was measured as the average per-image processing time on the validation set after 20 warm-up forward passes, with CUDA synchronisation before and after each measurement. For DIP and ZSSR, reconstruction time includes the image-specific optimisation and final super-resolved reconstruction.

Table \ref{tab:efficiency} summarises the results for BD2-BD6. S5-DSCR-S provides the lowest model complexity and reconstruction cost among all proposed architectures, while S5-DSCR also remains more efficient than both Unet-S5P variants in terms of parameters, computational complexity, and reconstruction time. Although S5-DSCR-S has the lowest model complexity, its training time is not the shortest, indicating that training time depends on the overall optimisation procedure rather than model complexity alone. For DIP and ZSSR, the reported GMACs correspond to a single forward pass of their respective networks and therefore do not represent the complete computational cost of image reconstruction. Both methods perform image-specific optimization for each input image, which is reflected in their higher reconstruction times. In contrast, the proposed models require only a single forward pass during reconstruction after the training stage. Consequently, all four proposed models achieve substantially lower per-image reconstruction costs than the image-specific baselines, ZSSR and DIP.

\begin{table*}[!htbp]
\centering
\caption{Model complexity, training time, and reconstruction time on  BD2-BD6 (497 spectral channels, $112\times 112$ input)}
\label{tab:efficiency}
\begin{tabular}{@{}lcccc@{}}
\hline
\textbf{Model} & \textbf{Parameters (M)$\downarrow$ }& \textbf{GMACs$\downarrow$ }& \textbf{Training (s)$\downarrow$} & \textbf{Reconstruction (s/img)$\downarrow$} \\ \hline
S5-DSCR-S     & 0.252 & 76.31  & 2090.73  & 0.012 \\
S5-DSCR       & 0.757 & 178.05 & 2263.19 & 0.027 \\
Unet-S5P-800k & 0.807 & 185.94 & 956.16 & 0.030\\
Unet-S5P-1M   & 1.075 & 224.34 & 2364.27  & 0.036 \\
ZSSR          & 0.794   & 160.40   & - & 43.91 \\
DIP           & 1.050   &  6.59    & - & 274.24 \\ \hline
\end{tabular}
\end{table*}

\subsection{Qualitative Validation Using EMIT}
To complement the qualitative evaluation, we perform an independent assessment using HS imagery acquired by the Earth Surface Mineral Dust Source Investigation (EMIT) sensor. Owing to its superior spatial resolution ($60 \times 60\,\mathrm{m}^2$) compared with TROPOMI \cite{thompson2024emitspec}, EMIT provides a valuable reference for visually assessing whether the super-resolved S5P outputs recover spatial structures that are realistic. EMIT data were processed independently and were not used during model training, and their use is strictly limited for evaluation. 

The analysis is based on a rare spatio-temporal coincidence between EMIT and TROPOMI acquired over Luanda, Angola, on 16 November 2025 at approximately 13:00 local time, identified using the OrbNav tool developed by the University of Wisconsin-Madison \cite{wiscSSECorbnav}. Although this represents a single case study rather than a comprehensive cross-sensor validation, it provides an opportunity to compare reconstructed spatial structures over the same geographical region under closely matched acquisition conditions. The scene contains coastline transitions and heterogeneous land-water boundaries with pronounced spatial variations, making it suitable for assessing fine-scale spatial reconstruction.

Because EMIT and S5P have different viewing geometries and acquisition characteristics, differences in atmospheric conditions, illumination, and surface reflectance may affect the observations. Moreover, a formal Simultaneous Nadir Overpass (SNO) is not achievable because of their different orbital configurations \cite{gil2020sno}. Therefore, the comparison is used only to assess the spatial plausibility of the reconstructed structures rather than radiometric agreement or quantitative sensor-to-sensor consistency.

For the comparison, the EMIT radiance product acquired during the coincidence is used as the higher-resolution reference. The proposed SSL framework is evaluated under the GT-SHR setting, and the resulting outputs are restricted to the EMIT spatial footprint, corresponding to approximately 70$\times$140 $\mathrm{km}^2$. To ensure spectral compatibility, the analysis is performed on S5P BD6 ($725$-$775\,\mathrm{nm}$), whose spectral range overlaps with the EMIT observations. For visualisation, both EMIT and S5P HS images are reduced to two-dimensional images by averaging radiance across the spectral channels.

\begin{figure}[pos=h,width=\linewidth]
    \centering
    \begin{minipage}{0.48\linewidth}
        \centering
        \subfloat[EMIT]{
            \includegraphics[angle=90, width=\linewidth]{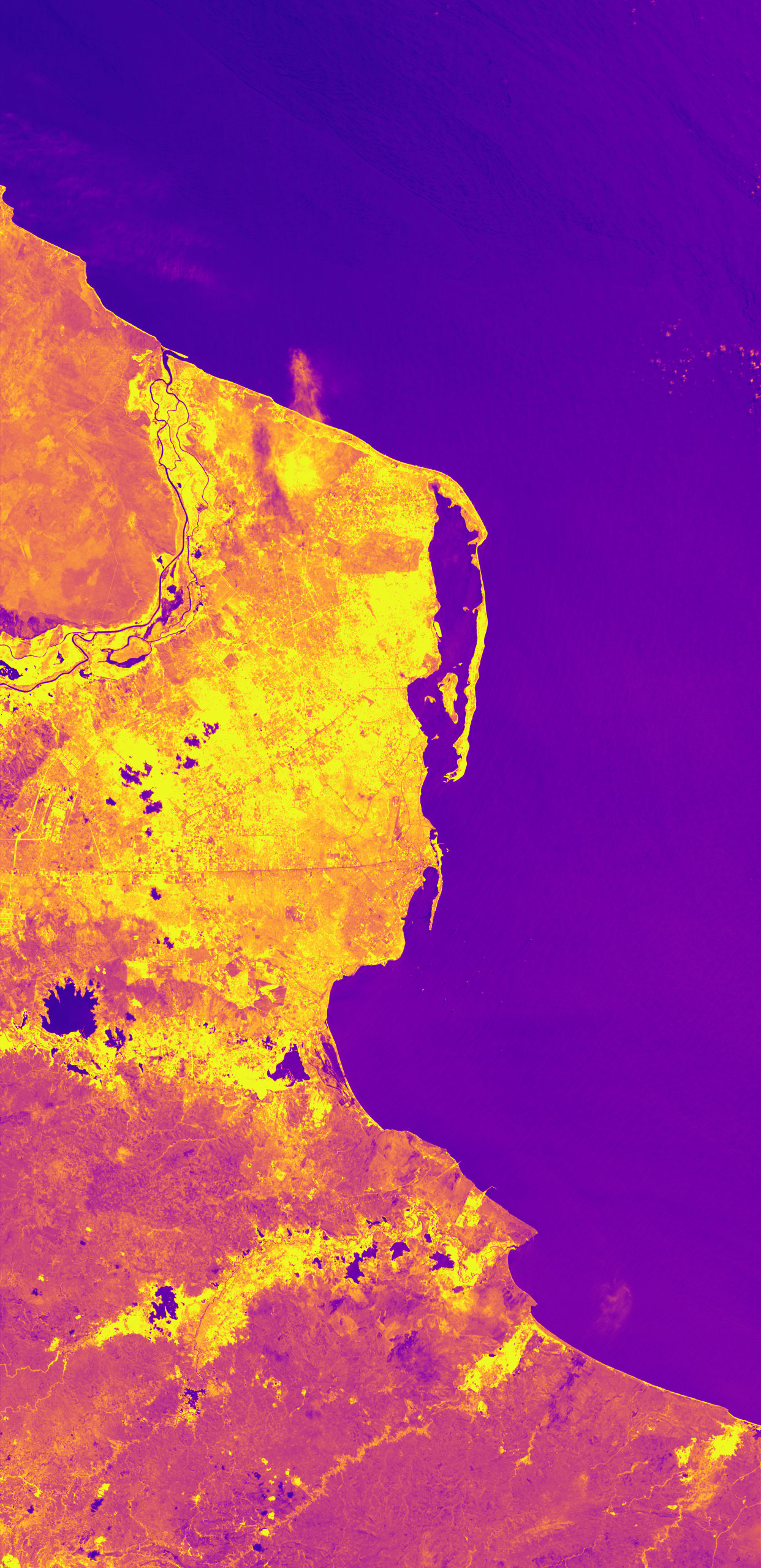}
        }

        \subfloat[Downscaled EMIT]{
            \includegraphics[angle=90, width=\linewidth]{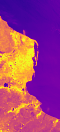}
        }
    \end{minipage}
    \hfill
    \begin{minipage}{0.48\linewidth}
        \centering
        \subfloat[S5P GT]{
            \includegraphics[angle=90, width=\linewidth]{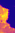}
        }

        \subfloat[Bicubic]{
            \includegraphics[angle=90, width=\linewidth]{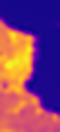}
        }

        \subfloat[SHR]{
            \includegraphics[angle=90, width=\linewidth]{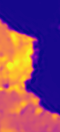}
        }
    \end{minipage}  
\caption{Qualitative cross-sensor comparison between EMIT and S5P for BD6. All images are visualised by averaging radiance across the spectral channels} \label{EMIT_fig}
\end{figure}

Fig. \ref{EMIT_fig} illustrates the qualitative comparison between EMIT, native-resolution S5P, bicubic interpolation, and the SSL SHR construction. The native S5P and bicubic images appear substantially smoother, particularly along coastlines and land-water boundaries. In contrast, the SHR reconstruction produces sharper coastlines and clearer inland water structures, such as lakes, whose shapes and boundaries become better defined. The downscaled EMIT image provides a higher-resolution reference at a comparable spatial scale and shows that these fine structures become progressively less distinct as spatial resolution decreases. This comparison provides complementary evidence that the spatial details recovered by the SSL framework are spatially plausible and correspond to structures visible in higher-resolution imagery, particularly in high-contrast regions.

\section{Conclusions}
This study investigated self-supervised SR of S5P hyperspectral observations in the absence of HR GT. The results support the use of the S5P degradation operator $\mathcal{A}$, band-dependent noise characteristics, and measurement consistency constraints for self-supervised reconstruction from native observations. The combination of SURE and the EI constraint exploits the known image formation process without relying on synthetic HR outputs, while the DSC-based architectures provide an efficient implementation tailored to the spatial and spectral characteristics of S5P.

The experiments show that SSL can approach SL performance in the LR-HR setting while preserving measurement consistency and spatial detail. For real S5P observations, the GT-SHR experiments demonstrate the ability to recover enhanced spatial structures beyond the native resolution. Uncertainty analysis indicates stable reconstructions, while comparisons with generic self-supervised SR methods provide complementary evidence of the effectiveness of the proposed framework. The EMIT comparison further supports the spatial plausibility of the recovered structures, particularly in high-contrast regions. These results suggest that self-supervised SR is a viable strategy for enhancing S5P observations when HR references are naturally unavailable.

An inherent limitation of this study is the absence of HR observations, which limits definitive quantitative assessment of the GT-SHR reconstructions, while the available cross-sensor validation is limited to the evaluated EMIT observations. In addition, the current band-wise training strategy does not explicitly exploit cross-band correlations that may provide complementary spatial and spectral information. Although the proposed degradation model explicitly reflects the S5P acquisition process, its direct transfer to other hyperspectral sensors would require adapting the degradation and noise models to their respective instrument characteristics.

Future work will focus on extending the framework to additional hyperspectral sensors and expanding cross-sensor validation. Uncertainty-aware learning strategies will also be investigated to improve robustness under challenging noise and LR conditions. 

\section*{Acknowledgments}
This work received support under the JUNON Program with the financial support of the R\'egion Centre-Val de Loire (France).


\printcredits

\bibliographystyle{unsrtnat}

\bibliography{cas-refs}

 \appendix
\section{Channel Configurations of Unet-S5P Variants} \label{App_unet}
This section provides detailed channel configurations for the Unet-S5P architectures across different spectral bands. Due to variations in the number of spectral channels between bands, the encoder structures are adapted accordingly, while the decoder structures remain consistent within each model variant.
\begin{enumerate}
    \item Unet-S5P-800k: the encoder consists of three levels of channel compression.
\begin{itemize}
    \item BD2–BD6: $497 \rightarrow 63 \rightarrow 8 \rightarrow 1$
    \item BD7–BD8: $480 \rightarrow 60 \rightarrow 8 \rightarrow 1$
\end{itemize}
The decoder follows a shared expansion method for all the bands: $1 \rightarrow 8 \rightarrow 64 \rightarrow 512$.

\item Unet-S5P-1M: the encoder consists of four levels.
\begin{itemize}
    \item BD2–BD6: $497 \rightarrow 180 \rightarrow 65 \rightarrow 24 \rightarrow 9$
    \item BD7–BD8: $480 \rightarrow 173 \rightarrow 63 \rightarrow 23 \rightarrow 9$
\end{itemize}
The decoder structure is shared across all bands: $9 \rightarrow 25 \rightarrow 70 \rightarrow 195 \rightarrow 542$.
\end{enumerate}



\end{document}